\documentclass[sigconf]{aamas}

\usepackage{balance} %
\usepackage{algorithm}
\usepackage{algpseudocode}
\usepackage{microtype}
\usepackage{amsfonts}
\usepackage{booktabs}
\usepackage{subcaption}
\usepackage{multirow}
\usepackage{xcolor}
\usepackage[capitalise,noabbrev]{cleveref}

\setcopyright{ifaamas}
\acmConference[AAMAS '24]{Proc.\@ of the 23rd International Conference
on Autonomous Agents and Multiagent Systems (AAMAS 2024)}{May 6 -- 10, 2024}
{Auckland, New Zealand}{N.~Alechina, V.~Dignum, M.~Dastani, J.S.~Sichman (eds.)}
\copyrightyear{2024}
\acmYear{2024}
\acmDOI{}
\acmPrice{}
\acmISBN{}

\acmSubmissionID{654}

\title[Causal Explanations for Sequential Decision-Making in Multi-Agent Systems]{Causal Explanations for Sequential \\ Decision-Making in Multi-Agent Systems}

\author{Balint Gyevnar}
\affiliation{
  \institution{University of Edinburgh}
  \city{Edinburgh}
  \country{United Kingdom}
  }
\email{balint.gyevnar@ed.ac.uk}

\author{Cheng Wang}
\affiliation{
  \institution{University of Edinburgh}
  \city{Edinburgh}
  \country{United Kingdom}
  }
\email{cheng.wang@ed.ac.uk}

\author{Christopher G. Lucas}
\affiliation{
  \institution{University of Edinburgh}
  \city{Edinburgh}
  \country{United Kingdom}
  }
\email{c.lucas@ed.ac.uk}

\author{Shay B. Cohen}
\affiliation{
  \institution{University of Edinburgh}
  \city{Edinburgh}
  \country{United Kingdom}
  }
\email{scohen@inf.ed.ac.uk}

\author{Stefano V. Albrecht}
\affiliation{
  \institution{University of Edinburgh}
  \city{Edinburgh}
  \country{United Kingdom}
  }
\email{s.albrecht@ed.ac.uk}

\begin{abstract}
    We present \textbf{CEMA}: \textbf{C}ausal \textbf{E}xplanations in \textbf{M}ulti-\textbf{A}gent systems; a framework for creating causal natural language explanations of an agent's decisions in dynamic sequential multi-agent systems to build more trustworthy autonomous agents.
    Unlike prior work that assumes a fixed causal structure, CEMA only requires a probabilistic model for forward-simulating the state of the system.
    Using such a model, CEMA simulates counterfactual worlds that identify the salient causes behind the agent's decisions.
    We evaluate CEMA on the task of motion planning for autonomous driving and test it in diverse simulated scenarios.
    We show that CEMA correctly and robustly identifies the causes behind the agent's decisions, even when a large number of other agents is present, and show via a user study that CEMA's explanations have a positive effect on participants' trust in autonomous vehicles and are rated as high as high-quality baseline explanations elicited from other participants.
    We release the collected explanations with annotations as the HEADD dataset.
\end{abstract}


\newcommand{\BibTeX}{\rm B\kern-.05em{\sc i\kern-.025em b}\kern-.08em\TeX}
\algrenewcommand\algorithmicrequire{\textbf{Input:}}
\algrenewcommand\algorithmicensure{\textbf{Output:}}

\makeatletter
\gdef\@copyrightpermission{
	\begin{minipage}{0.3\columnwidth}
		\href{https://creativecommons.org/licenses/by/4.0/}{\includegraphics[width=0.90\textwidth]{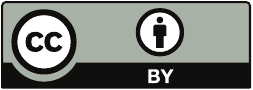}}
	\end{minipage}\hfill
	\begin{minipage}{0.7\columnwidth}
		\href{https://creativecommons.org/licenses/by/4.0/}{This work is licensed under a Creative Commons Attribution International 4.0 License.}
	\end{minipage}
	\vspace{5pt}
}
\makeatother

\begin{document}

\pagestyle{fancy}
\fancyhead{}

\maketitle

\section{Introduction}\label{sec:intro}

\begin{figure}
    \centering
    \includegraphics[width=\linewidth]{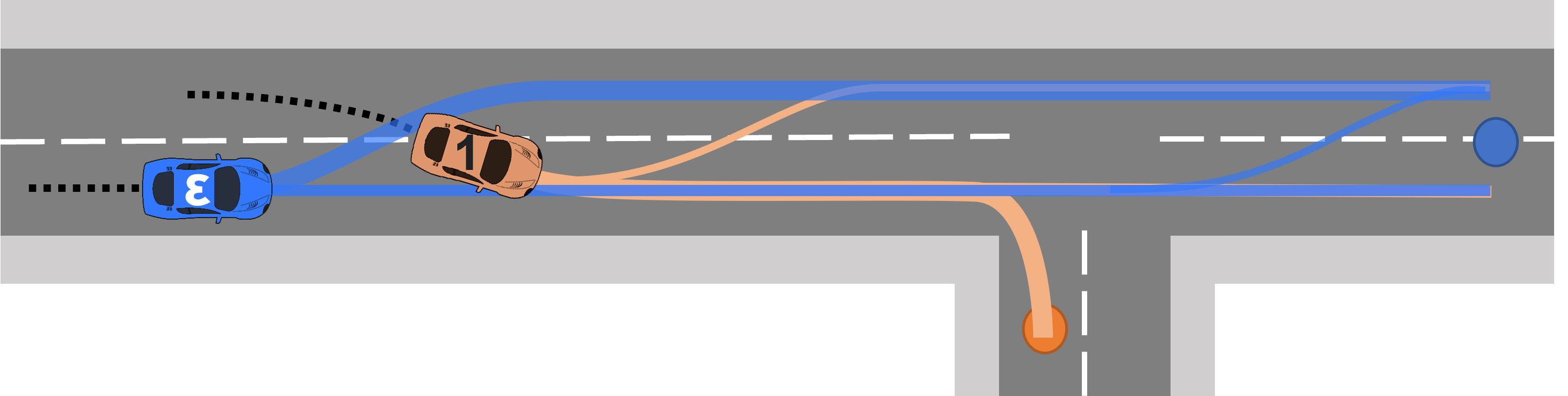}
    \caption{The \textcolor{blue}{autonomous vehicle ($\varepsilon$)} is heading to the blue goal. 
    It decided to change lanes after the \textcolor{orange}{other vehicle (1)} cut in front of it and began to slow down.
    A passenger asks: 
    \textit{Why did you change lanes?} ``To decrease the time to reach the goal.'' [teleological]
    \textit{Why was changing lanes faster?} ``Because the other vehicle is slower than us and is decelerating.'' [mechanistic]
    -- Actual explanations by CEMA with explanation types in brackets. Blue/orange lines illustrate forward simulations using the probabilistic forward model.}
    \label{fig:intro_example}
\end{figure}

Artificial Intelligence (AI) is subject to heightened social and regulatory scrutiny where trust, or a lack thereof, has proven a barrier to public adoption~\cite{kaurTrustworthyArtificialIntelligence2022}, especially in safety-critical systems such as autonomous driving (AD)~\cite{haspielExplanationsExpectationsTrust2018}.
This is in part attributed to the inherent lack of transparency of current black box deep learning-based systems~\cite{aliExplainableArtificialIntelligence2023a}.
In response, explainable AI (XAI) has gained popularity. 
Most XAI methods focus on explanations for supervised learning using tabular or image data~\cite{burkartSurveyExplainabilitySupervised2021}.
However, these explanations are often purely numeric, and alone have little utility for non-experts who lack domain knowledge to understand the system's internal representations~\cite{gyevnarTransparencyGap2023}.
To address this, XAI is increasingly drawing inspiration from philosophy and the social sciences~\cite{millerExplanationArtificialIntelligence2019} which has created what we call the subfield of \textit{social XAI}. %

An essential part of social XAI is the ability to generate causal explanations.
There are several methods for this task~\cite{stepinSurveyContrastiveCounterfactual2021} and some were proposed for causally explaining sequential decision-making in single-agent systems~\cite{tsirtsisCounterfactualExplanationsSequential2021,chakrabortiEmergingLandscapeExplainable2020}.
However, complex and dynamic multi-agent systems, such as the case with AD, involve tightly coupled interactions among agents where the decisions of any one agent may be difficult to explain even for humans, and there have been few works in XAI addressing this problem.
An additional important feature of social XAI is the ability to communicate the extracted causes in the form of intelligible and easy to understand natural language explanations (NLE) as part of a conversational process.
A conversation lets users target the pertinent or unclear actions of the agent, while a social XAI system can adjust the user's mental model without excessive cognitive overhead, thereby contributing to more trustworthy interactions with people~\cite{dazeleyLevelsExplainableArtificial2021}.

To advance the social explainability of multi-agent systems, we introduce a new method called \textbf{CEMA}, which stands for \textbf{C}ausal \textbf{E}xplanations in \textbf{M}ulti-\textbf{A}gent systems.
As illustrated in \cref{fig:intro_example}, CEMA is a social XAI method that generates intelligible causal NLEs about an \textit{ego agent's} decisions in sequential multi-agent systems both in terms of the ego's intrinsic motivations (i.e., teleological explanation) and the actions of other agents in the ego's neighborhood (i.e., mechanistic explanation). %
At the core of CEMA is a novel causal selection algorithm based on the Counterfactual Effect Size Model~\cite{quillienCounterfactualsLogicCausal2023}, which builds on a large body of research into how people select causes for explanations.
Instead of creating a specific fixed causal structure, CEMA only relies on a probabilistic model for forward-simulating the joint state of the system, which makes it generally applicable where such models are available.
By creating counterfactual simulations of what has occurred, CEMA ranks the salient causes behind the ego's actions based on which causes are most correlated with the ego's actions across counterfactual worlds.
Causal selection follows a three-step process:

\begin{enumerate}
    \item \textbf{Roll back} the current factual state of the system to a previous point in time, such that the actions of the ego that we would like to explain have not yet occurred;
    \item \textbf{Simulate} a set of counterfactual worlds from this past time point using a probabilistic forward model of the system;
    \item \textbf{Calculate} the counterfactual causal effect size by correlating the ego's actions with changes in its rewards and actions of other agents across counterfactuals.
\end{enumerate}

We evaluate CEMA on AD using diverse simulated driving scenarios from the literature with expert explanations~\cite{albrechtInterpretableGoalbasedPrediction2021}, we show that CEMA correctly selects causes of the ego's decisions that are congruent with the expert explanations, even when a large number of agents are present.
We show that CEMA is robust to changes in the number of counterfactual simulations and the accuracy of the predictive forward model.
We also perform a user study to measure the perceived quality and effects of CEMA's explanations on people.
First, we collect a set of high-quality human-written explanations as our baseline.
We then show that CEMA's explanations are rated on average at least as high as this baseline while positively affecting participants' trust in AD.
In summary, our contributions are:\footnote{CEMA available at: \url{https://github.com/uoe-agents/cema} \\ HEADD available at: \url{https://datashare.ed.ac.uk/handle/10283/8714}}

\begin{itemize}
    \item CEMA: a framework to generate intelligible causal explanations of the decisions of an ego agent in dynamic multi-agent systems based on the Counterfactual Effect Size Model~\cite{quillienCounterfactualsLogicCausal2023};
    \item Evaluation of CEMA on motion planning for AD, showing its ability to robustly identify correct causes even when a large number of agents are present;
    \item HEADD: a dataset of Human Explanations for Autonomous Driving Decisions consisting of human-written explanations with a minimum of 5 unique annotations regarding the causal content and trustworthiness of the explanations~\cite{gyevnar2024HEADD};
    \item User study showing CEMA's explanations are ranked at least as high as human explanations and a positive effect of CEMA's explanations on trust in AD.
\end{itemize}

\section{Background and Related Work}\label{sec:background}

Causality is a cornerstone of useful human-centric explanations.
A common approach for causal selection is to first model the system in the form of a structural causal model (SCM)~\cite{pearlCausality2009}, but this has some drawbacks for complex and dynamically evolving systems.
First, it is challenging to model all causal factors in the system, such as the state, action, or reward influences, while keeping the SCM interpretable and useful for end users.
Second, the SCM may grow to intractable sizes depending on the desired coverage of causal factors and the complexity of the system.
Third, due to the temporal and non-stationary nature of dynamic systems, an SCM may frequently need to be recomputed to adapt to changes.
Thus, existing work has applied SCMs only in simpler single-agent systems where, e.g., the agent is trained with a specific algorithm~\cite{nashedCausalExplanationsSequential2023,madumalExplainableReinforcementLearning2020}.  

In addition, AI models have grown complex enough that generating explanations by ``opening the black box'', i.e., relying on an understanding of the intrinsic causal properties of the trained model, is often infeasible~\cite{wachterCounterfactualExplanationsOpening2017}.
Instead, we can rely on the \textit{counterfactual model of causation}, which is a well-understood formulation of causation in philosophical literature~\cite{hiltonLogicCausalAttribution1988,lewisCausation1973}.
Counterfactual cases uncover causes in relation to the factual case by highlighting events whose absence resulted in the counterfactual case rather than the factual case.
Implementing the counterfactual model of causation for complex multi-agent systems is challenging in practice.
We rely on~\citet{quillienCounterfactualsLogicCausal2023}'s Counterfactual Effect Size Model which is an empirically validated model to operationalize causal selection based on two assumptions about how humans themselves might select causes for explanations.
First, people cognitively simulate counterfactual worlds by sampling from a distribution over possible alternative worlds that are grounded in, i.e., not too different from the factual world.
Second, people approximate causal effect sizes by correlating variables (i.e., potential causes) in the world with the presence of an outcome across counterfactual simulations.
This means that if we have a probabilistic model for forward-simulating a multi-agent system then we can rank and select the most important causes behind the ego agent's actions by simulating counterfactuals.

Furthermore, how a cause is used for the explanation determines its \textit{explanatory mode}.
We consider Aristotle's system as it stood the test of time and is still frequently used in the modern discourse of philosophy of explanations~\cite{lombrozoFunctionalExplanationFunction2006}. 
Aristotle argued for four modes: mechanistic, teleological, material, and formal~\cite{hankinsonCauseExplanationAncient1998}.
The \textit{mechanistic} mode gives an explanation describing the mechanisms of the cause of a change, while the \textit{teleological} mode explains to what end or goal a change has occurred. 
For example in \cref{fig:intro_example}, ``other vehicle slowing down'' is a mechanistic cause while ``reaching goal faster'' is a teleological cause behind the decision of the blue autonomous vehicle to change lanes.
The material and formal modes stay constant in the systems we study, so we do not consider them.

An increasing body of literature studies the generation of explanations for sequential decision-making.
However, most methods focus on deterministic planning in well-defined domains~\cite{chakrabortiEmergingLandscapeExplainable2020}.
Prior work in explainable reinforcement learning does address single and multi-agent settings in dynamic systems~\cite{qingSurveyExplainableReinforcement2022}, but causal methods are sparser.
\citet{madumalExplainableReinforcementLearning2020} is the first to take a causal approach by building an SCM for the action-influence of agents in model-free RL, while \citet{nashedCausalExplanationsSequential2023} generates explanation by mapping the algorithmic process of solving a Markov Decision Process into an SCM.
Others use surrogate interpretable representations of agents' policies with, e.g., decision trees~\cite{silvaOptimizationMethodsInterpretable2020a} and programs~\cite{vermaProgrammaticallyInterpretableReinforcement2019}. %
We are not aware of methods for social XAI in multi-agent systems. %

We use AD for evaluation, where probabilistic models for forward simulating the system are widely available~\cite{badueSelfdrivingCarsSurvey2021}.
Goal recognition methods predict other agents' future states~\cite{brewittGRITFastInterpretable2021,brewitt2023ogrit}, while motion planning generates optimal behavior for agents~\cite{albrechtInterpretableGoalbasedPrediction2021,hanna2021interpretable}.
Social XAI also received some attention in AD. 
For example, \citet{zhangTacticalLevelExplanationNot2022} found that explanations in terms of purely high-level tactical causes (e.g., lane change, turn) had little effect on drivers' trust, therefore, more fine-grained insights are required, e.g., in terms of relative position or acceleration.
However, prior methods for social XAI in AD do not consider the sequential nature of decision-making~\cite{omeizaAccountabilityProvidingIntelligible2021}, rely on a complex neural model which is impossible to certify for safety~\cite{kimTextualExplanationsSelfDriving2018}, or only provide high-level explanations~\cite{gyevnarHumanCentricMethodGenerating2022}.

\section{CEMA: Causal Explanations in Multi-Agent Systems}\label{sec:method}

We assume that CEMA functions in goal-based sequential multi-agent systems with partial observability, and follow the system definition of~\citet{albrechtInterpretableGoalbasedPrediction2021}.  
Let $\mathcal{I}$ be the set of indexed agents in the environment.
At timestep $t \in \mathbb{N}$, each agent $i \in \mathcal{I}$ is in local state $s^i_t \in \mathcal{S}^i$ and receives a local observation $o^i_t \in \mathcal{O}^i$ that probabilistically depends on $s^i_t$ through $p(o^i_t \mid s^i_t)$.
In addition, agent $i$ selects an action $a^i_t \in \mathcal{A}^i$ in reaction to observations through $p(a^i_t \mid o^i_{1:t})$, where the notation $o^i_{a:b}$ denotes a tuple for the sequence $(o^i_a, \dots, o^i_b)$.
The joint state of all agents is denoted $s_t \in \mathcal{S}$ where $\mathcal{S} = \times_i \mathcal{S}^i$ and similarly for $o_t \in \mathcal{O}$ and $a_t \in \mathcal{A}$.
Further, we assume that agent $i$ is aiming to reach a goal $G^i \subset \mathcal{S}^i$ defined as any partial local state description, such as destination coordinates.
The goal $G^i$ may not be observable to other agents.
If a state sequence $s_{1:t}$ achieves $G^i$ for agent $i$, it receives reward $R^i(s_{1:t}) \in \mathbb{R}^d$ which is a $d$-dimensional vector of reward values where each element in $R^i$ is indexed by a label from a set $\mathcal{R}$ of reward components, such as the time taken to reach the destination.
We define the problem of explaining the actions of a particular ego agent $\varepsilon \in \mathcal{I}$ as creating the explanatory function $f \colon (\mathcal{O}^\varepsilon)^* \times (\mathcal{A}^\varepsilon)^* \rightarrow \mathcal{H}$ that maps a sequence of local observations and actions to an explanation from a set of possible explanations $\mathcal{H}$.
For example, one could define $\mathcal{H} \subset \mathcal{A}^*$, so that an explanation is a partial sequence of actions.  
We use $\hat{s}_{a:b}$ to indicate that the sequence may contain counterfactual states.
We write $s_{x:y} \prec s_{a:b}$ if $s_{x:y}$ is a subsequence of $s_{a:b}$.

We also assume the existence of a probabilistic model that can be used to stochastically forward simulate the system.
These are readily available in existing multi-agent literature, for example, in the form of planners or trained reinforcement learning policies~\cite{huang2023multimodal,ALBRECHT201866}.
Such probabilistic models define a conditional probability distribution over subsequent joint states of the system given previous observations and actions.
We denote this model with $p(\hat{S}_{t+1:n} \mid o^\varepsilon_{1:t},a^\varepsilon_{1:t})$, where $n$ is the last timestep.
In the case when the local state is fully observable to the ego agent (such as in our evaluation), this model can be replaced with $p(\hat{S}_{t+1:n} \mid s^\varepsilon_{1:t},a^\varepsilon_{1:t})$, dropping $a^\varepsilon_{1:t}$ for notational simplicity.
Note, that the goals of other agents remain unobservable even under this assumption.

\subsection{Social XAI Framework}\label{ssec:method:framework}

\begin{figure}
    \centering
    \includegraphics[width=0.95\linewidth]{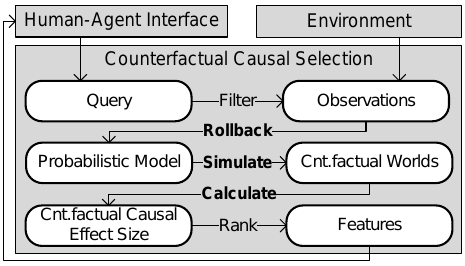}
    \caption[The overall structure of CEMA that generates social explanations for multi-agent decision-making.]{ 
    First, irrelevant observations are filtered out based on the query.
    Second, CEMA rolls back the filtered observations to a previous timestep so that the queried action is erased.
    From then, CEMA simulates counterfactual worlds to calculate the counterfactual causal effect size for the queried actions, which are used to rank the features of the system.
    }
    \label{fig:framework}
\end{figure}

The process of CEMA (\cref{fig:framework}) begins with the user asking a question about an ego agent $\varepsilon$ and an action they would like explained.
The question is parsed by an external human-agent interface into a machine-readable \textit{query}, denoted $q$, encoding a description of the state sequence $\hat{s}_{u:v}$ that corresponds to the ego's queried action.
Here, $u$ is the start timestep and $v$ is the final timestep of the queried action.
Irrelevant states may be then filtered out from the observed states $s_{1:t}$ based on $\hat{s}_{u:v}$. 
For example, if $\hat{s}_{u:v}$ refers to an action in the past ($v<t$), then we can ignore states after timestep $v$.
The queried action $\hat{s}_{u:v}$ need not be a subsequence of $s_{1:t}$, instead it can also be a hypothetical sequence that appears, e.g., in a counterfactual world.
This allows the user to ask contrastive questions, for example of the form \textit{``Why did you not do Y instead of X?''}
The filtered observations and the query are then passed to the counterfactual causal selection module discussed in detail in \cref{ssec:method:causal}.

As the focus of CEMA is to generate intelligible explanations for end users, in this framework explanations are composed from a set of \textit{features} $\mathcal{F}$ which describe semantically meaningful properties of a state and/or action sequence.
For discreet $\mathcal{S}$ and $\mathcal{A}$ with inherent interpretations, the set of features might simply equal $\mathcal{S} \cup \mathcal{A}$.
For continuous spaces, such as in AD, $\mathcal{F}$ might include a discretized summary of actions, such as average acceleration or distance to the leading vehicle.
The set of reward components $\mathcal{R} \subset \mathcal{F}$ are also considered features.
For example in autonomous driving, these might be time to destination or presence of collisions. 
CEMA does not assume anything about the actual meaning or properties of features except that there is some feature function $\phi: \mathcal{S}^* \times \mathcal{A}^* \rightarrow \mathcal{F}$ converting a  state and action sequence to features.
Given the above, for CEMA we define the set of all explanations as $\mathcal{H}=(\mathcal{F} \times \mathbb{R})^*$, so that the output of the counterfactual causal selection process is a subset of features $\mathcal{F}$ with corresponding ranking by counterfactual causal effect size.
Finally, the explanation is converted into an NLE and returned to the user via the human-agent interface.

\subsection{Counterfactual Causal Selection}\label{ssec:method:causal}
The counterfactual causal selection process has three main steps.
First, it rolls back time before the start timestep $u$ of the queried action, erasing the queried action (\cref{alg:counterfactuals}).
Second, this rollback allows CEMA to simulate counterfactual alternatives to the queried action (\cref{alg:counterfactuals}).
Third, the counterfactual simulations inform us about which features of the system are most important for the queried action to occur and we use this information to calculate the counterfactual causal effect size for both the teleological and the mechanistic explanatory mode presented in \cref{sec:background} (\cref{alg:explanations}).

\Cref{alg:counterfactuals} starts by rolling back the joint state sequence $s_{1:t}$ to a timestep $\tau$, such that $\tau \leq u$, resulting in a truncated sequence $s_{1:\tau}$ that assures that the queried action $\hat{s}_{u:v}$ is erased from $s_{1:t}$.
The value of $\tau$ can be a fixed distance from $u$ or it can be determined to, for example, correspond to the start of a distinct qualitative change in the ego's behavior prior to $u$.
The algorithm then performs $K$ number of forward simulations of the system from time $\tau$ according to the probabilistic model $p(\hat{S}_{\tau+1:n} \mid s_{1:\tau})$.
For each simulation, we obtain a sequence of future joint states of the system denoted $\hat{s}_{\tau+1:n}$, determine the reward $r \in \mathbb{R}^d$ for the ego, and whether the queried action $\hat{s}_{u:v}$ of the ego was present in the simulation ($y \in \{0,1\}$).
This process gives a dataset of simulations denoted $\mathcal{D}$.

\begin{algorithm}[t]
    \caption{Counterfactual dataset simulation}
    \label{alg:counterfactuals}
    \begin{algorithmic}[1] %
    \Require Parsed query $q$; observed joint state sequence $s_{1:t}$. 
    \Ensure Counterfactual dataset $\mathcal{D} = \{(\hat{s}^{(k)}_{\tau+1:n},y^{(k)},r^{(k)})\}_{k=1}^K$.
        \State $\mathcal{D} \gets \varnothing$.
        \State $\tau \gets$ Determine from $s_{1:t}$ assuring that $q.\hat{s}_{u:v}$ is erased.
        \For{$K$ iterations}
            \State Get $\hat{s}_{\tau+1:n} \sim p(\hat{S}_{\tau+1:n} \mid s_{1:\tau})$ via forward simulation.
            \State Determine reward for ego $r \gets R^\varepsilon(\hat{s}_{\tau+1:n})$.
            \State Presence of query $y \gets 1$ if $q.\hat{s}_{u:v} \prec \hat{s}_{\tau+1:n}$ else $0$. 
            \State $\mathcal{D} \gets \mathcal{D} \cup \{(\hat{s}_{\tau+1:n},y,r)\}$.
        \EndFor
    \end{algorithmic}
\end{algorithm}

\begin{algorithm}[t]
    \caption{Calculate counterfactual causal effect size}
    \label{alg:explanations}
    \begin{algorithmic}[1] %
    \Require Counterfactual dataset $\mathcal{D}$.
    \Ensure Mechanistic ($\mathcal{F}^m$) or teleological ($\mathcal{R}^t$) explanation.
        \item[\textit{Mechanistic explanation}]
        \State $\mathcal{F}^m \gets []$.
        \For{interval end-point $p_j \in P$}
            \State $\mathcal{D}_j \gets$ Slice $\hat{s}^{(k)}_{\tau+1:n} \in \mathcal{D}$ from $p_{j-1}$ to $p_j$ giving $\hat{s}^{(k)}_{p_{j-1}:p_j}$.
            \State $\mathcal{X}, \mathcal{Y} \gets$ convert $\mathcal{D}_j$ to features $\phi(\hat{s}^{(k)}_{p_{j-1}:p_j})$ and targets $y^{(k)}$.
            \State $\mathcal{M} \gets$ Fit an interpretable classifier to $\mathcal{X}$ predicting $\mathcal{Y}$.
            \State $w, I \gets $ Feature importance attributions $w$ of $\mathcal{M}$ indexed in descending order by $I$.
            \State Append $\mathcal{F}^m_j = \{(\mathcal{F}_i,w_i) \mid i \in I\}$ to $\mathcal{F}^m$.
        \EndFor

        \item[\textit{Teleological explanation}]
        \State $\mathcal{X} \gets$ Filter $\mathcal{D}$ by $y^{(k)}=1$ for match with query.
        \State $\mathcal{Y} \gets \mathcal{D} \setminus \mathcal{X}$, all samples not matching the query.
        \State $w, I \gets \mathbb{E}_{\mathcal{X}}[r] - \mathbb{E}_{\mathcal{Y}}[r]$ indexed by $I$ in absolute desc. order.
        \State $\mathcal{R}^t \gets \{(\mathcal{R}_i,w_i) \mid i \in I\}$.
    \end{algorithmic}
\end{algorithm}

\Cref{alg:explanations} has two parts, one for each mode of explanation.

Mechanistic explanations are formulated in terms of the actions of other agents in the neighborhood of the ego vehicle.
Actions of the other agents can have different causal effects on the ego at different times, so we first increase the granularity of explanations by cutting sequences into $|P|$ slices defined by their end-points $P=(p_1,\dots,p_{|P|})$ with $p_0 = \tau+1$ assumed implicitly.
Each slice is then converted to a set of features using the feature function $\phi$.
Following \citet{quillienCounterfactualsLogicCausal2023}, features that co-occur more frequently with the queried action across counterfactuals should be ranked higher as a salient cause by humans.
Therefore, for each slice $p_j \in P$ of a counterfactual simulation, \cref{alg:explanations} measures the counterfactual causal effect size of features on the presence of the queried action $y$ by correlating features with the presence of the action across the simulated counterfactuals.
For this, an interpretable classifier $\mathcal{M}$ (e.g., logistic regression) is used to predict the presence of the queried action $y$ from the features.
The counterfactual causal effect sizes are given by importance attributions for features from $\mathcal{M}$, giving a mechanistic selection and ranking of features $\mathcal{F}^m_j \in \mathcal{H}$.

Teleological explanations are formulated in terms of the intrinsic reward components of the ego agent.
For this explanatory mode, the counterfactual simulations inform us how the rewards of the ego, as measured by the reward vector $r \in \mathbb{R}^d$, change depending on the presence $y$ of the queried action of the ego. 
For binary $y$, this means that \Cref{alg:explanations} splits $\mathcal{D}$ into two disjoint sets: one where the queried action was observed ($y=1$) and one where it was not ($y=0$).
Following the average treatment effect for randomized controlled trials~\cite{austinIntroductionPropensityScore2011} we take the difference between the expected reward vectors of each set, then order the elements of the difference decreasingly by absolute value, giving a teleological ordering of reward components $\mathcal{R}^t \in \mathcal{H}$ by the causal effect of $y$.

\section{Application to Motion Planning}\label{sec:system}
We give a full demonstration of CEMA's capabilities by applying it to the problem of motion planning for AD which is a challenging reasoning task due to the tightly coupled interactions of many agents in a dynamically evolving system~\cite{schwartingPlanningDecisionMakingAutonomous2018}.
Specifically, we use CEMA to automatically explain the decisions of the Interpretable Goal-based Prediction and Planning (IGP2) system for AD~\cite{albrechtInterpretableGoalbasedPrediction2021}.
We give a summary of IGP2 to the extent necessary for the following sections, but for full details please refer to the original paper.%

The local state $s^i$ of a vehicle $i$ contains its pose (position and heading), velocity, and acceleration
A sequence of temporally adjacent local states is called a trajectory.
Local observation $o^i$ contain the local states of nearby traffic participants.
Actions $a^i$ set low-level controls such as acceleration and steering, while goals $G^i$ are spatial destinations.
Reward components $\mathcal{R}$ are longitudinal and lateral acceleration, presence of collisions, time to reach a destination, and goal completion.
IGP2 uses a hierarchy of systems rather than an end-to-end architecture. 
It defines a set of action sequence templates called \textit{maneuvers} with dynamically generated trajectories for vehicles to follow, including \texttt{lane-follow}, \texttt{lane-change}-\{\texttt{left},\texttt{right}\}, \texttt{turn}-\{\texttt{left},\texttt{right}\}, \texttt{give-way}, and \texttt{stop}. 
Common sequences of maneuvers are then further chained into high-level \textit{macro actions}: \texttt{Continue}, \texttt{Change}-\{\texttt{Left},\texttt{Right}\}, \texttt{Exit}, and \texttt{Stop}.

IGP2 uses macro actions to predict for each non-ego vehicle~$i$ a joint distribution over possible goals and future trajectories given the observed joint local states $s_{1:t}$. 
Monte Carlo Tree Search (MCTS) is then used to forward simulate the world and obtain driving trajectories for the ego vehicle.
In every MCTS simulation, the previously predicted joint goal and trajectory distribution is used to randomly sample a goal and corresponding trajectory for each non-ego vehicle.
MCTS generates a trajectory for the ego in a simulation by sequentially choosing macro actions based on backpropagated preference values (i.e., \textit{Q}-values) until the ego reaches its goal.

\begin{figure*}
    \renewcommand{\thesubfigure}{S\arabic{subfigure}}
    \begin{subfigure}[b]{0.24\textwidth}
        \centering
        \includegraphics[width=\textwidth]{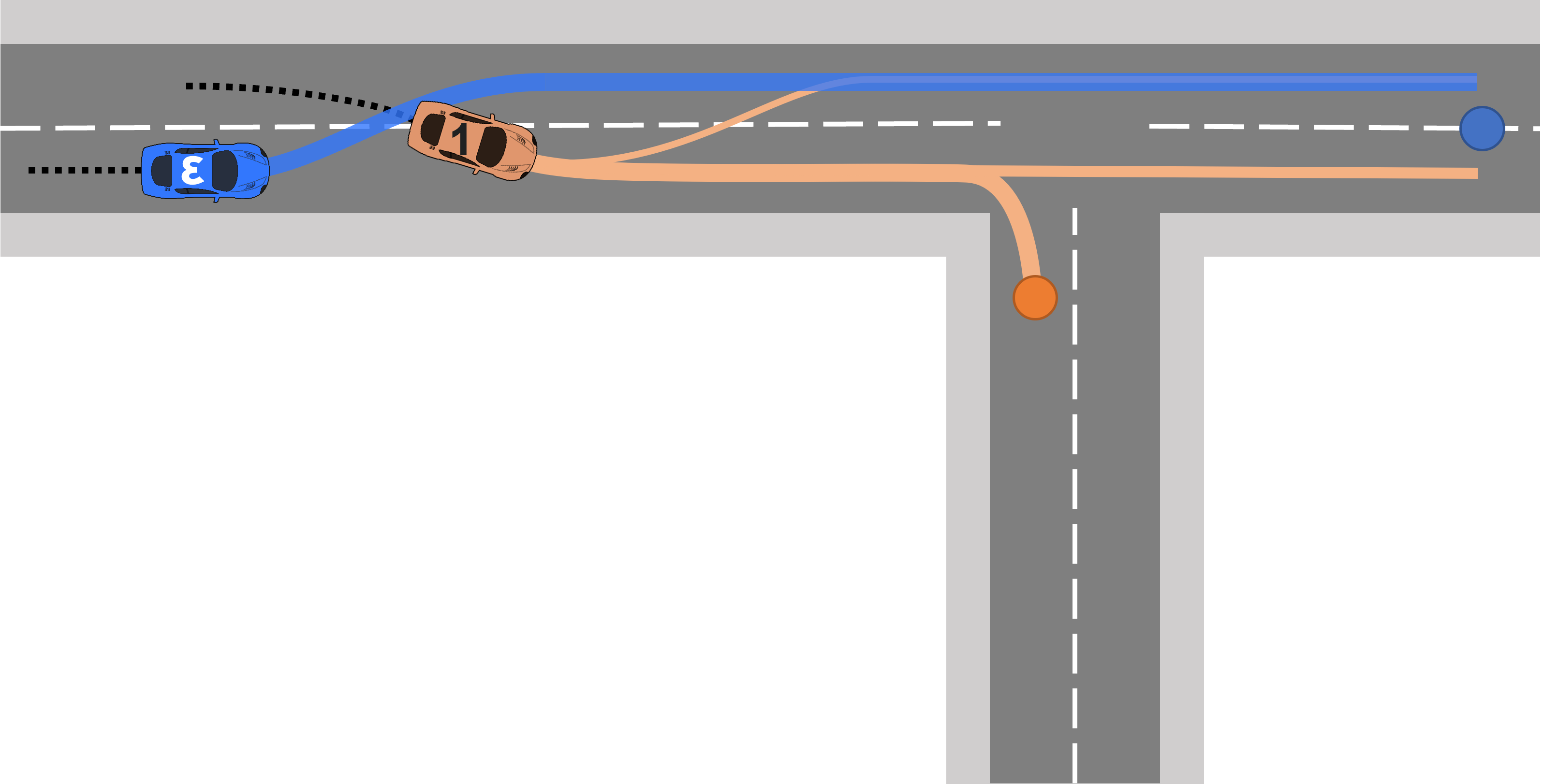}
        \caption{The \textcolor{orange}{non-ego} in front of the \textcolor{blue}{ego} changes lanes and begins to slow down. This is indicative of its intention to turn right at the junction. To avoid being slowed down, the ego decides to change lanes as it is heading straight.}
        \label{fig:s1}
    \end{subfigure}
    \hfill
    \begin{subfigure}[b]{0.24\textwidth}
        \centering
        \includegraphics[width=\textwidth]{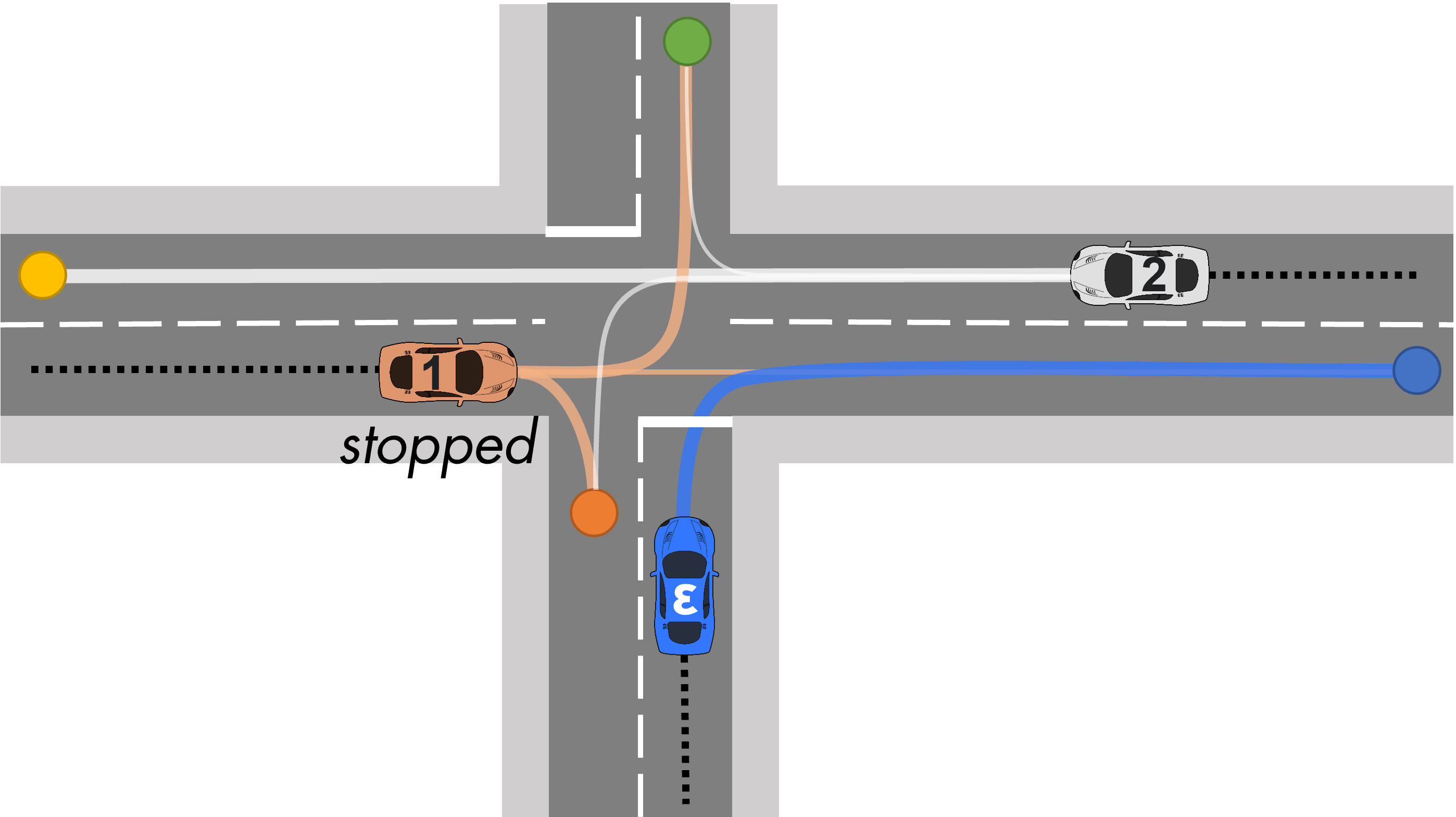}
        \caption{The \textcolor{blue}{ego} is turning right but must give way. It observes the \textcolor{orange}{vehicle} on the left stopping. This is only rational if it is trying to turn left and is giving way for the oncoming \textcolor{gray}{vehicle}. The \textcolor{blue}{ego} can use this to enter the road earlier.}
        \label{fig:s2}
    \end{subfigure}
    \hfill
    \begin{subfigure}[b]{0.24\textwidth}
        \centering
        \includegraphics[width=\textwidth]{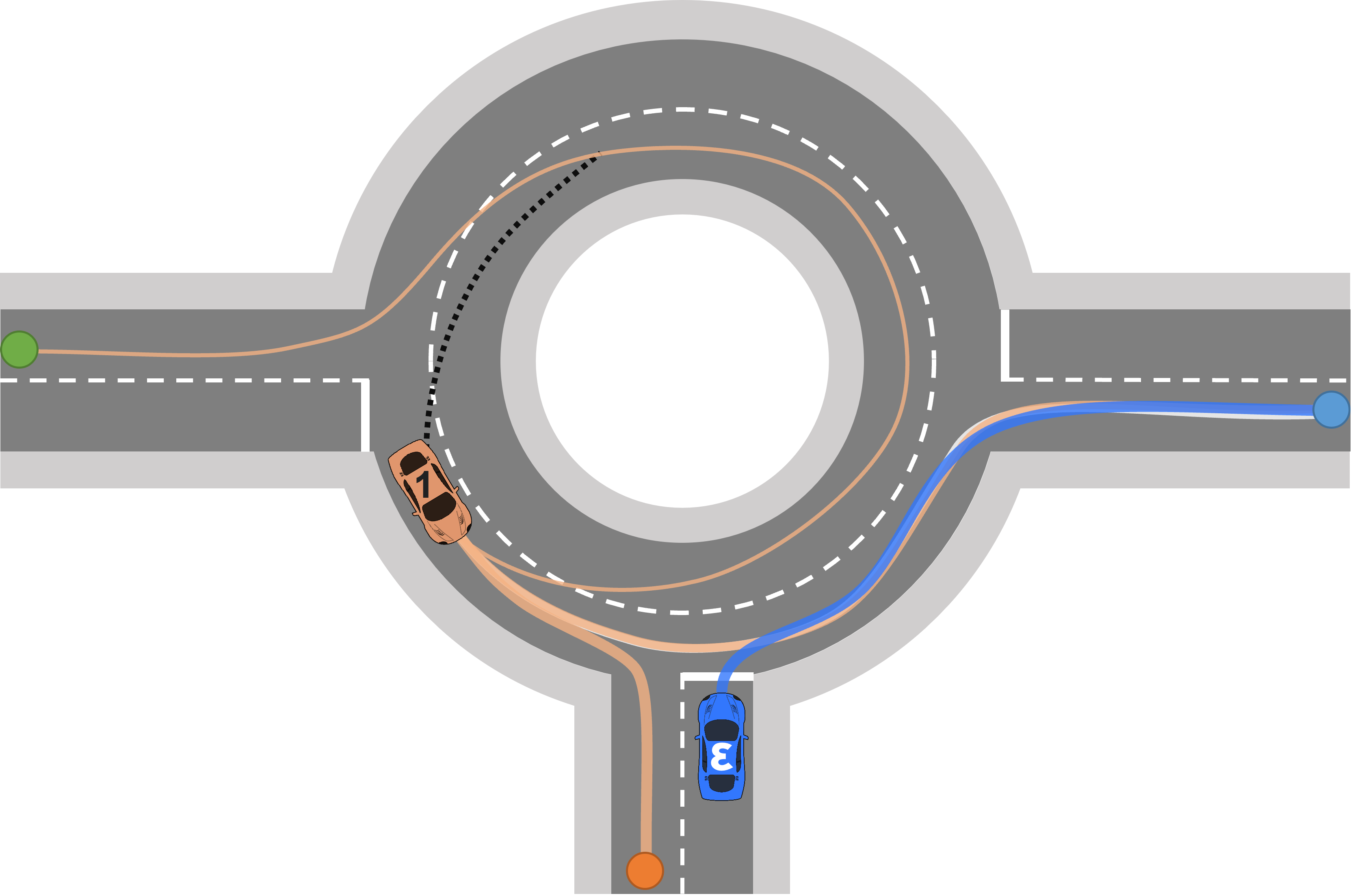}
        \caption{The \textcolor{blue}{ego} observes the \textcolor{orange}{non-ego} changing lanes to the right. This is only rational if the \textcolor{orange}{non-ego} is leaving the roundabout at the next exit. The \textcolor{blue}{ego} can therefore enter the roundabout faster without waiting to give way.}
        \label{fig:s3}
    \end{subfigure}
    \hfill
    \begin{subfigure}[b]{0.24\textwidth}
        \centering
        \includegraphics[width=\textwidth]{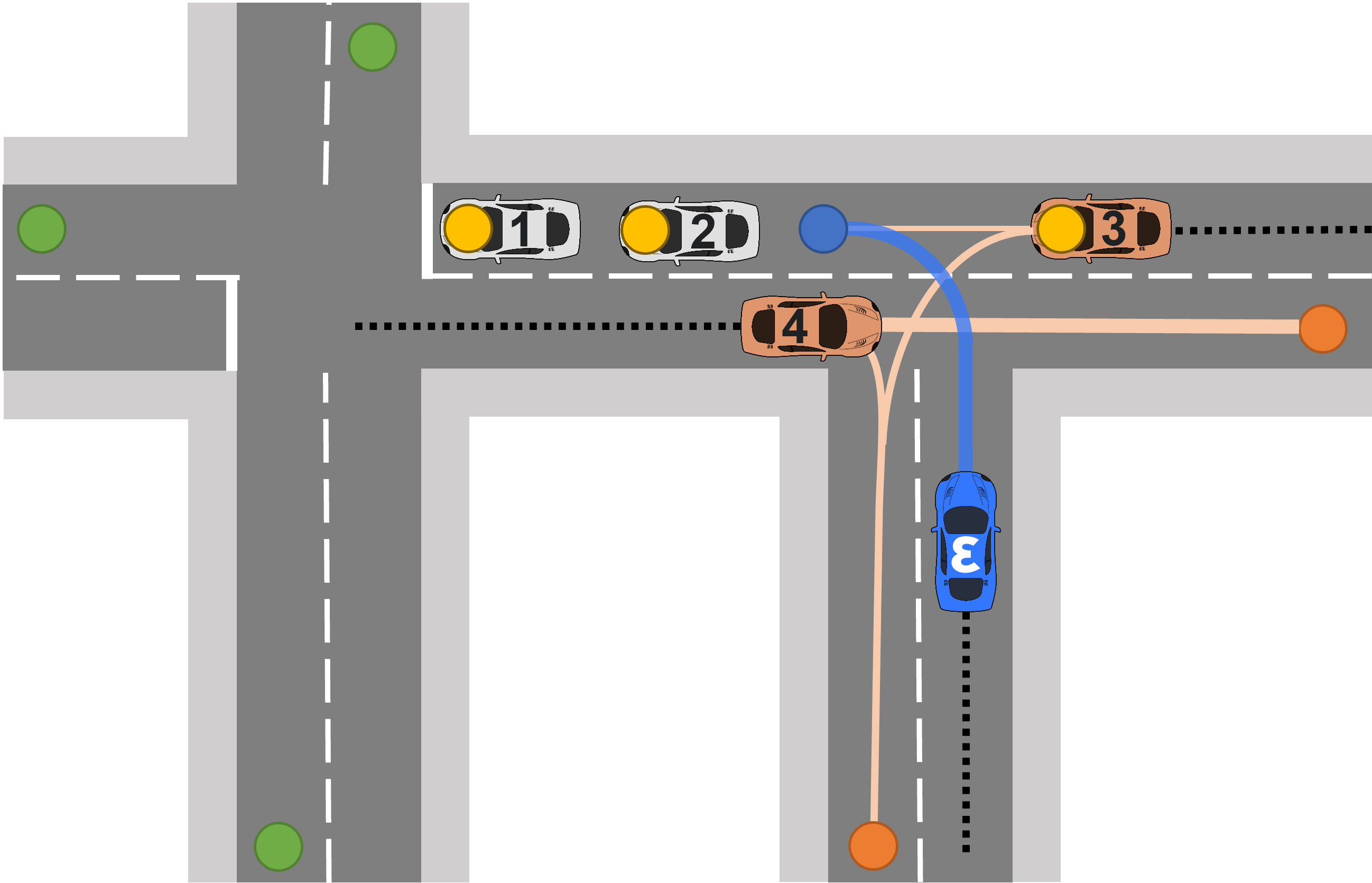}
        \caption{\textcolor{orange}{Non-ego 3} is slowing down to stop. Once \textcolor{orange}{non-ego 4} drives past as indicated by its maintained high speed, the stopping of \textcolor{orange}{non-ego 3} stays rational only if it is to allow the \textcolor{blue}{ego} to merge without waiting for \textcolor{orange}{non-ego 4} to pass.}
        \label{fig:s4}
    \end{subfigure}
    \caption{The four scenarios used for evaluation based on~\protect\citet{albrechtInterpretableGoalbasedPrediction2021}. Colored circles are goals. Solid lines are predicted trajectories of non-egos with thickness corresponding to predicted probability. Black dotted lines are observations.
    }
    \label{fig:scenarios}
\end{figure*}

\begin{table}
    \centering
    \caption{Binary features $\mathcal{F}$ to describe the fundamental motions and high-level actions of vehicles (including ego). For continuous values, the mean value is calculated along the length of the trajectory and thresholded with small value $\delta$.}
    \label{tab:features}
    \begin{tabular}{lll}
        \toprule
        \textbf{Feature} & \textbf{Calculation}    & \textbf{Explanation} \\
        \midrule
        \textit{Acceleration}    & $a^i > \delta_a$                               & Accelerate \\
                        & $a^i < -\delta_a$                              & Decelerate \\
                        & $a^i \in [-\delta_a, \delta_a] $               & Maintain velocity \\
        \textit{Relative}        & $v^i-v^\varepsilon > \delta_v$                 & Faster than ego \\
        \textit{speed}           & $v^i-v^\varepsilon < -\delta_v$                & Slower than ego \\
                        & $v^i-v^\varepsilon \in [-\delta_v, \delta_v]$  & Same speed as ego \\
        \textit{Stop}            & $v^i \in [0, \delta_s]$                        & Does it stop \\
        \textit{Maneuver}        & One-hot encode                                 & Longest maneuver \\
\textit{        Macro Action}    & One-hot encode                                 & Longest macro action \\
        \bottomrule
    \end{tabular}
\end{table}

\subsection{Implementing CEMA}\label{ssec:system:implementation}

We define our set of features $\mathcal{F}$ in \cref{tab:features}, which were chosen to describe both fundamental motions and high-level maneuvers of all vehicles including the ego.
Features average along the length of the trajectory and may encounter the issue that at one timestep they have a positive causal effect, while at a later timestep, they have a negative causal effect, resulting in aggregate zero causal effect.
The slicing operation in \cref{alg:explanations} assures that this issue is avoided.

To focus on causal selection and avoid the ambiguities of natural language, we hand-code each query $q$ to contain a description of the queried subsequence $\hat{s}_{u:v}$ given as a subset of features from $\mathcal{F}$.
For natural language generation, we use a deterministic realization engine called SimpleNLG~\cite{gattSimpleNLGRealisationEngine2009}, which generates a grammatically correct English sentence from a content specification, e.g., subject and verb.
This a better fit than neural generation algorithms, due to a lack of annotated data and hallucinations in neural models. 

Since IGP2 can assign to some (reachable) goals and trajectories near-zero probabilities, we use additive smoothing -- detailed in Appendix~A.3 -- with parameter $\alpha$ to make sure every goal and trajectory can be sampled for the non-ego vehicles.
We then generate two datasets with~\cref{alg:counterfactuals}.
For teleological explanations, we set $\tau=u$, rolling back time just before the queried action of the ego.
This is because teleological explanations are determined by the MCTS reward components which only depend on the ego's present and future actions.
For mechanistic explanations, we set $\tau$ to the start time of the last action prior to $u$, erasing both the queried action of the ego and the action that came before it.
For slicing the trajectories in~\cref{alg:explanations}, we set $P \gets (u,n)$ which slices the trajectory $\hat{s}_{\tau+1:n}$ into a past $\hat{s}_{\tau+1:u}$ and present-future $\hat{s}_{u:n}$ subsequence in reference to the start of the ego's queried action.

We use feature weights from logistic regression with K-fold cross-validation to determine feature importance values.
We found logistic regression to work best as it is simple, inherently interpretable, and all features are binary so their scale does not affect the importance values.

\section{Computational Evaluation}\label{sec:eval}

We evaluate CEMA on the four scenarios (S1--S4) used by~\citet{albrechtInterpretableGoalbasedPrediction2021}.
The scenarios are shown in \cref{fig:scenarios} with expert explanations of the ego's behavior by~\citet{albrechtInterpretableGoalbasedPrediction2021}
In line with our focus on social XAI, we test CEMA on many user queries regarding different ego agents and behaviors, and the generated outputs of CEMA are presented through five simulated conversations (\cref{tab:conversations}), highlighting CEMA's ability to correctly identify the causes behind each queried action.
For all queries, we simulate $K=100$ counterfactual worlds with a smoothing weight $\alpha=0.1$.
Further details about the experimental setup are given in Appendix B.
We focus on S1 for presentation, but all results are confirmed across all scenarios and all presented in Appendix C.
We show that:

\begin{enumerate}
    \item CEMA correctly finds and ranks the relevant causes of the ego's actions that are congruent with expert explanations;
    \item It correctly identifies the relevant causes even in the presence of a large number of agents;
    \item The causal selection process is robust to changes in the sampling size $K$ and the accuracy of the probabilistic model.
\end{enumerate}

\subsection{Correctness of Causal Selection}\label{ssec:eval:causes}

\begin{table}
    \centering
    \caption[Conversations demonstrating the applicability of our framework.]{Actual responses of CEMA to queries (\textit{in italic}).
    (\textbf{S1-A}) The passenger sees on an onboard display, that the ego is planning to change to the left lane. They find this unexpected and inquire. 
    (\textbf{S1-B}) The passenger observes the ego changing lanes and asks for the reasons behind the maneuver while it is ongoing.
    (\textbf{S2}) The passenger observed that the ego had entered the junction without stopping to give way, which could be dangerous.
    (\textbf{S3}) The passenger sees that the ego will turn right at the roundabout without stopping despite the oncoming vehicle. 
    (\textbf{S4}) Once non-ego 4 has passed, the ego merges onto the main road which makes the passenger wonder why the ego did not give way to non-ego 3.}
    \label{tab:conversations}    
    \scalebox{0.96}{

    \begin{tabular}{p{4cm}p{3.8cm}}
        \toprule
        \textbf{Scenario 1 (S1-A)} & \textbf{Scenario 1 (S1-B)} \\
        \midrule
        \textit{Why will you change lanes?} \newline
        It will decrease the time to goal. \newline
        \textit{Why does it decrease the time to the goal?} \newline
        Because vehicle 1 will be slower than us. \newline
        \textit{Why will it be slower?} \newline
        It will decelerate and turn right. \newline
        \textit{What if it hadn't changed lanes before?} \newline
        We would've gone straight.   
         & 
        \textit{Why aren't you going straight?} \newline
        It would increase the time to the goal. \newline
        \textit{Why is it slower to go straight?} \newline
        Because vehicle 1 is slowing down. \newline
        \textit{Any other reasons?} \newline
        It is slower than us. \newline
        \textit{What will you do after this?} \newline
        We will go straight and accelerate. \\

        \midrule
        \textbf{Scenario 2 (S2)} & \textbf{Scenario 3 (S3)} \\
        \midrule

        \textit{Why did you not stop to give way?} \newline
        It would've increased the time to the goal. \newline 
        \textit{Couldn't it have caused a collision?} \newline
        A collision would not have occurred. \newline
        \textit{How did you know?} \newline
        Vehicle 1 was turning left and stopped. \newline
        \textit{What if it went straight?} \newline
        We would've given way and slowed down.  
            &
        \textit{What will you do at the roundabout?} \newline
        We will turn right and accelerate. \newline
        \textit{Would not stopping lead to a collision?} \newline
        Not stopping doesn't cause a collision, but stopping increases the time to reach the goal. \newline
        \textit{How do you know we won't collide with the oncoming car?} \newline
        It has been changing lanes right and is turning right. \\

        \midrule
        \textbf{Scenario 4 (S4)} \\
        \midrule
        
         \multicolumn{2}{p{7.8cm}}{
        \textit{Why are you not stopping to give way?} \newline
        Stopping and giving way would increase our time to reach the goal. \newline
        \textit{Is it safe to turn left early?} \newline
        Accelerating and turning left does not cause a collision. \newline
        \textit{Why not?} \newline
        Because vehicle 3 stops. \newline
        \textit{What if vehicle 3 went straight?} \newline
        We would slow down and give way.} \\ 
        \bottomrule
    \end{tabular}
    }
\end{table}

As shown in~\cref{tab:conversations}, CEMA correctly selects causes which are congruent with the expert explanations of~\citet{albrechtInterpretableGoalbasedPrediction2021}.

\begin{figure}
    \centering
    \includegraphics[width=0.95\linewidth]{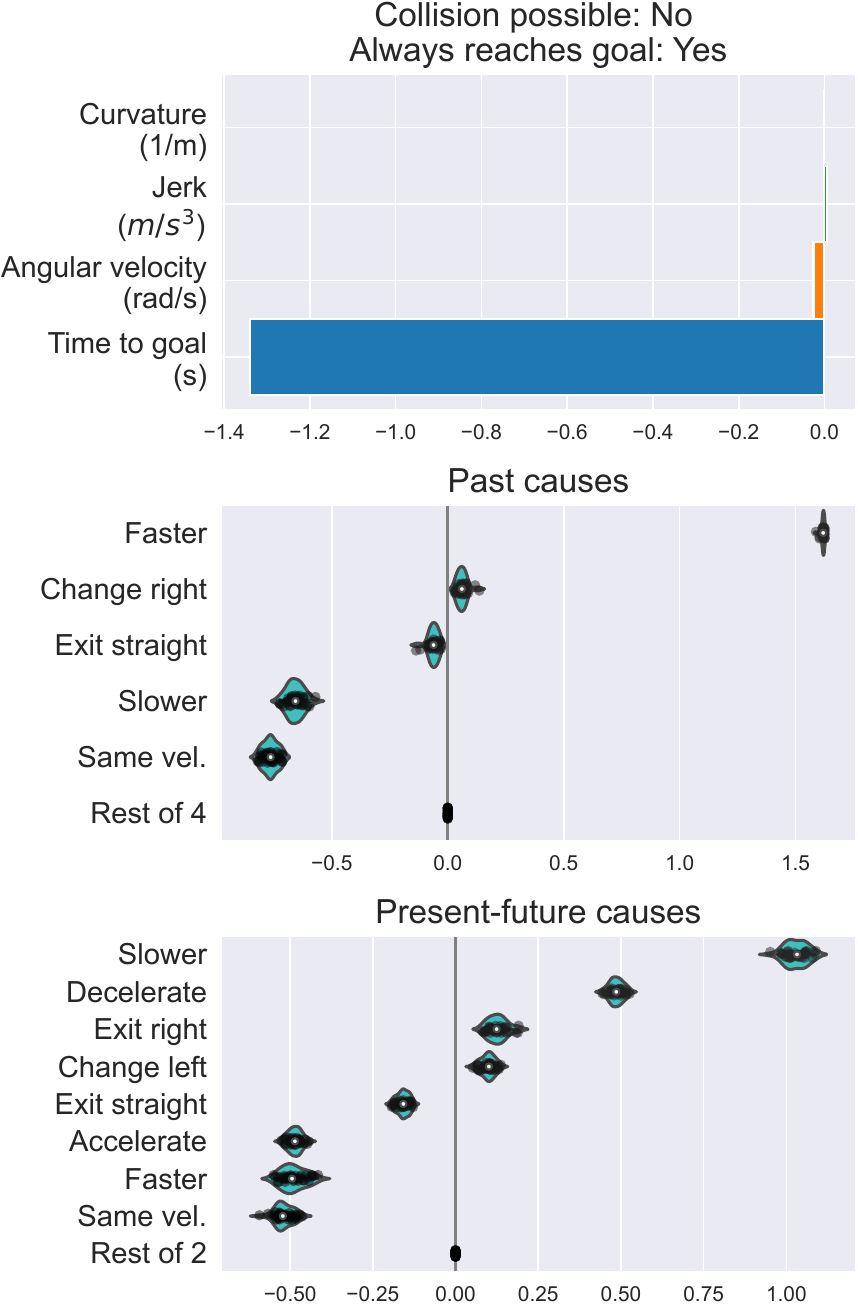}
    \caption[Plots of causal attributions for the first three queries in conversation S1-A for teleological and mechanistic explanations.]{\textbf{[Top]} Signed differences between expected reward components correctly identify time-to-goal as the most significant teleological cause.
    [\textbf{Mid/Bot}] Feature importance attributions for the slice before and during/after the queried subsequence correctly rank mechanistic causes.
    Violin plots show 5-fold cross-validation repeated 7 times.}
    \label{fig:attributions}
\end{figure}

In conversation S1-A, the causes behind the factual lane change of the ego are queried.
The top plot in \cref{fig:attributions} shows that CEMA correctly finds that a decrease in time-to-goal is the most significant teleological cause.
As the bottom plot in \cref{fig:attributions} shows, CEMA correctly identifies that the non-ego slowing down is a mechanistic cause of the ego's lane change. 
CEMA also determines that this slowing down is due to the non-ego vehicle decelerating in order to turn right.
The middle plot of \cref{fig:attributions} confirms that the initially faster non-ego vehicle cutting in front of the ego is also a mechanistic cause of the ego's lane change.
This shows the importance of slicing the trajectories into segments as CEMA produces more fine-grained causes that focus on action in a particular time interval.

In conversations S1-B to S4, we also see that CEMA correctly identifies causes for contrastive questions -- for example, \textit{``Why aren't you going straight?''} -- in which the user asks about an alternative action (i.e., foil) that the ego could have done as opposed to the factual observed actions (i.e., fact).
Leveraging the counterfactual simulations, CEMA contrasts the simulations containing the foil to simulations containing the fact and derives the appropriate teleological causes.
CEMA delivers consistent explanations even when queries target the same action but are phrased differently.
For example, \textit{``Why will you change lanes?''} is a direct question, while \textit{``Why aren't you going straight?''} is contrastive, yet they both refer to the same changing lane action of the ego and CEMA finds consistent causes for both queries.
In S4, CEMA correctly finds that the stopping of non-ego 3 is the most relevant cause behind the ego's early merging behavior and it also finds other intuitive causes.
For example, the vehicle at the front of the waiting line of cars is stopped.
Would this vehicle move, the waiting line of cars would begin moving and non-ego 3 could not allow the ego to merge.

CEMA can also correctly find the relevant causes even when a large number of agents are present.
For this, we greatly increase the number of agents in all scenarios and rerun CEMA. 
For example, we extend S1, adding two extra lanes to the east-west road and increasing the number of agents to 20.
This gave 180 features, most of which had no causal influence on the ego, but CEMA could still identify the most important causes as in the original scenario.

\subsection{Robustness of Causal Selection}\label{ssec:eval:robustness}

We demonstrate robustness to changes in \textbf{(a)} the sampling size $K$, and \textbf{(b)} in the accuracy of the probabilistic simulation model, to show that correct explanations are generated even when sampling is limited by resources and that our system works with prediction algorithms of varying performance.
For size robustness, we randomly sample a dataset of $K \in \{5, 10, \dots, 100\}$ sequences 50 times and calculate the causal attributions for each dataset.
For robustness, we interpolate between the true predicted and uniformly distributed behaviors by increasing the smoothing strength $\alpha$ on a log scale.

\begin{figure}
    \centering
    \begin{subfigure}[b]{0.95\linewidth}
        \includegraphics[width=\linewidth]{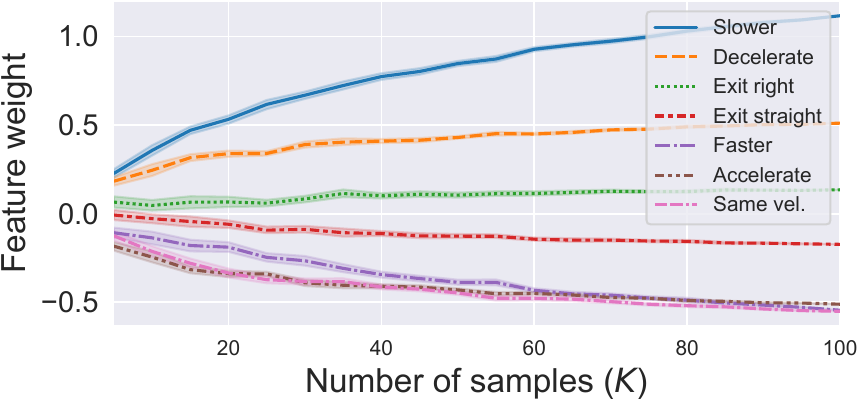}
    \end{subfigure}
    \begin{subfigure}[b]{0.95\linewidth}
        \includegraphics[width=\linewidth]{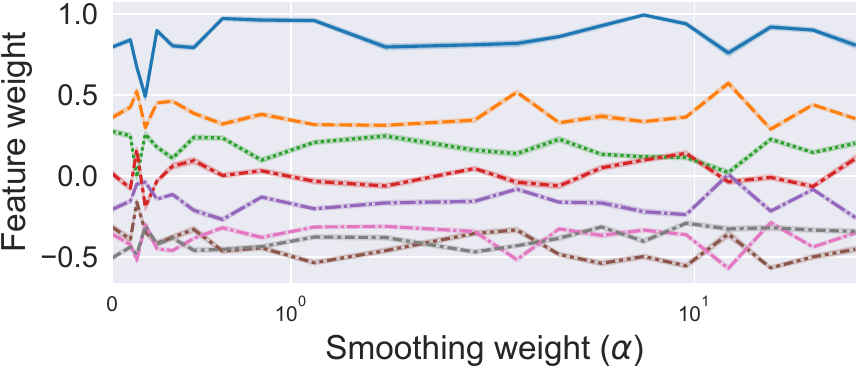}
    \end{subfigure}
    \caption{Changes to causal attributions with [\textbf{Top}] different sample sizes and [\textbf{Bot}] different smoothing weights for present-future mechanistic causes in conversation S1-A. Shaded regions are bootstrapped 95\% confidence intervals. }
    \label{fig:coef_evol}
\end{figure}

The top plot in \Cref{fig:coef_evol} shows the evolution of causal attributions as we increase $K$ in S1.
We see that CEMA becomes increasingly confident in its attributions as $K$ increases, while confidence intervals remain tight.
Even with few samples, CEMA identifies causes correctly.
The bottom plot of \Cref{fig:coef_evol} shows how causal attributions change as $\alpha$ increases which corresponds to increasing uncertainty in behavior predictions.
We see that feature importance values are little affected by changes in the sample distributions as they fluctuate around the same values.
Similar patterns are observed across scenarios, which demonstrates that CEMA is robust to changes in both the sampling size and the accuracy of external predictions.

\section{User Study}

So far, we have focused on the technical details of CEMA.
Ultimately, however, the primary target of CEMA is non-expert end users, so we must evaluate the quality of CEMA's explanations and their various effects on humans with actual participants via a user study.
We aim to answer the following research questions:

\begin{enumerate}
    \item How do people perceive the quality of CEMA's explanations as compared to a human baseline?
    \item What are the effects of CEMA's explanations on people's trust in autonomous vehicles?
\end{enumerate}

We used Prolific to recruit participants from the USA whose first language is English.
As most people have not had first-hand experience with autonomous vehicles (AV), we engaged them via animated videos of the scenarios.
We design two surveys and summarize our methodology below with full details in Appendix D.

In the first survey (N=54; Male=25, Female=29), participants were asked to describe and explain in their own words the behavior of the AV in all four scenarios.
This gave 408 explanations across scenarios, of which we excluded 26 vacuous responses (e.g., \textit{``I don't know", ``None", etc.}), and annotated the remaining explanations with a different set of participants regarding their causal content, overall quality and complexity, and trustworthiness.
We release an extended version of this annotated dataset of natural language explanations, called the \textit{Human Explanations for Autonomous Driving Decisions (HEADD)} dataset, containing 14 scenarios with several agents and environmental elements, including occlusions, pedestrians, and 1308 explanations. 
We collected explanations as we are not aware of any reproducible and publicly available methods for AD that would allow for a meaningful comparison to CEMA's explanations.\footnote{We explored ChatGPT as a baseline, but it was inadequate as its responses were very inconsistent and only sometimes correct (see Appendix D.4 for details).}
Comparing against a human baseline is also a better fit for CEMA as its explanations are intended to have low cognitive overhead and be easy to understand.
In contrast, more complex expert explanations would likely be less effective for end users~\cite{dazeleyLevelsExplainableArtificial2021}.

In the second survey (N=200; M=99, F=101), we designed two tasks, one for each research question.
First, to measure the quality of explanations, we asked participants to rate a random sample of 10 explanations from a set of 30 explanations (5 from CEMA and 25 from HEADD with the highest quality ratings) for each scenario on a 5-point Likert scale.
With 50\% chance, we highlighted in bold the explanations from CEMA.
Second, to measure trust, we use the 9 trust scales proposed by~\citet{hoffmanMetricsExplainableAI2019} adapted to the AD domain.
We use a between-subjects design: half the participants are shown the trust scales prior to ranking explanations, and the other half after having ranked explanations.
We also asked participants about their driving experience and previous exposure to AVs using the SAE automation scale~\cite{saeinternationalTaxonomyDefinitionsTerms2021}.
We hypothesize that \textbf{H1:} the explanations generated by CEMA are scored on average as highly as the human baseline explanations; \textbf{H2} participants who saw explanations from CEMA have on average higher levels of trust than those who have not. %
We analyze our data by fitting linear mixed-effects models for each hypothesis.
We report the estimated means ($\hat{\beta}$) and standard errors ($\sigma$) for each variable and use the Wald test~\cite{waldtest1943} to determine whether the effects of a variable are statistically significant on the outcome.

For \textbf{H1}, we found that CEMA's explanations were rated significantly higher when its explanations were not highlighted and were not significantly worse when they were highlighted. 
On average, explanation ratings ($\hat{\beta}_0$=$3.31$, $\sigma$=$0.08$) were marginally lower for human-written explanations ($\hat{\beta}$=$-0.16$, $\sigma$=$0.08$, $p$=$0.21$), and ratings were significantly lower for human explanations when CEMA's explanations were not highlighted to participants ($\hat{\beta}$=$-0.22$, $\sigma$=$0.08$, $p<0.05$).
Variations across scenarios were negligible ($SD$=$0.07$). 
We also found that people tend to rank CEMA's explanations higher when they had exposure to AVs previously ($\hat{\beta}$=$0.1$, $\sigma$=$0.06$, $p=0.09$).

For \textbf{H2}, we found that, on average, participants' trust ratings ($\hat{\beta}_0$=$1.53$, $\sigma$=$0.5$) were significantly higher after seeing explanations ($\hat{\beta}$=$0.11$, $\sigma$=$0.05$, $p<0.05$), which aligns with expectations from literature~\cite{omeizaAccountabilityProvidingIntelligible2021}. 
Participants' trust also increased significantly when they rated CEMA's explanations higher ($\hat{\beta}$=$0.35$, $\sigma$=$0.15$, $p\approx0$) or when they had previous exposure to AVs ($\hat{\beta}$=$0.33$, $\sigma$=$0.05$, $p\approx0$), but trust remained largely unchanged by human explanations ($\hat{\beta}$=$0.12$, $\sigma$=$0.15$, $p$=$0.85$).
Trust ratings were not significantly affected by whether CEMA's explanations were highlighted ($\hat{\beta}$=$0.02$, $\sigma$=$0.05$, $p=0.66$) and there were no significant interaction effects between the average ratings of explanations and highlighting ($\hat{\beta}$=$-0.04$, $\sigma$=$0.04$, $p=0.36$).
The estimated trust levels varied across the 9 trust scales ($SD$=$0.53$) but not the observed tendencies.
Our results suggest that people who had some exposure to AVs or had a preference for CEMA's explanations were more likely to trust AVs in general, regardless of whether they knew which explanations came from CEMA.
Taken together with the result for \textbf{H1}, this suggests that CEMA's explanations may be more effective at improving people's trust in AVs than non-expert human explanations.

\section{Discussion and Future Work}\label{sec:discussion}

Our primary goal with CEMA is to advance the field of social XAI applied to dynamic multi-agent systems. 
A crucial component of intelligible explanations is the use of semantically meaningful features~\cite{dazeleyLevelsExplainableArtificial2021}.
Importantly, the challenge of designing useful features is not unique to CEMA but is a necessary step for any automated explanation generation system in social XAI.
With CEMA, we assumed that there is a feature function $\phi$ which performs the translation from the raw representations of state and action spaces to the more abstract semantic feature space.
This translation from state to feature space is domain-dependent and should be considered a crucial step during the deployment of social XAI systems.
However, CEMA is feature-agnostic so that counterfactual causal selection does not depend on $\phi$ or the interpretations of features.

CEMA also does not rely on a fixed causal graph to model dynamic multi-agent systems.
Instead, it assumes that there is a probabilistic model, such as a stochastic planner, trained joint policy, or autoregressive model trained on observational data, which can be used to forward simulate the state of the system.
Based on the work of~\citet{quillienCounterfactualsLogicCausal2023} and the counterfactual model of causation~\cite{hiltonLogicCausalAttribution1988,lewisCausation1973}, CEMA can derive causes to an ego agent's actions in any system where such a model is obtainable.
The assumption here is that these models cover alternatives that are grounded in factual observations with a non-zero probability, and any reasonably expressive algorithm would fulfill these criteria.

The user study suggests that people may prefer explanations generated by CEMA, however, trust levels are still low. 
This may be -- as several participants indicated in their feedback -- because people prefer to see agents act more conservatively, without exploiting potentially riskier but more efficient actions. %
Explanations that justify efficient but less safe decisions then have to overcome the inherent wariness of people, which was indeed high among participants, though it somewhat decreased after seeing explanations.

We designed CEMA to be used in conversations with users, but we did not focus on natural language processing in this work.
For example, we assume that queries unambiguously describe the timing of actions -- allowing us to focus on causal selection -- but actual natural language queries are fuzzy and imprecise.
By building modern NLP components, we can strengthen the social and conversational aspects of CEMA.
Future work will involve the integration of language parsing~\cite{liSurveyDiscourseParsing2022} and dialogue systems~\cite{chenSurveyDialogueSystems2017} leveraging modern neural language models to deliver explanations.

Our implementation of CEMA for AD improves on existing social XAI methods for AD in several aspects.
In contrast to \citet{omeizaAccountabilityProvidingIntelligible2021}, we avoid using a surrogate model and generate causal explanations that take the temporal nature of driving into account.
Compared to \citet{gyevnarHumanCentricMethodGenerating2022}, CEMA supports multiple modes of explanations with both high-level and low-level features.

To conclude, our goal is to address some of the transparency-related social concerns of AI.
CEMA fills a gap in social XAI by enabling causal explanation generation in dynamic sequential multi-agent systems.
As we expect to see autonomous agents proliferate in everyday environments, social explanations will be crucial for building user trust and for the acceptance of new technologies.

\begin{acks}
The authors thank C. Brewitt, M. Tamborski, and Y. Ziser for their helpful comments on earlier drafts.
This work was supported in part by the UKRI Centre for Doctoral Training in Natural Language Processing (grant EP/S022481/1) and the Edinburgh Laboratory for Integrated Artificial Intelligence.
\end{acks}

\bibliographystyle{ACM-Reference-Format} 
\bibliography{refs}

\appendix

\section{Implementation Details}

\subsection{User Queries}\label{sec:queries}

As we our focus is on causal selection and NLP, we do not use NLU systems to parse users' questions.
Instead, we define a standard query template and code queries as a structured JSON file.
The template is given in \cref{tab:json} with each field and the corresponding type.
The queries are given in terms of three main components: a query type (Type), a list of actions (Actions), and tense of the question (Tense); with an optional list of factual actions (Factuals).
A list of actions is given as a combination of maneuvers and macro actions of IGP2.
We identify for each timestep of a trajectory which maneuver and macro action are executed allowing for a timestep-by-timestep comparison to the queried list of actions.

The above information is used to infer the start $u$ and end timestep $v$ necessary to determine the state sequence $\hat{s}_{u:v}$.
First, we use tense to filter out those timesteps of the observed trajectory $s^\varepsilon_{1:t}$ that do not match the tense of the question. 
For example, if the question is in future tense, then all past and present timesteps are removed.
We then \textit{group} consecutive timesteps by actions and filter out those action groups that do not match the user-given list of actions.
If after this process, we still end up with multiple groups, then the first timestep of the group closest to the current timestep $t$ is picked as $u$ for the explanation.
The end timestep $v$ is then given by the last timestep of the action group to which $u$ belongs.

\begin{table}[b]
    \centering
    \caption{The format template to encode queries. Types marked with a star are optional.}
    \label{tab:json}
    \begin{tabular}{lll}
    \toprule
         \textbf{Field} & \textbf{Type}  & \textbf{Explanation}\\
         \midrule
         Type        & \textit{str}        & Type of the user's question. \\
         VID (VID)   & \textit{int}        & The VIDentifier for $\varepsilon$. \\
         Tense       & \textit{str}        & Grammatical tense of the \\
         {}          & {}                  & user's question \\ 
         {}          & {}                  &(past, present, future). \\
         Actions     & List[\textit{str}]  & A list of actions the user \\ 
         {}          & {}                  & is interested in. \\
         Query Time  & \textit{int}        & Timestep of the user \\ 
         {}          & {}                  & asking the question. \\
         Action Time & \textit{int}*       & Start time of queried actions. \\
         {}          & {}                  & Only given for \textit{what} \\ 
         {}          & {}                  &query type. \\
         Negated     & \textit{bool}*      & Whether the user's \\ 
         {}          & {}                  & question is negated. \\
         Factuals    & List[\textit{str}]* & The factual actions of $\varepsilon$. \\
         {}          & {}                  & Only used if Actions is \\ 
         {}          & {}                  & counterfactual. \\
    \bottomrule
    \end{tabular}
\end{table}

\subsection{Query Types}
We also define three query types: \textit{what}, \textit{what if}, and \textit{why} queries, which also affect the selected state sequence $\hat{s}_{u:v}$.

A \textit{why}-query asks why a list of factual actions is executed.
For example, \textit{``Why did you change lanes to the left?''}
It follows the process described above to determine $\hat{s}_{u:v}$.
We concurrently generate both teleological and mechanistic explanations for this query type.

A \textit{what if}-query asks about the actions of $\varepsilon$ had some other vehicle $i$ executed a different counterfactual list of actions.
For example, \textit{``What if vehicle 1 had stopped (instead of going straight)?''}
Therefore, \textit{what if}-queries encode a contrastive user question.
We assume that the (counterfactual) list of actions directly replaces some factual list of actions that is also given for vehicle $i$ as part of the user's query.
However, we cannot directly apply the process in \cref{sec:queries} to this list of actions to determine $\hat{s}_{u:v}$ as these actions have never been observed. 
Instead, we sample a set of alternative trajectories from the generative model $p(\hat{S}_{t+1:n}|s_{1:t})$ and find the counterfactual trajectory where vehicle $i$ was executing the counterfactual list of actions.
We apply the process in \cref{sec:queries} to the selected counterfactual trajectory which filters out all action groups that do not correspond to the counterfactual list of actions.
If multiple action groups remain, then we look at the additional factual list of actions from the query
We use the observed trajectory and the factual list of actions to then find the factual action group of greatest overlap with any remaining counterfactual action groups.
The values $u$ and $v$ are selected as the start and end timesteps of this overlapping region.
This process guarantees that $u$ and $v$ happen at times when both the counterfactual and factual list of actions were executed in some possible worlds.
We use the term \textit{associative explanation} to refer to an explanation that describes the alternative actions of vehicle $\varepsilon$ had vehicle $i$ executed the counterfactual list of actions.
We generate teleological and mechanistic explanations for why vehicle $\varepsilon$ would have executed those actions.

A \textit{what}-query asks what vehicle $\varepsilon$ is doing at some given timestep.
Thus, it always results in an associative explanation.
Here, we do not have access to the list of actions since that is what we are trying to determine. 
Instead, we assume that the timestep $u$ of the start of the action (Action Time) is given.

Users' questions can be negated (Negated) sentences.
The effect of negation is flipping the value of the boolean outcome variable $y$ in Algorithm 1, but it also affects how we determine $u$ and $v$.
Negation turns a \textit{why}-question into a \textit{why not} question, as in \textit{``Why did you not turn right?''} 
We treat this as a \textit{what if} query asking about a counterfactual list of actions about vehicle $\varepsilon$ itself.
For \textit{what if} queries, negation means that a factual list of actions is already given, as in \textit{``What if you hadn't stopped?''}
Therefore, we can directly apply the process in \cref{sec:queries} to determine $u$ and $v$.

Finally, the user can query the future (e.g., future ego plan can be shown on a screen).
For this, we concatenate the observed joint states $s_{1:t}$ with the \textit{maximum a posteriori}-predictions of $p(\hat{S}_{t+1:n}|s_{1:t})$ giving $\hat{s}_{1:n}$.
If the queried sequence is hypothetical, i.e., $\hat{s}_{u:v} \nprec \hat{s}_{1:n}$, then we assume that we are given a corresponding factual subsequence to allow for the inference of the timings $u$ and $v$.

\subsection{Additive Smoothing}
Since IGP2 can assign to some (reachable) goals and trajectories near-zero probabilities, we use additive smoothing with parameter $\alpha$ to make sure every trajectory can get sampled from $p(\hat{S}_{t+1:n}|s_{1:t})$.
Given a discrete probability distribution $p_\theta:\Omega \mapsto [0,1]$ parametrised by $\theta=[\theta_1,\dots,\theta_d]$ where $\Omega$ is a finite non-empty set of events with size $d$, additive smoothing creates a new discrete probability distribution $p_{\phi}:\Omega \mapsto [0,1]$ with new parameter vector $\phi=[\phi_1,\dots,\phi_d]$ defined as: 
\[
    \phi_i = \frac{\theta_i + \alpha}{1 + d \alpha}, \quad (i=1,\dots,d),
\]
where we assumed that $\sum_i \theta_i = 1$.

\section{Experimental Setup}
We implement CEMA using Python, building on the publicly available code repository of IGP2.\footnote{IGP2 available at \url{https://uoe-agents.github.io/IGP2/}.}
The source code for CEMA is available as supplementary material, is well-documented, and contains detailed instructions on how to reproduce our results.
Our experiments were run on a modern Windows 10 PC with 32 GiB of RAM and a 12-core CPU. 
CEMA does not need a GPU.

Each scenario is defined using a semantic road layout in the ASAM OpenDrive 1.6 format, and using a configuration file that describes the behaviour and starting regions of agents on the road. 
Scenarios are executed in a simple, discrete-time simulation environment with an execution rate of 20 frames per second.
For reproducing our results exactly, a random seed of 21 should be used whenever the algorithm uses randomness.
Note, the results were confirmed with multiple seeds, we only fixed the seed for presentation in the paper.

Our scenarios are usually only a few seconds long (in simulation time), therefore, to avoid finding irrelevant causes and to speed up our implementation, we remove all states that are more than 5 seconds away from the current time step $t$ of the simulation.
For all scenarios, we sample $K=100$ counterfactual sequences with a smoothing weight $\alpha=0.1$, and limits of $\tau_{min}=2$ and $\tau_{max}=5$ seconds. 
We picked $K$ so that it is large enough that a diverse range of trajectories is sampled considering that IGP2 predicts up to 3 distinct trajectories for each agent.
The value for $\tau_{min}$ is the planning period of IGP2 (i.e., the number of seconds between two calls to IGP2), while $\tau_{max}$ is the maximum temporal distance as described above.
We do not rely on any external datasets and no pre-processing steps are needed to run our code.

\begin{figure}
    \centering
    \includegraphics[width=\linewidth]{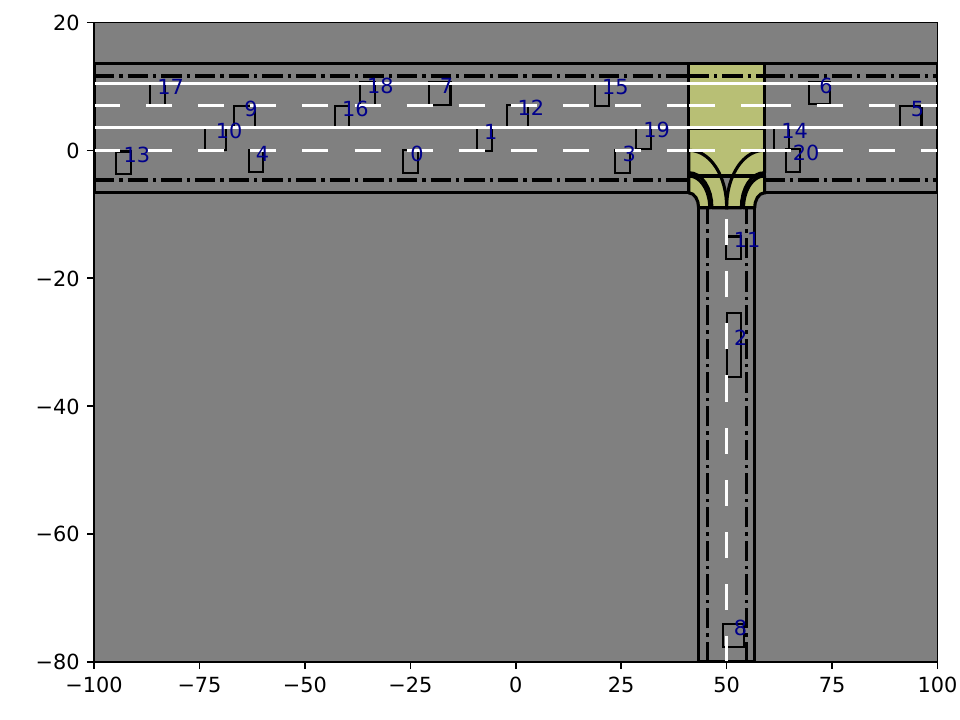}
    \caption{Extended S1 with 20 spawn locations for agents.}
    \label{fig:map_5}
\end{figure}

\begin{figure}
    \centering
    \includegraphics[width=\linewidth]{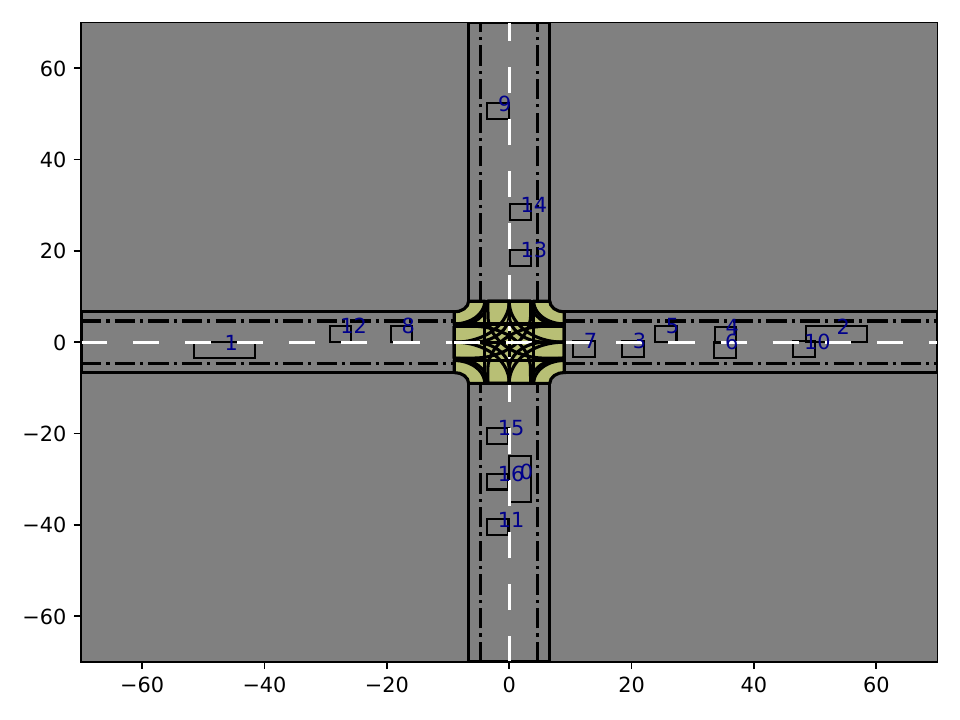}
    \caption{S2 with 16 spawn locations for agents.}
    \label{fig:map_6}
\end{figure}

\begin{figure}
    \centering
    \includegraphics[width=\linewidth]{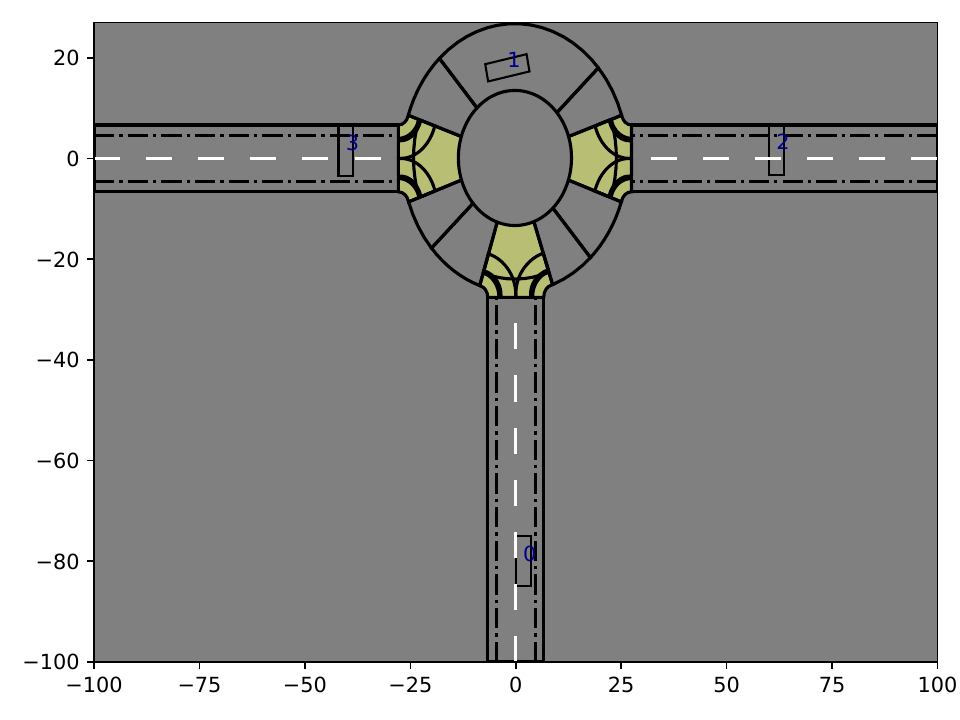}
    \caption{S3 with 4 spawn locations for agents.}
    \label{fig:map_7}
\end{figure}

\begin{figure}
    \centering
    \includegraphics[width=\linewidth]{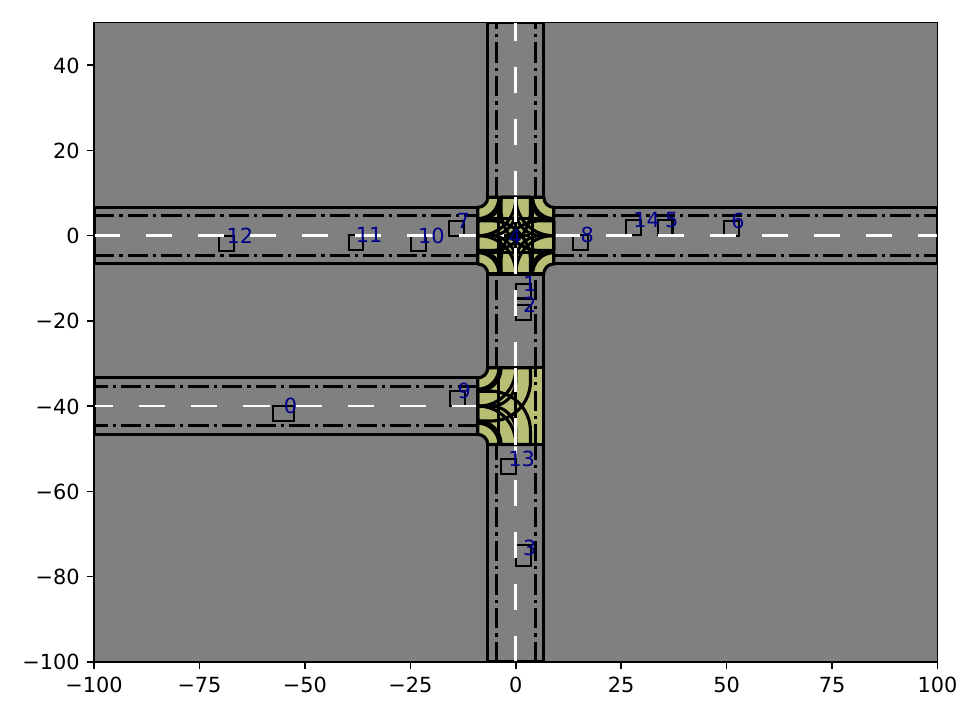}
    \caption{S4 with 16 spawn locations for agents.}
    \label{fig:map_8}
\end{figure}

\subsection{Scaling with the Number of Agents}\label{apx:scaling}
CEMA can scale to a large number of agents, which means that it can identify the relevant causes behind the actions of the ego agent, even if there are numerous irrelevant agents around, which did not have a causal effect on the actions of the ego.
To show that this is true, we increase the number of agents in all four scenarios while making sure that the ego executes the same actions
We then show that the same relevant causes, both teleological and mechanistic, are found by CEMA behind the ego's actions. 

We alter the number of total agents in each scenario as follows: S1: 20 agents; S2: 16 agents; S3: 4 agents; S4: 16 agents.
We only increase S3 to 5 agents, as the scenario map was too compact to allow for more agents without a significant change to the behaviour of the ego.
\Cref{fig:map_5,fig:map_6,fig:map_7,fig:map_8} show the spawn regions of each agent, in which they are placed randomly.

\subsection{Size and Sampling Robustness}
We test the size robustness of mechanistic explanations by sampling increasingly larger datasets using Algorithm 1 and for each dataset we rerun Algorithm 2 to determine causal attributions.
We randomly sample $K \in \{5,10,\dots,100\}$ different sample sizes.
For a given $K$, we repeat the sampling process 50 times.
We plot the mean and 95\% bootstrapped confidence intervals of the causal attributions against $K$.

Using the fact that $\lim_{\alpha \rightarrow \infty} \phi_i = \frac{1}{d}$ for all $i$, we see that in the limit as $\alpha$ approaches infinity, $p_\phi$ approaches a uniform distribution with probability $\frac{1}{d}$ for all elements in $\Omega$.
Thus we can interpolate between the original $p_\theta$ and an approximately uniform distribution by setting $\alpha$ to larger and larger values.

Using the above, we perform our experiment for testing the robustness of our system against increasing uncertainty in external predictions of other vehicles' goals and trajectories.
We define a list of twenty $\alpha$ values spaced on a logarithmic scale plus zero giving in total 21 distinct alphas (\cref{tab:alphas}).
For each $\alpha$ in this list, we sample a dataset of size $K=50$ and determine causal attributions for mechanistic causes.

\begin{table}
    \centering
    \caption{The list of $\alpha$ values used for interpolation in the sampling robustness experiments.}
    \label{tab:alphas}       
    \begin{tabular}{p{8.1cm}}
        \toprule
         $0, 0.1,  0.14,  0.18,  0.25,  0.34,  0.45,  0.62,  0.83,  1.13,  1.53,$ \\
         $2.07,  2.8,  3.79,  5.13,  6.95,  9.41, 12.74, 17.25, 23.36, 31.62$ \\
        \bottomrule
    \end{tabular}
\end{table}

\section{Computational Evaluation}

\begin{figure*}
    \begin{subfigure}[b]{\textwidth}
        \centering
        \caption{Corresponding questions: \textit{``Why will you change lanes?''}; \textit{``Why does it decrease the time to the goal?''}; \textit{``Why will it be slower?''}}
        \begin{tabular}{l|llllllll}
        \toprule
             \textbf{Conv.} & \textbf{Type} & \textbf{VID} & \textbf{Tense} & \textbf{Actions} & \textbf{Query Time} & \textbf{Action Time} & \textbf{Negated} & \textbf{Factuals}  \\
            \midrule
             S1-A & Why & 0 (AV) & Future & Change left & 40 & --- & No & --- \\
             \bottomrule
        \end{tabular}
    \end{subfigure}    
    \vspace{0em}\\    
    \begin{subfigure}[b]{\textwidth}
        \centering
        \includegraphics[width=\textwidth]{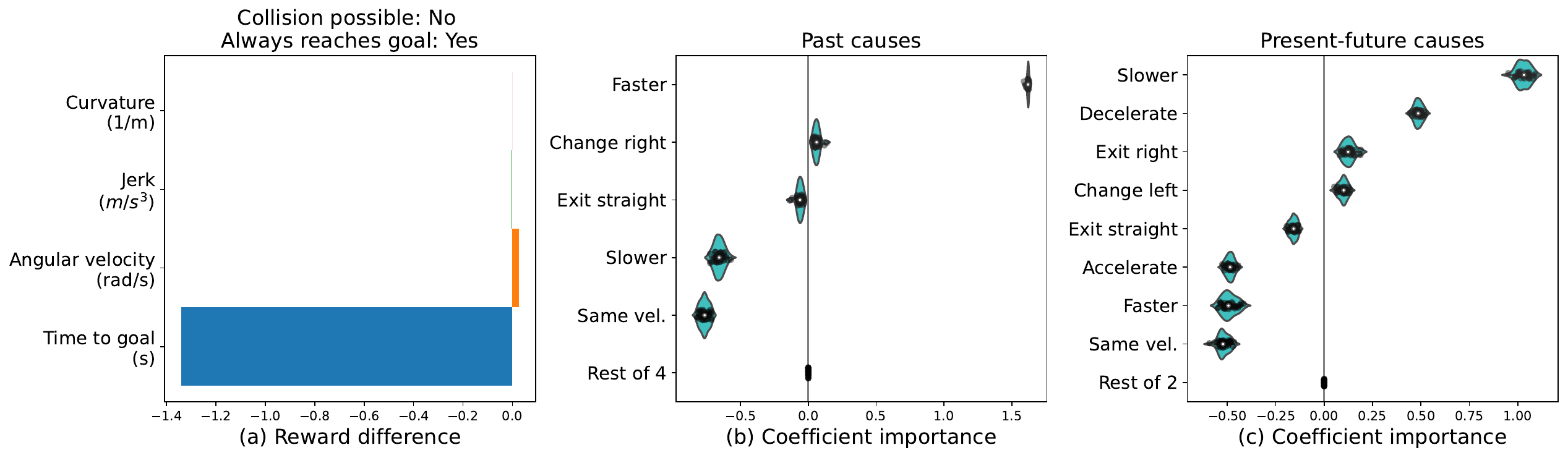}
    \end{subfigure}
    \vspace{0em}\\
    \begin{subfigure}[b]{0.49\textwidth}
        \centering
        \includegraphics[width=\textwidth]{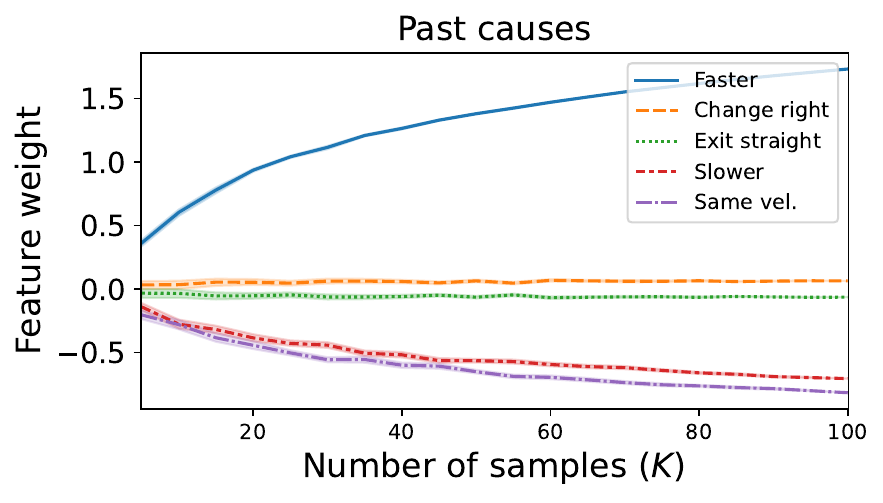}
    \end{subfigure}
    \hfill
    \begin{subfigure}[b]{0.49\textwidth}
        \centering
        \includegraphics[width=\textwidth]{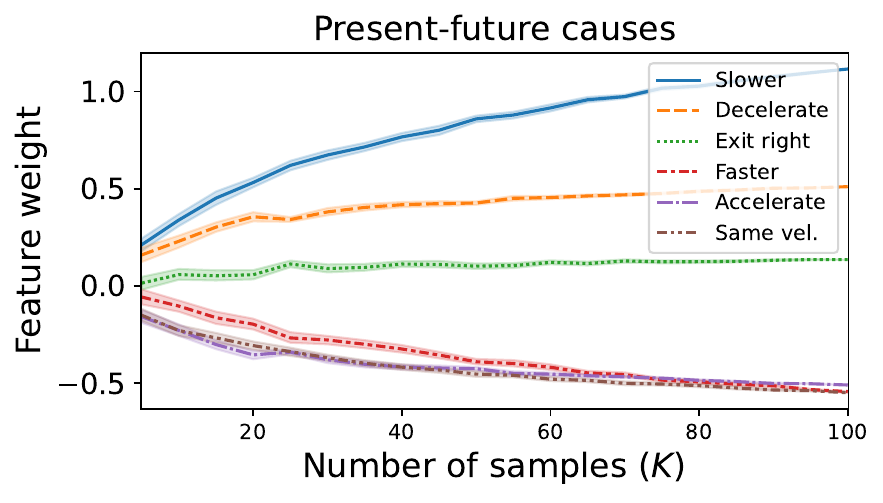}
    \end{subfigure}
    \vspace{1em}\\
    \begin{subfigure}[b]{0.49\textwidth}
        \centering
        \includegraphics[width=\textwidth]{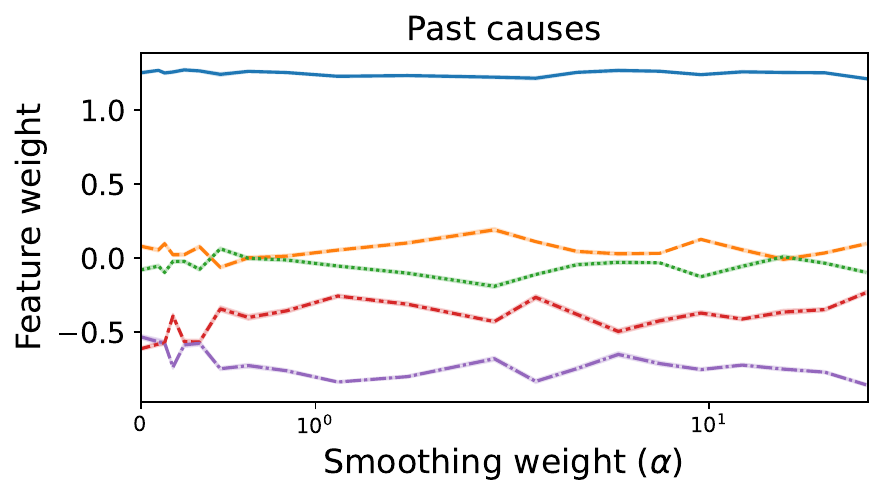}
    \end{subfigure}
    \hfill
    \begin{subfigure}[b]{0.49\textwidth}
        \centering
        \includegraphics[width=\textwidth]{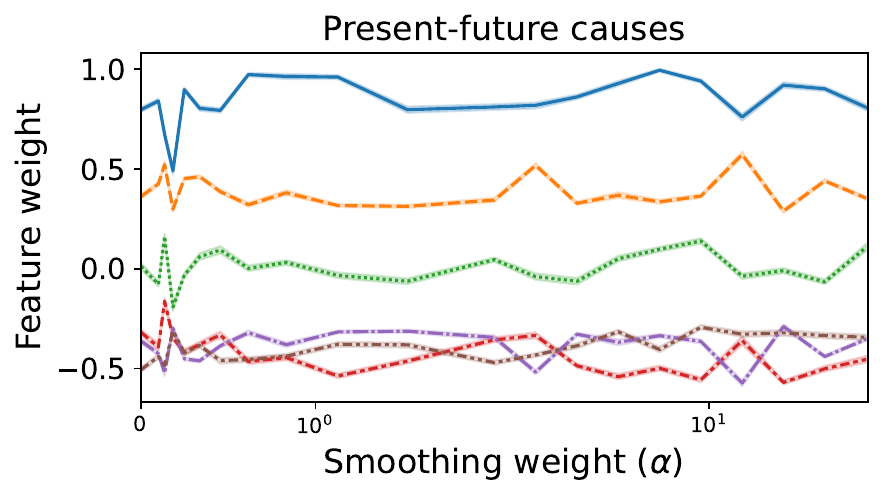}
    \end{subfigure}       
    \caption{Results for the first three questions in conversation S1-A.}
    \label{fig:s1-1}
\end{figure*}

\begin{figure*}
    \centering
    \caption{Corresponding question: \textit{``What if vehicle 1 hadn't changed lanes to the right before?''}}
    \begin{subfigure}[b]{\textwidth}
        \centering
        \begin{tabular}{l|llllllll}
        \toprule
             \textbf{Conv.} & \textbf{Type} & \textbf{VID} & \textbf{Tense} & \textbf{Actions} & \textbf{Query Time} & \textbf{Action Time} & \textbf{Negated} & \textbf{Factuals}  \\
            \midrule
             S1-A & What if & 1 & Past & Change right & 75 & --- & Yes & --- \\
             \bottomrule
        \end{tabular}
    \end{subfigure}    
    \vspace{0em}\\    
    \begin{subfigure}[b]{\textwidth}
        \centering
        \includegraphics[width=\textwidth]{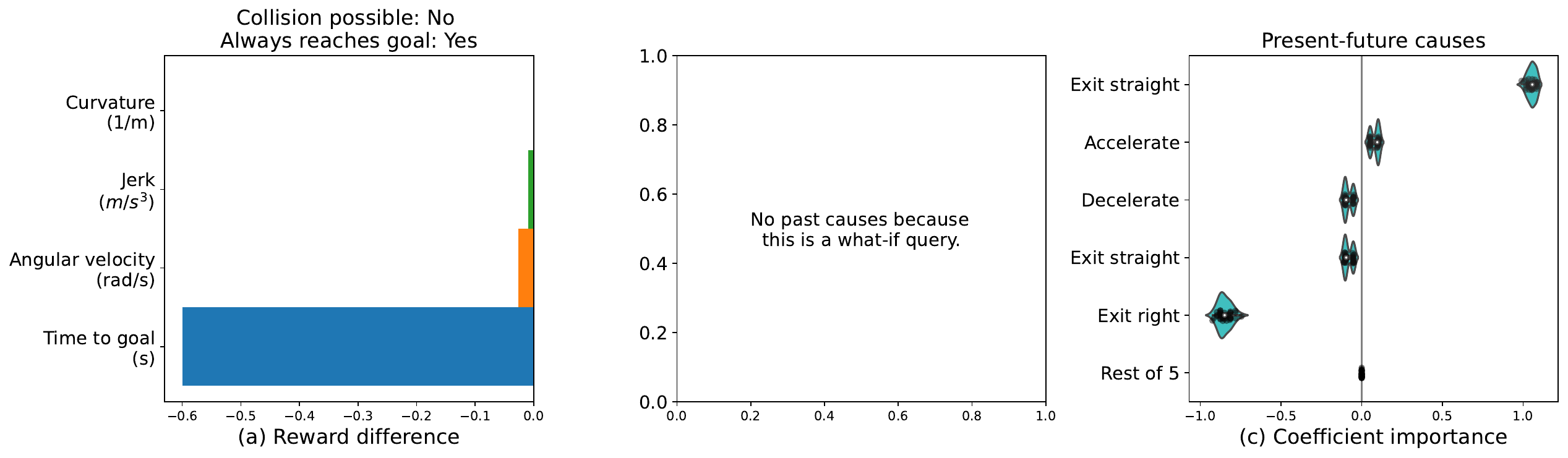}
    \end{subfigure}
    \vspace{0em}\\
    \begin{subfigure}[b]{0.5\textwidth}
        \centering
    \end{subfigure}
    \hfill
    \begin{subfigure}[b]{0.5\textwidth}
        \centering
        \includegraphics[width=\textwidth]{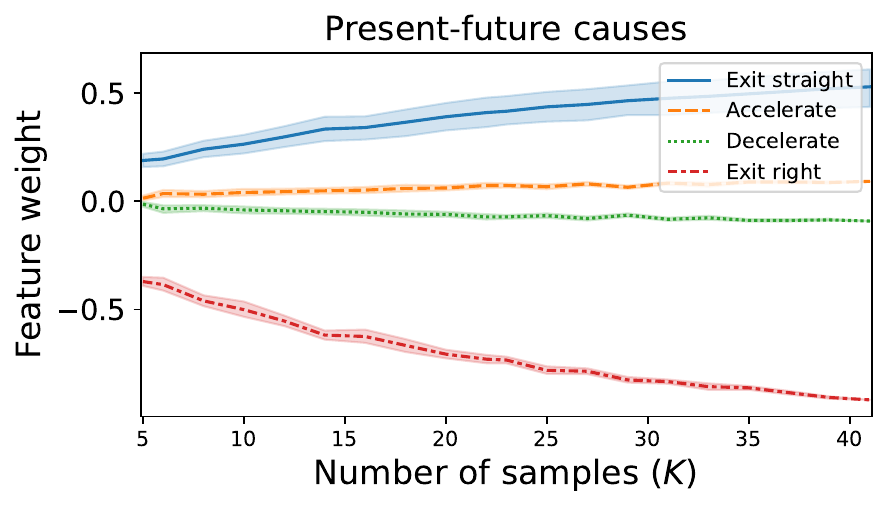}
    \end{subfigure}
    \hfill
    \vspace{1em}\\
    \begin{subfigure}[b]{0.5\textwidth}
        \centering
    \end{subfigure}
    \hfill
    \begin{subfigure}[b]{0.5\textwidth}
        \centering
        \includegraphics[width=\textwidth]{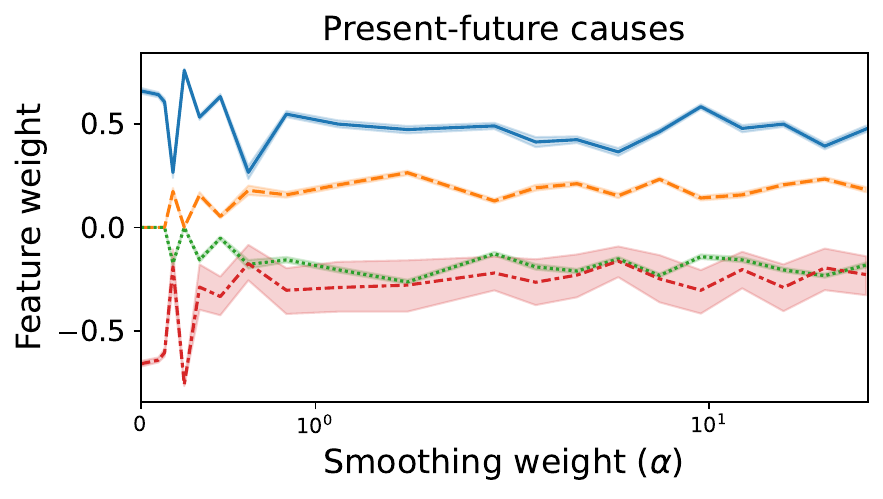}
    \end{subfigure}       
    \hfill
    \vspace{0em}\\
    \caption{Results for the last question in conversation S1-A.}
    \label{fig:s1-2}
\end{figure*}

\begin{figure*}
    \begin{subfigure}[b]{\textwidth}
        \centering
        \caption{Corresponding questions: \textit{``Why are you not going straight?''}, \textit{``Why is it slower to go straight then?''}, \textit{``Any other reasons?''}}        
        \begin{tabular}{l|llllllll}
        \toprule
             \textbf{Conv.} & \textbf{Type} & \textbf{VID} & \textbf{Tense} & \textbf{Actions} & \textbf{Query Time} & \textbf{Action Time} & \textbf{Negated} & \textbf{Factuals}  \\
            \midrule
             S1-B & Why & 0 & Present & Go straight & 45 & --- & Yes & Change left \\
             \bottomrule
        \end{tabular}
    \end{subfigure}    
    \vspace{0em}\\    
    \begin{subfigure}[b]{\textwidth}
        \centering
        \includegraphics[width=\textwidth]{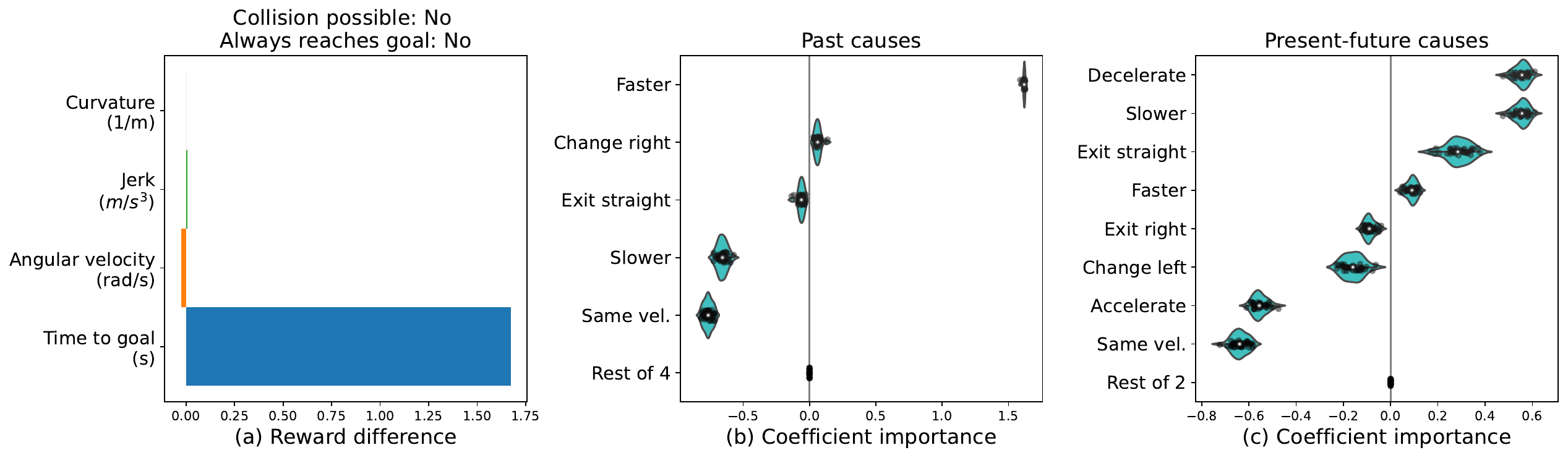}
    \end{subfigure}
    \vspace{0em}\\
    \begin{subfigure}[b]{0.49\textwidth}
        \centering
        \includegraphics[width=\textwidth]{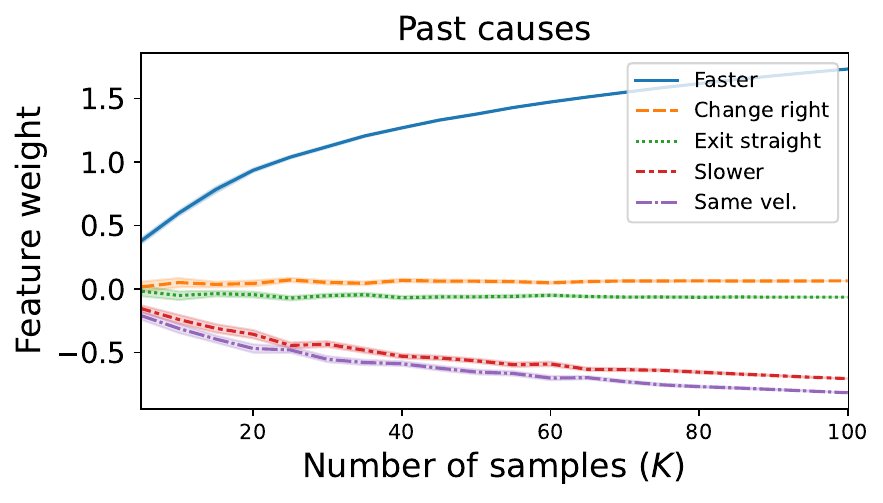}
    \end{subfigure}
    \hfill
    \begin{subfigure}[b]{0.49\textwidth}
        \centering
        \includegraphics[width=\textwidth]{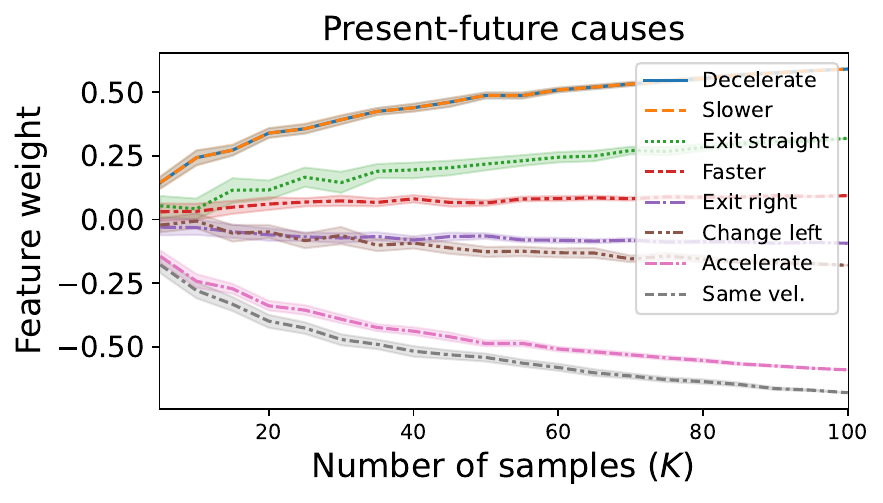}
    \end{subfigure}
    \vspace{1em}\\
    \begin{subfigure}[b]{0.49\textwidth}
        \centering
        \includegraphics[width=\textwidth]{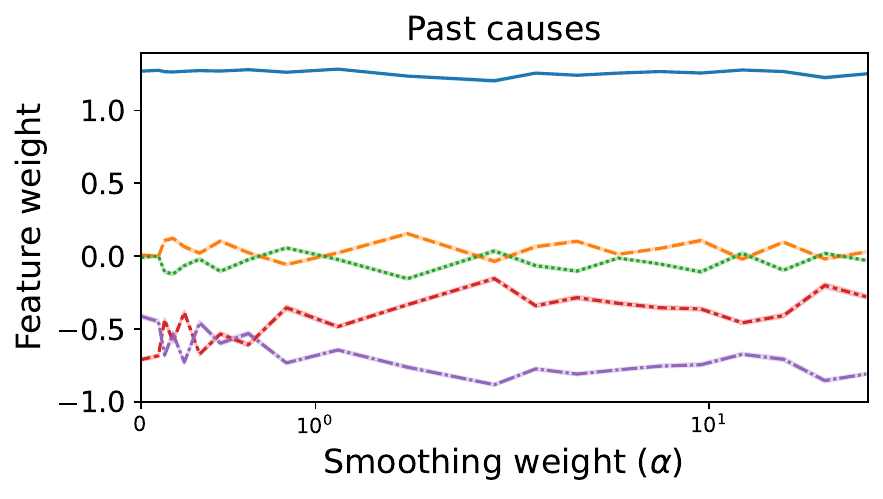}
    \end{subfigure}
    \hfill
    \begin{subfigure}[b]{0.49\textwidth}
        \centering
        \includegraphics[width=\textwidth]{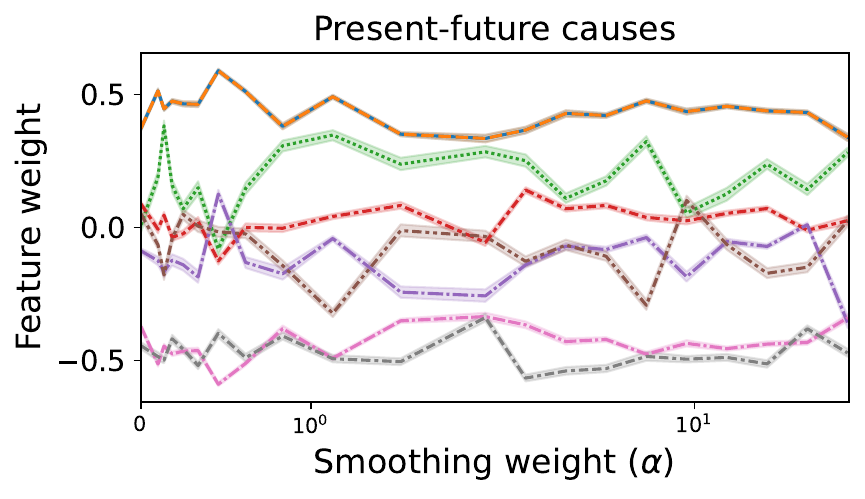}
    \end{subfigure}       
    \caption{Results for the first three questions in conversation S1-B.}
    \label{fig:s1-3}
\end{figure*}

\begin{figure*}
    \begin{subfigure}[b]{\textwidth}
        \centering
        \caption{Corresponding questions: \textit{``Why didn't you stop to give way?''}, \textit{``Couldn't it have caused a collision?''}, \textit{``How did you know?''}}        
        \begin{tabular}{l|llllllll}
        \toprule
             \textbf{Conv.} & \textbf{Type} & \textbf{VID} & \textbf{Tense} & \textbf{Actions} & \textbf{Query Time} & \textbf{Action Time} & \textbf{Negated} & \textbf{Factuals}  \\
            \midrule
             S2 & Why & 0 & Past & Give way \& stop & 160 & --- & Yes & Give way \& accelerate \\
             \bottomrule
        \end{tabular}
    \end{subfigure}    
    \vspace{0em}\\    
    \begin{subfigure}[b]{\textwidth}
        \centering
        \includegraphics[width=\textwidth]{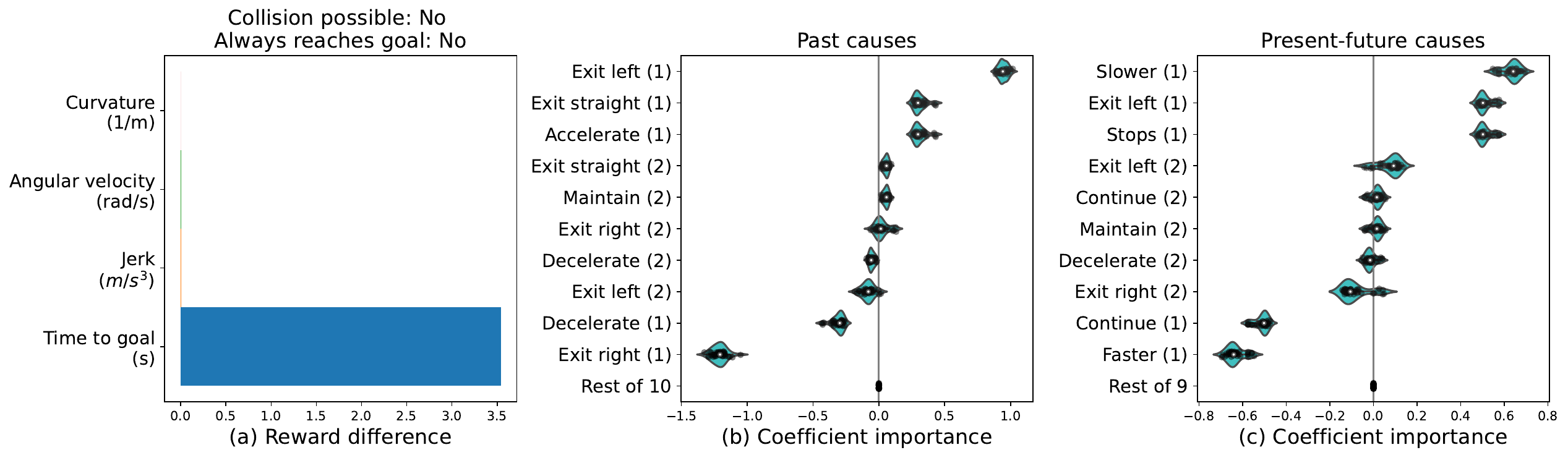}
    \end{subfigure}
    \vspace{0em}\\
    \begin{subfigure}[b]{0.49\textwidth}
        \centering
        \includegraphics[width=\textwidth]{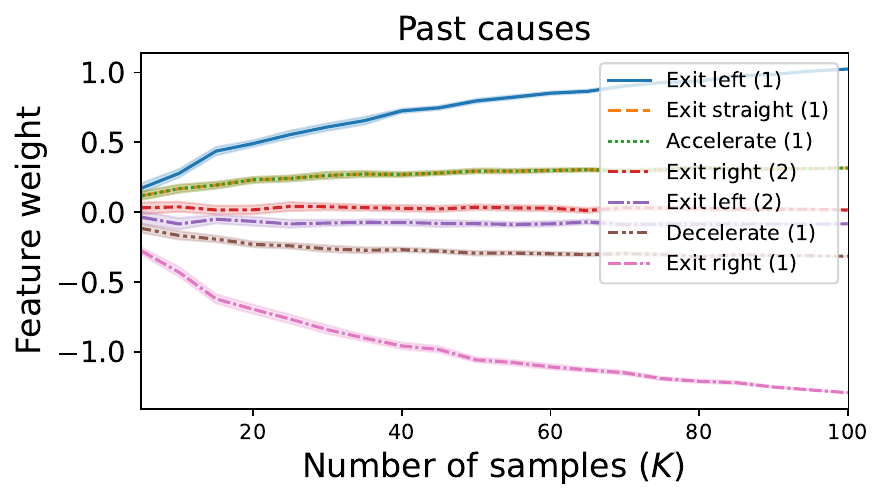}
    \end{subfigure}
    \hfill
    \begin{subfigure}[b]{0.49\textwidth}
        \centering
        \includegraphics[width=\textwidth]{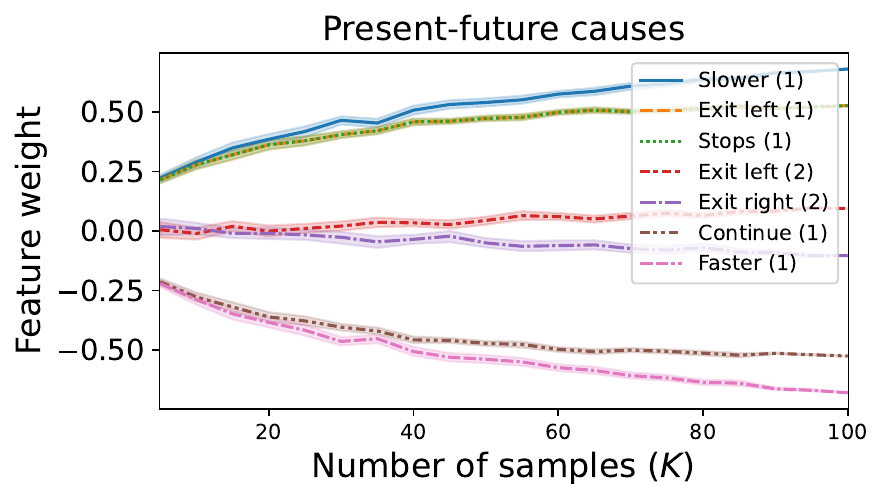}
    \end{subfigure}
    \vspace{1em}\\
    \begin{subfigure}[b]{0.49\textwidth}
        \centering
        \includegraphics[width=\textwidth]{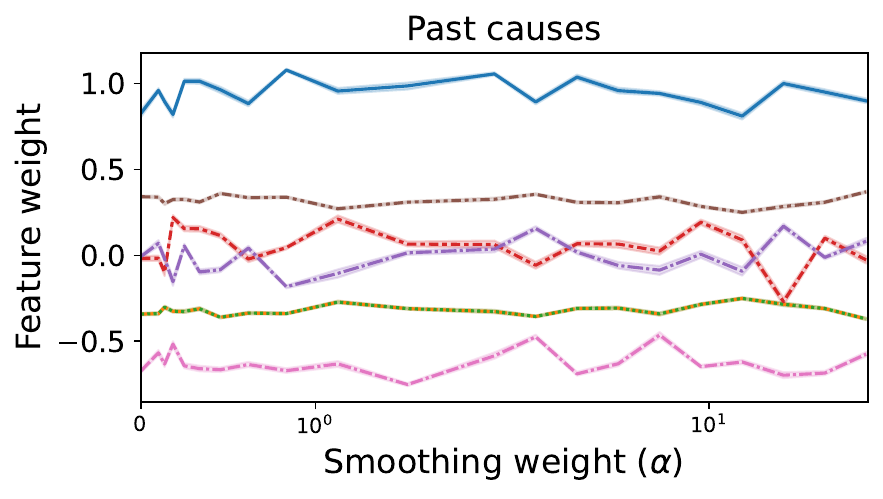}
    \end{subfigure}
    \hfill
    \begin{subfigure}[b]{0.49\textwidth}
        \centering
        \includegraphics[width=\textwidth]{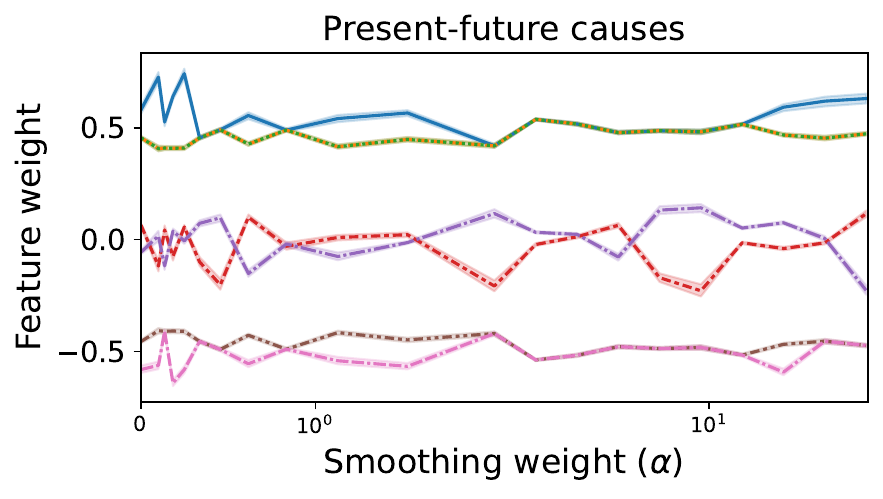}
    \end{subfigure}       
    \caption{Results for the first three questions in conversation S2.}
    \label{fig:s2-1}
\end{figure*}

\begin{figure*}
    \begin{subfigure}[b]{\textwidth}
        \centering
        \caption{Corresponding question: \textit{``What if it went straight instead?''}}
        \begin{tabular}{l|llllllll}
        \toprule
             \textbf{Conv.} & \textbf{Type} & \textbf{VID} & \textbf{Tense} & \textbf{Actions} & \textbf{Query Time} & \textbf{Action Time} & \textbf{Negated} & \textbf{Factuals}  \\
            \midrule
             S2 & What if & 1 & Past & Go straight & 110 & --- & No & Turn left \\
             \bottomrule
        \end{tabular}
    \end{subfigure}    
    \vspace{0em}\\    
    \begin{subfigure}[b]{\textwidth}
        \centering
        \includegraphics[width=\textwidth]{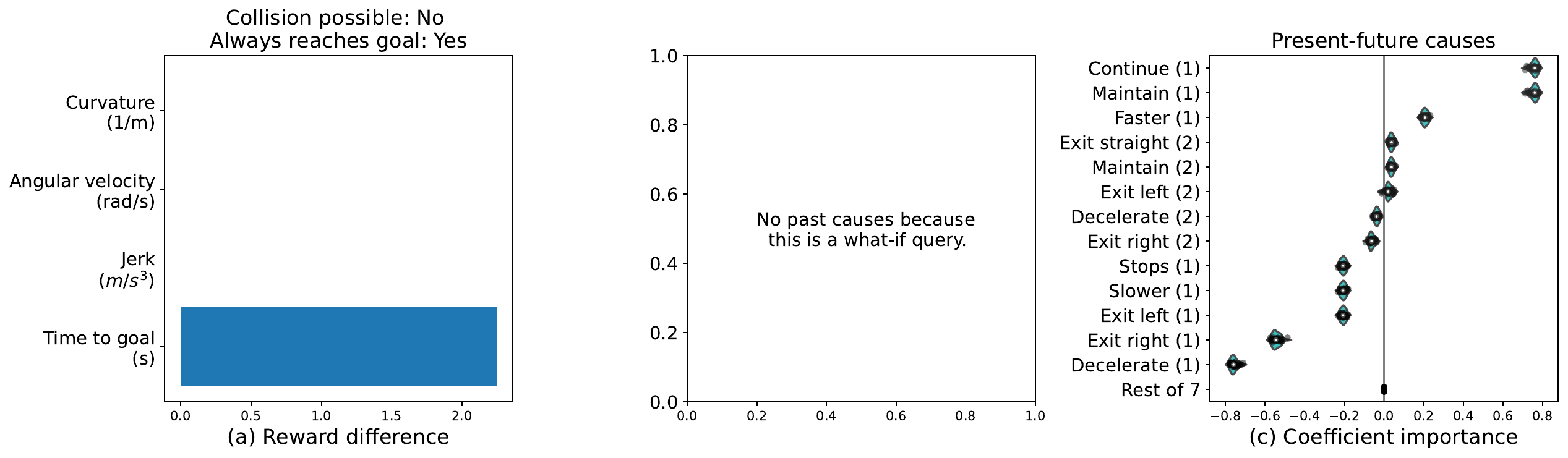}
    \end{subfigure}
    \vspace{0em}\\
    \begin{subfigure}[b]{0.5\textwidth}
        \centering
    \end{subfigure}
    \hfill
    \begin{subfigure}[b]{0.5\textwidth}
        \centering
        \includegraphics[width=\textwidth]{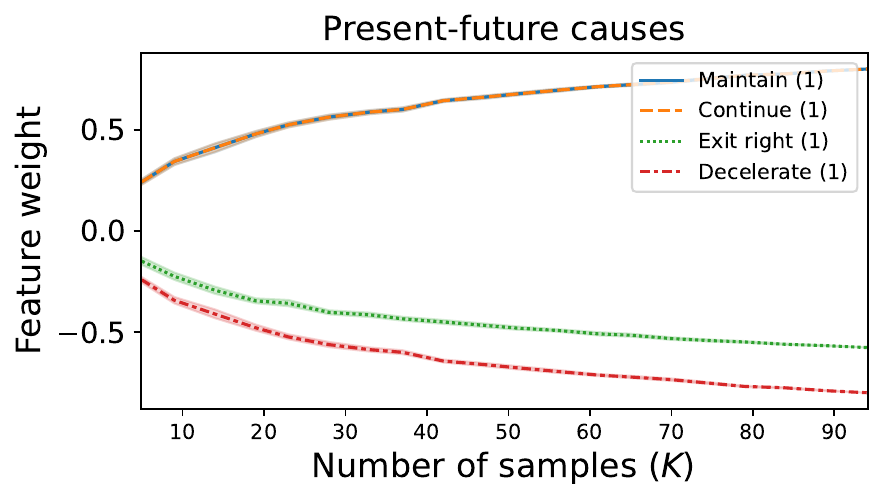}
    \end{subfigure}
    \hfill
    \vspace{1em}\\
    \begin{subfigure}[b]{0.5\textwidth}
        \centering
    \end{subfigure}
    \hfill
    \begin{subfigure}[b]{0.5\textwidth}
        \centering
        \includegraphics[width=\textwidth]{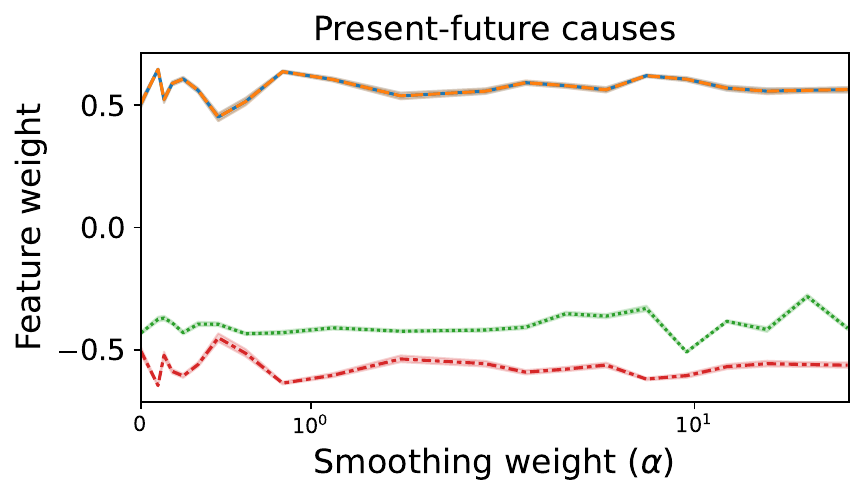}
    \end{subfigure}       
    \hfill
    \vspace{0em}\\    
    \caption{Results for the last question in conversation S2.}
    \label{fig:s2-2}
\end{figure*}

\begin{figure*}
    \begin{subfigure}[b]{\textwidth}
        \centering
        \caption{Corresponding questions: \textit{``Will you stop to give way?''}, \textit{``How do you know we won't collide with the oncoming car?''}}
        \begin{tabular}{l|llllllll}
        \toprule
             \textbf{Conv.} & \textbf{Type} & \textbf{VID} & \textbf{Tense} & \textbf{Actions} & \textbf{Query Time} & \textbf{Action Time} & \textbf{Negated} & \textbf{Factuals}  \\
            \midrule
             S3 & Why & 0 & Present & Give way \& stop & 90 & --- & Yes & Give way \& accelerate \\
             \bottomrule
        \end{tabular}
    \end{subfigure}    
    \vspace{0em}\\    
    \begin{subfigure}[b]{\textwidth}
        \centering
        \includegraphics[width=\textwidth]{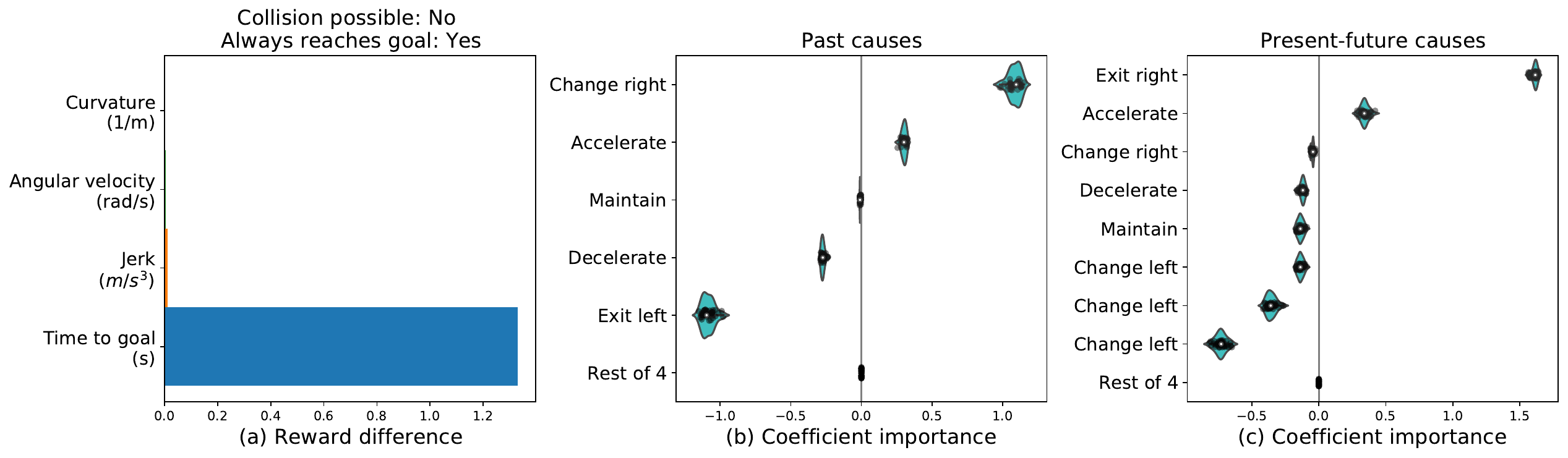}
    \end{subfigure}
    \vspace{0em}\\
    \begin{subfigure}[b]{0.49\textwidth}
        \centering
        \includegraphics[width=\textwidth]{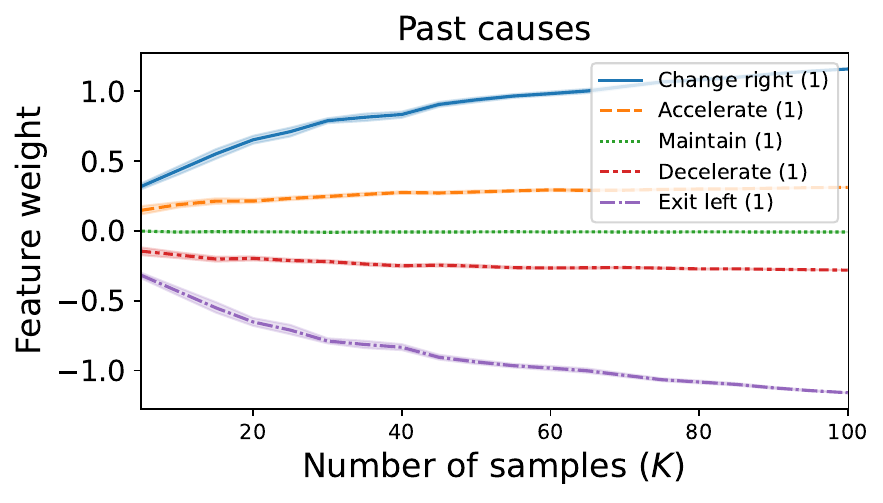}
    \end{subfigure}
    \hfill
    \begin{subfigure}[b]{0.49\textwidth}
        \centering
        \includegraphics[width=\textwidth]{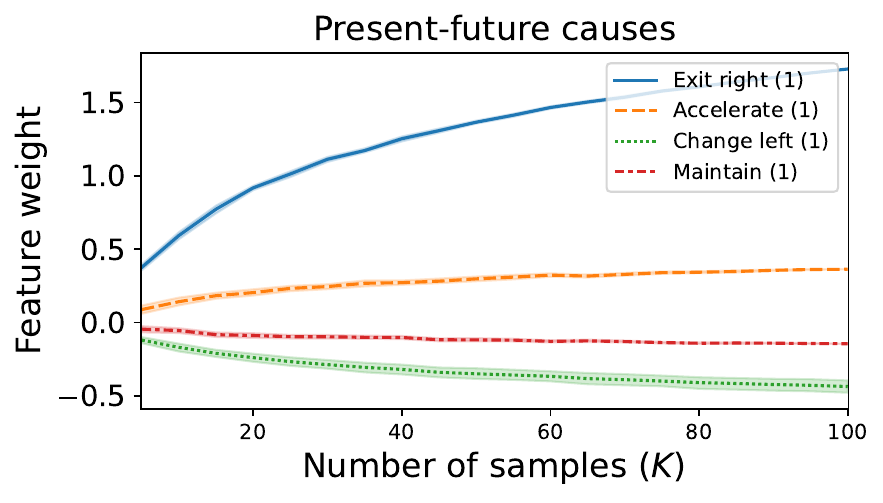}
    \end{subfigure}
    \vspace{1em}\\
    \begin{subfigure}[b]{0.49\textwidth}
        \centering
        \includegraphics[width=\textwidth]{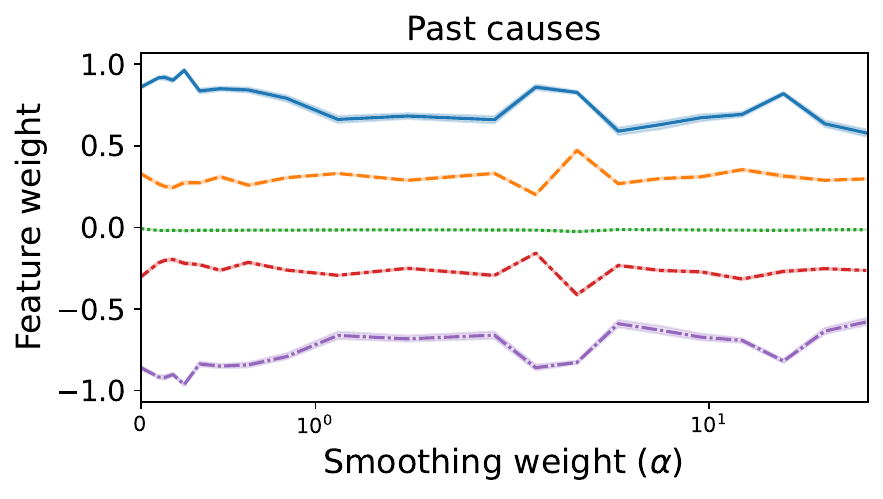}
    \end{subfigure}
    \hfill
    \begin{subfigure}[b]{0.49\textwidth}
        \centering
        \includegraphics[width=\textwidth]{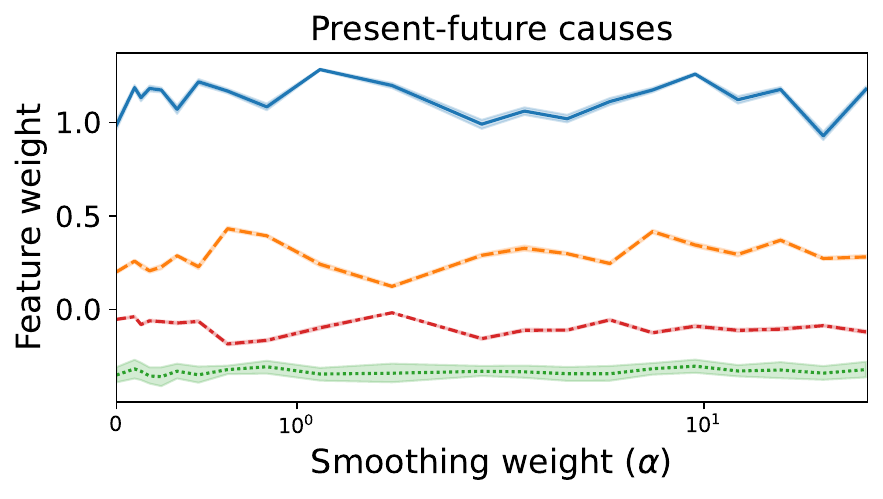}
    \end{subfigure}       
    \caption{Results for the last two questions in conversation S3. There are multiple ``change left'' among the present-future causes as the non-ego could change lanes left at many points in the roundabout. CEMA differentiates between all of them, we only show it here this way for brevity.}
    \label{fig:s3-1}
\end{figure*}

\begin{figure*}
    \begin{subfigure}[b]{\textwidth}
        \centering
        \caption{Corresponding questions: \textit{``Why are you not stopping to give way?''}, \textit{``Why is it safe to turn left?''}}
        \begin{tabular}{l|llllllll}
        \toprule
             \textbf{Conv.} & \textbf{Type} & \textbf{VID} & \textbf{Tense} & \textbf{Actions} & \textbf{Query Time} & \textbf{Action Time} & \textbf{Negated} & \textbf{Factuals}  \\
            \midrule
             S4 & Why & 0 & Present & Give way \& stop & 140 & --- & Yes & Turn left \& accelerate \\
             \bottomrule
        \end{tabular}
    \end{subfigure}    
    \vspace{0em}\\    
    \begin{subfigure}[b]{\textwidth}
        \centering
        \includegraphics[width=\textwidth]{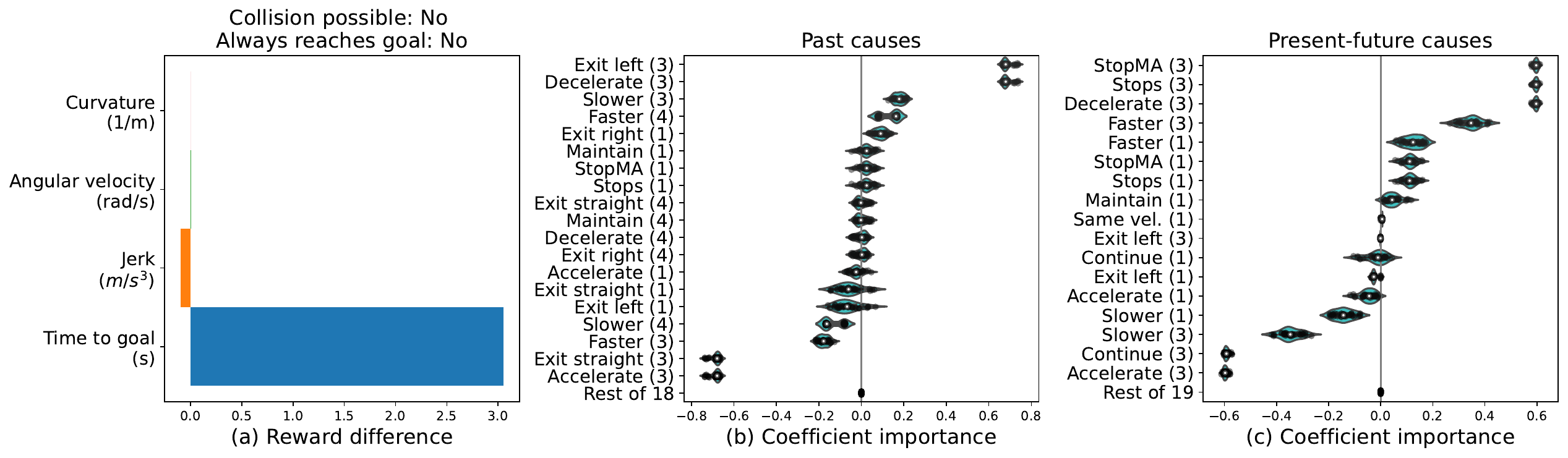}
    \end{subfigure}
    \vspace{0em}\\
    \begin{subfigure}[b]{0.49\textwidth}
        \centering
        \includegraphics[width=\textwidth]{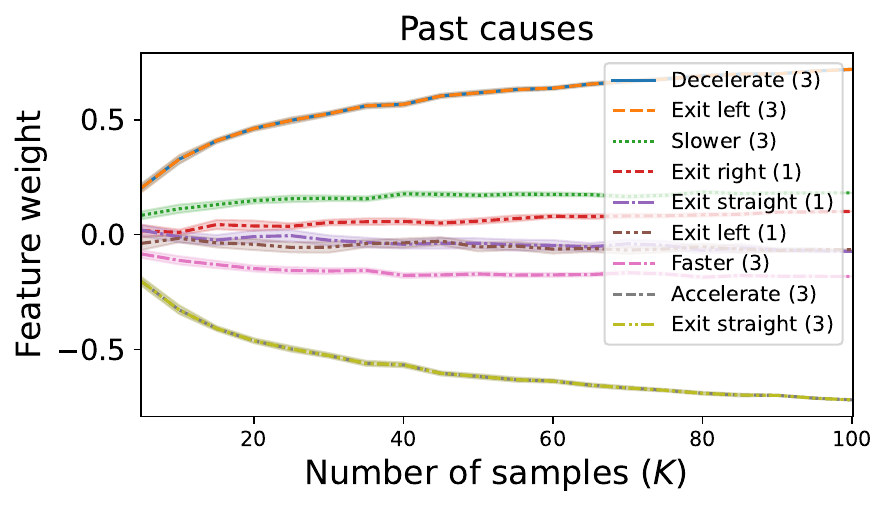}
    \end{subfigure}
    \hfill
    \begin{subfigure}[b]{0.49\textwidth}
        \centering
        \includegraphics[width=\textwidth]{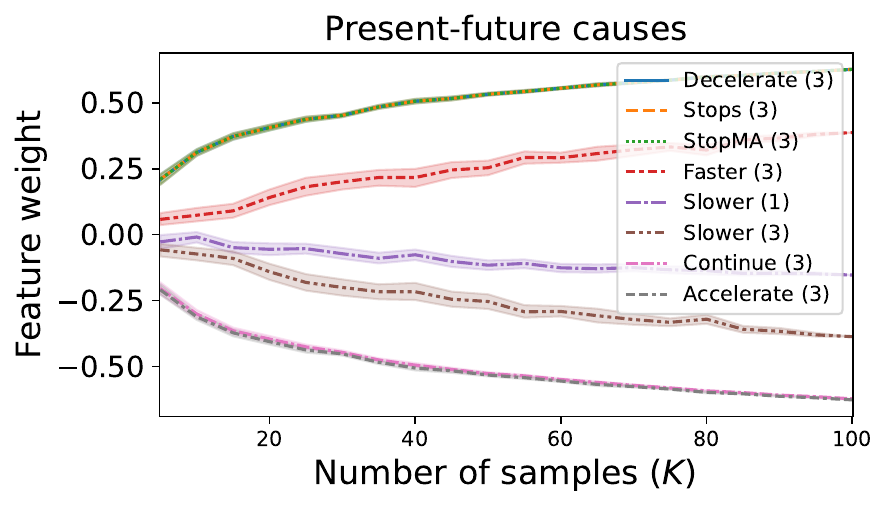}
    \end{subfigure}
    \vspace{1em}\\
    \begin{subfigure}[b]{0.49\textwidth}
        \centering
        \includegraphics[width=\textwidth]{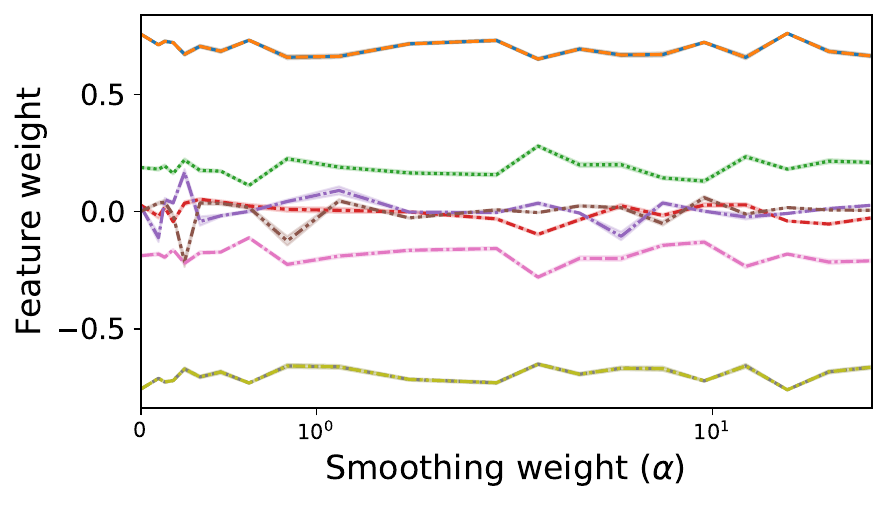}
    \end{subfigure}
    \hfill
    \begin{subfigure}[b]{0.49\textwidth}
        \centering
        \includegraphics[width=\textwidth]{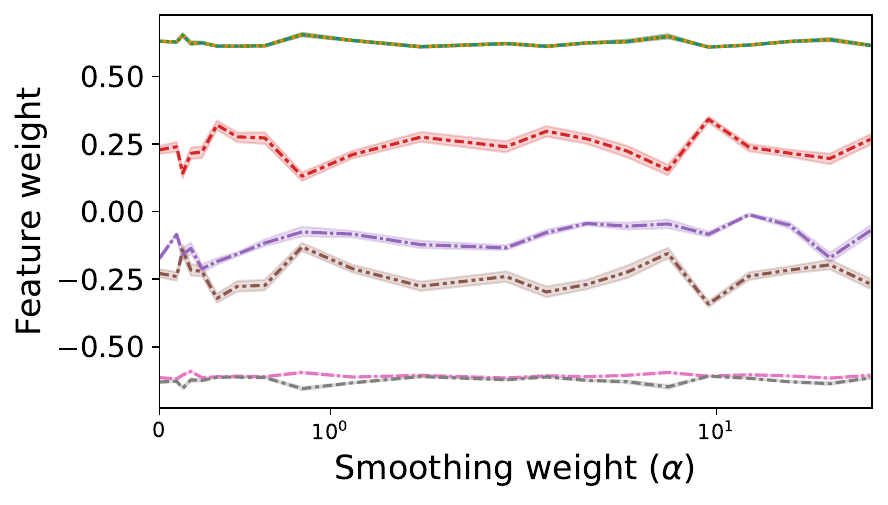}
    \end{subfigure}       
    \caption{Results for the first two questions in conversation S4.}
    \label{fig:s4-1}
\end{figure*}

\begin{figure*}
    \begin{subfigure}[b]{\textwidth}
        \centering
        \caption{Corresponding questions: \textit{``What if vehicle 3 went straight?''}}
        \begin{tabular}{l|llllllll}
        \toprule
             \textbf{Conv.} & \textbf{Type} & \textbf{VID} & \textbf{Tense} & \textbf{Actions} & \textbf{Query Time} & \textbf{Action Time} & \textbf{Negated} & \textbf{Factuals}  \\
            \midrule
             S4 & What if & 3 & Future & Go straight & 100 & --- & No & Stop \\
             \bottomrule
        \end{tabular}
    \end{subfigure}    
    \vspace{0em}\\    
    \begin{subfigure}[b]{\textwidth}
        \centering
        \includegraphics[width=\textwidth]{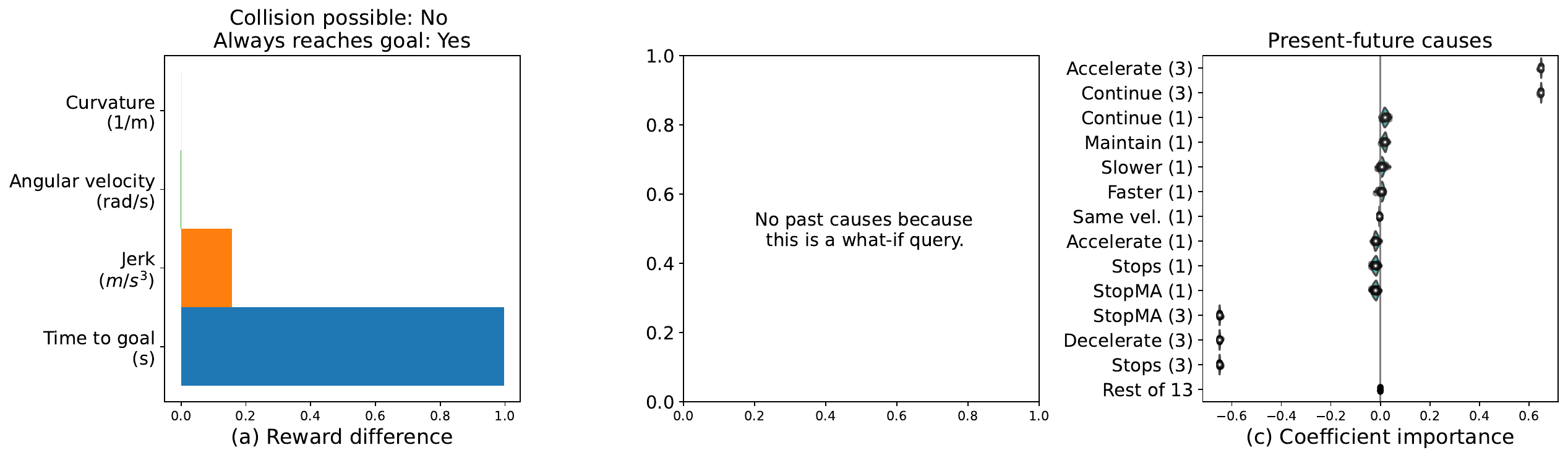}
    \end{subfigure}
    \vspace{0em}\\
    \begin{subfigure}[b]{0.5\textwidth}
        \centering
    \end{subfigure}
    \hfill
    \begin{subfigure}[b]{0.45\textwidth}
        \centering
        \includegraphics[width=\textwidth]{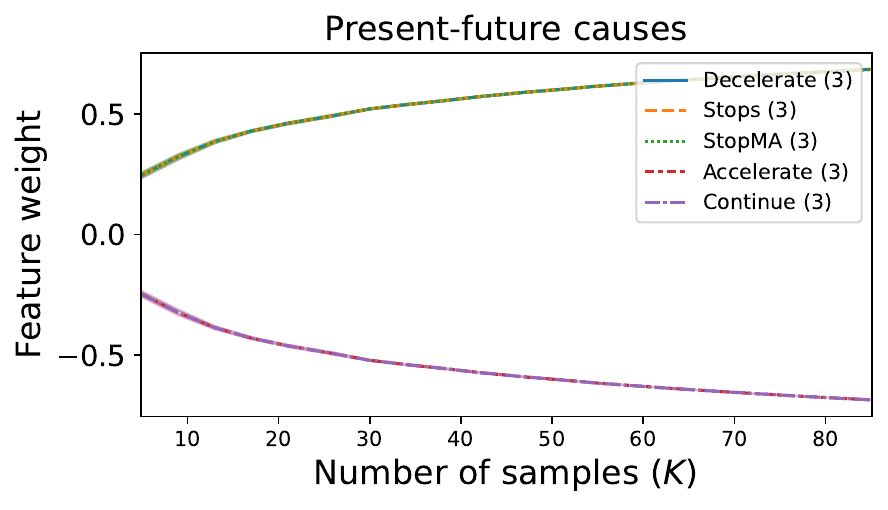}
    \end{subfigure}
    \hfill
    \vspace{1em}\\
    \begin{subfigure}[b]{0.5\textwidth}
        \centering
    \end{subfigure}
    \hfill
    \begin{subfigure}[b]{0.45\textwidth}
        \centering
        \includegraphics[width=\textwidth]{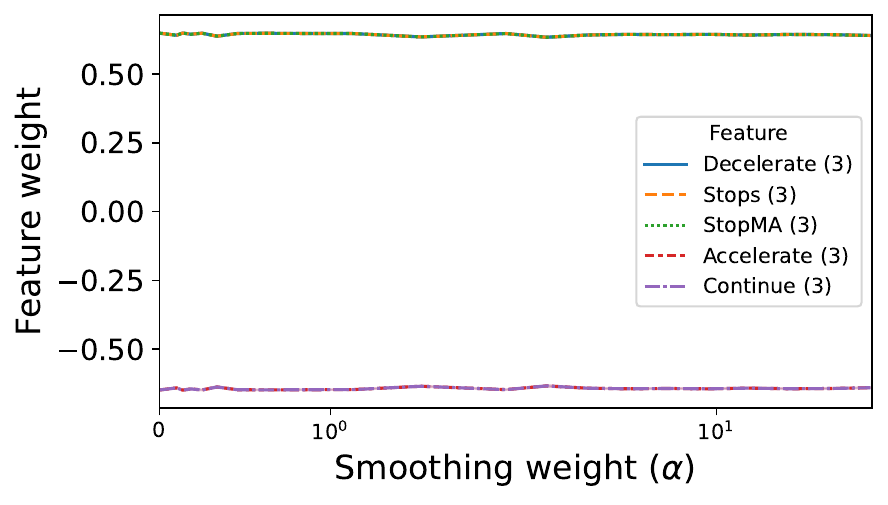}
    \end{subfigure}       
    \hfill
    \vspace{0em}\\    
    \caption{Results for the last question in conversation S4.}
    \label{fig:s4-2}
\end{figure*}

\begin{figure*}
    \centering
    \includegraphics[width=0.85\textwidth]{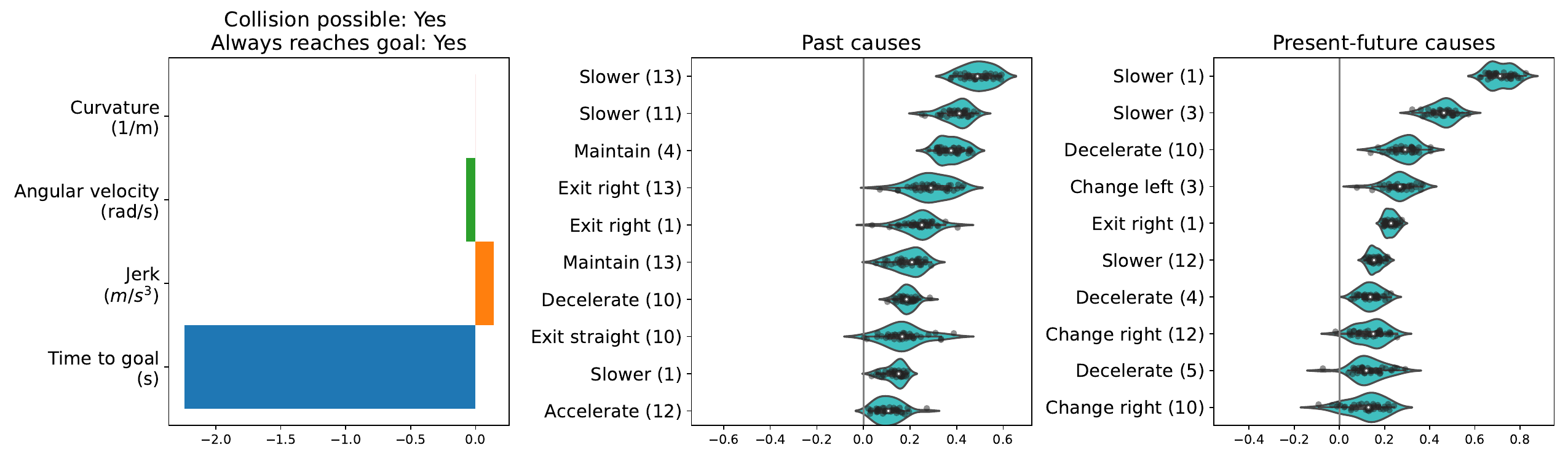}
    \caption{Top 10 causal importances in the extended scenario 1 with 20 agents.}
    \label{fig:s5}
\end{figure*}

\begin{figure*}
    \centering
    \includegraphics[width=0.85\textwidth]{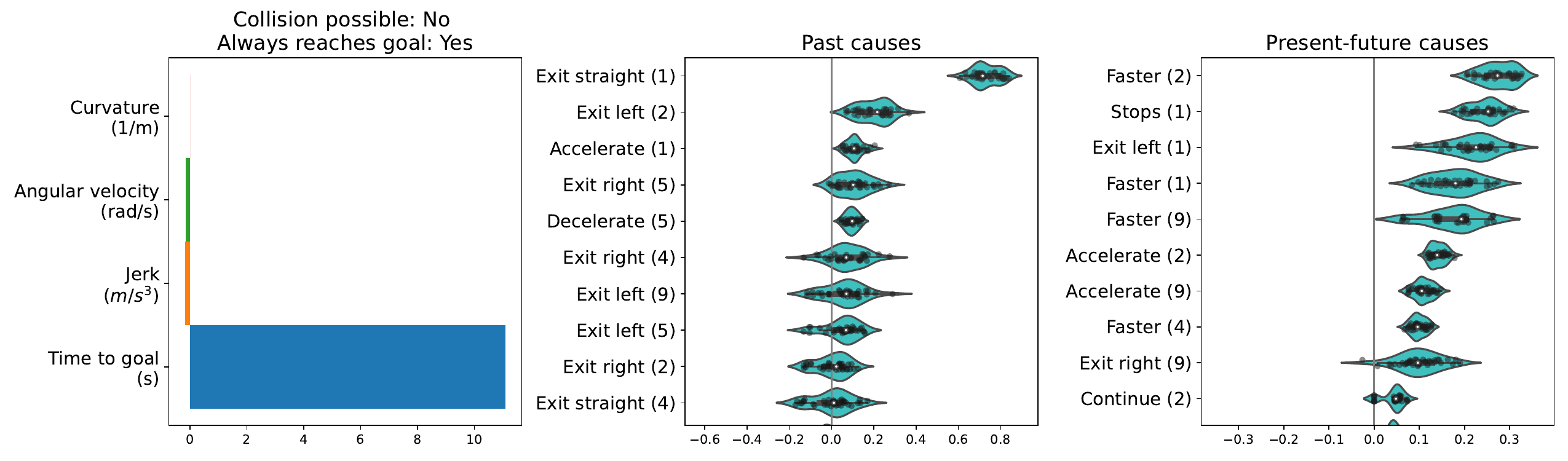}
    \caption{Top 10 causal importances in the extended scenario 2 with 16 agents.}
    \label{fig:s6}
\end{figure*}

\begin{figure*}
    \centering
    \includegraphics[width=0.85\textwidth]{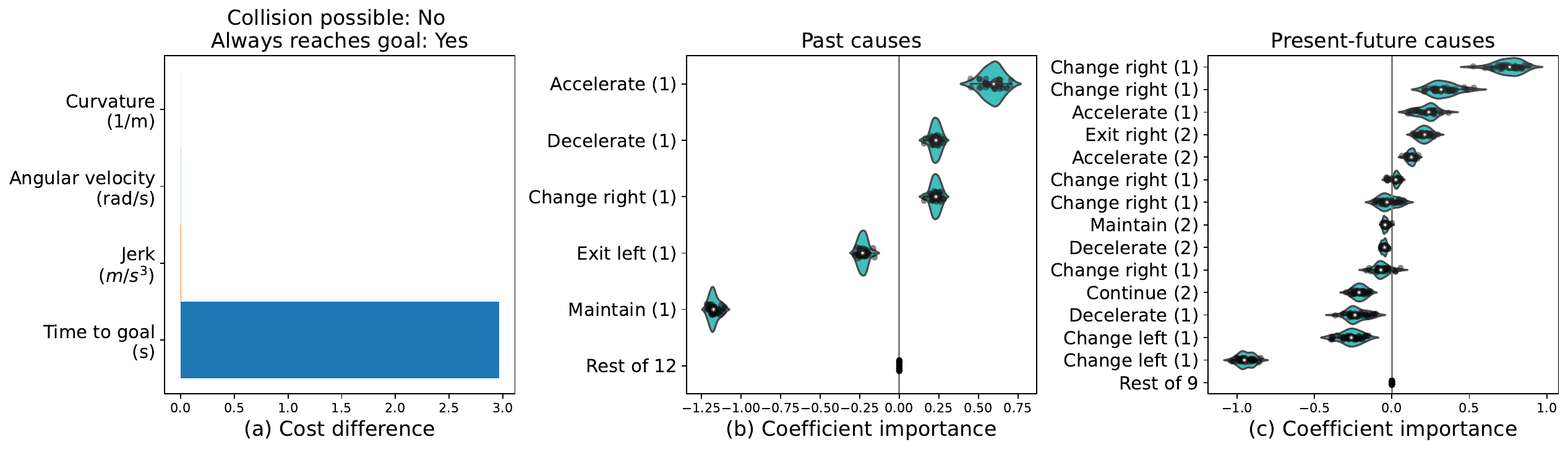}
    \caption{Top 10 causal importances in the extended scenario 3 with 4 agents.}
    \label{fig:s7}
\end{figure*}

\begin{figure*}
    \centering
    \includegraphics[width=0.85\textwidth]{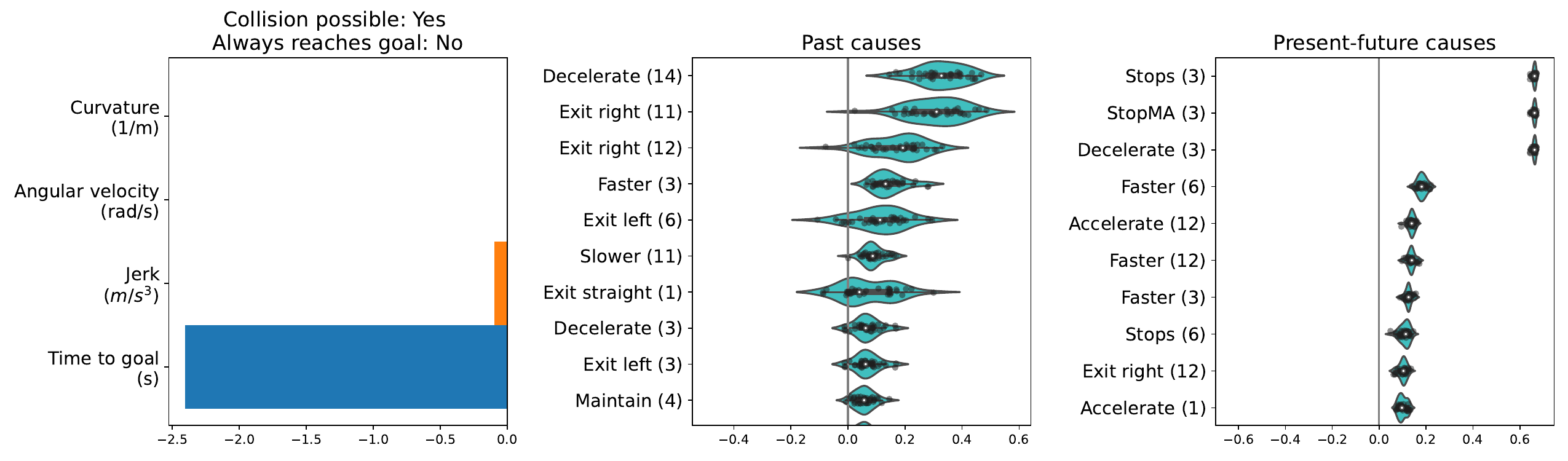}
    \caption{Top 10 causal importances in the extended scenario 4 with 16 agents.}
    \label{fig:s8}
\end{figure*}

We present our full computational results for all queries in \cref{fig:s1-1,fig:s1-2,fig:s1-3,fig:s2-1,fig:s2-2,fig:s3-1,fig:s4-1,fig:s4-2,fig:assc}.
\Cref{fig:assc} shows all \textit{associative} (i.e., descriptive) queries.
\Cref{fig:s5,fig:s6,fig:s7,fig:s8} show the causal selection results for the scaling experiments described in~\cref{apx:scaling}
Each figure gives natural language questions and the corresponding query content.
We also give the full plot of causal attributions as well as the results for robustness experiments.
In the plots of mechanistic causes, numbers in parentheses identify the ID of non-ego vehicles, if there is more than one non-ego.
The code to reproduce these results is part of the supplementary material.

Details about the user study follows the results of this section.

\begin{figure*}
    \centering
    \caption{All queries that provide an \textit{associative} explanation; across all conversations.}
    \label{fig:assc}\begin{tabular}{l|llllllll}
    \toprule
         \textbf{Conv.} & \textbf{Type} & \textbf{VID} & \textbf{Tense} & \textbf{Actions} & \textbf{Query Time} & \textbf{Action Time} & \textbf{Negated} & \textbf{Factuals}  \\
        \midrule
         S1-A & What if & 1 & Past & Change right & 75 & --- & Yes & --- \\        
         S1-B & What    & 0 & --- & --- & 45 & 70 & No & --- \\
         S2   & What if & 0 & Past & Go straight & 110 & --- & No & Turn left \\
         S3   & What    & 0 & --- & --- & 105 & 80 & No & ---\\      
         S4   & What if & 1 & Future & Go straight & 100 & --- & No & Stop        \\
         \bottomrule
    \end{tabular}
\end{figure*}

\clearpage
\clearpage
\section{User Study}

\subsection{Survey Methodology}
We performed two surveys as part of the user study. 
The first elicited high-quality explanations from participants about the behaviour of the autonomous vehicle as shown in pre-recorded video clips.
The second survey was used to compare the explanations of CEMA against this collected baseline dataset of human-written explanations.
We include the dataset of high-quality human explanations, responses to the second survey, and code to reproduce our statistical analysis of the second survey as part of the supplementary materials.

\subsubsection{Participants}
We used the crowd-sourcing platform Prolific to recruit participants.
We recruited from the USA, as the video recordings were in right-handed traffic, and filtered for participants whose first language was English.
Participants were paid an average of £11/h and were shown our ethics approval and a consent form which they had to accept before being allowed to fill out the surveys. 

Regarding the first survey, 54 participants filled it out with a median duration of completion of 25 minutes and 37 seconds.
The sex distribution was 25 males and 29 females.
The participants' age ranged between 19 to 73 years, with a median of 36 years. 
No participants were excluded for failing attention checks in the first survey.

Regarding the second survey, 200 participants filled it out with a median duration of completion of 9 minutes and 34 seconds.
Sexes were distributed as 99 males and 101 females.
The participants' age ranged between 19 to 77 years with a median of 35 years.
One participant was excluded for failing attention checks in the second survey.

\subsubsection{Survey 1: Design and Procedure}

In the first survey, people were shown 7 scenarios out of a collection of 14 including the four scenarios used in the evaluation of CEMA.
For each scenario, people were told what the goal of the ego agent is, and then they were shown a short (about 5 to 15-second-long) video of the scenario.
The videos were top-down animations recorded in the software RoadRunner 2023a by MathWorks.
We include the videos for the four scenarios used in our evaluation in the main paper.
After watching the videos, participants were asked to answer the following four questions:
\begin{enumerate}
    \item Describe the actions of the blue {car, self-driving car}.
    \item Explain why the blue {car, self-driving car} took these actions over different actions to reach its goal.
    \item Explain how the blue {car, self-driving car} was influenced in the scenario to take these actions.
    \item Describe changes to the scenario so that the blue {car, self-driving car} takes different actions? \textit{(The new actions need not be the best actions in the original scenario)}
\end{enumerate}
With 50\% chance, participants were shown either ``car'' or ``self-driving'' car.
We did this in order to get wide coverage of explanations, in case people were \textit{a priori} biased against autonomous vehicles.
Participants had to answer all four questions in their own words, entering their responses into unconstrained free-text boxes.
We also introduced two attention checks at the end of the page of random scenarios \#3 and \#6.

\subsubsection{Creating the Baseline Dataset}
Having collected explanations in Survey 1, we needed to assemble a useful and high-quality baseline dataset for the second survey.
As the second survey is only concerned with our four main scenarios, we excluded all explanations from Survey 1 that were not about scenarios S1-4.
This left a total of 408 explanations.
From this, we further removed another 26 explanations which were vacuous, that is, they did not contain any useful information about the scenario, or they expressed highly personal or obviously incorrect opinions.

To select explanations for the baseline dataset for Survey 2, we set up an annotation scheme with the following elements:
\begin{itemize}
    \item Over quality: Scale on 1 to 5. The overall quality of an explanation as determined by its relevancy to the scenario, its linguistic correctness, and subjective clarity;
    \item Complexity: Scale on 1 to 5. The complexity of the explanation as determined by its length and the number of causal relationships mentioned in it;
    \item Type: Whether the explanation is a teleological (1), mechanistic (2), or associative (i.e., descriptive) (3) explanation, or some combination of all three;
    \item Counterfactual: Boolean. Whether the explanation is contrastive or not.
\end{itemize}
We annotated all remaining 382 explanations with this scheme and randomly picked 25 explanations for each of the four scenarios S1-4.
We also corrected any obvious typos in each explanation.
Our selection criteria included only the highest quality explanations (level 4 or 5) and a wide range of combinations of complexities, types, and counterfactuals was covered as well.
This process in the end resulted in a selected dataset of high-quality baseline human-written explanations with a total size of 100 explanations.

\subsubsection{Survey 2: Design and Procedure}\label{sssec:survey2}
We used Survey 2 to measure the effects of explanations on trust and to understand what kind of explanations people preferred: the ones generated by CEMA or the human-written baseline explanations.
We randomised several parts of this survey, so for later reference, we will give names to random variables starting with an uppercase letter (shown in italics in parentheses), for example, \textit{Type}.

Before the start of the survey (after giving consent), we asked participants to answer three questions to measure their previous exposure to autonomous driving systems (\textit{AVExperience}), as classified by the SAE automation scale~\cite{saeinternationalTaxonomyDefinitionsTerms2021}.
The question were:
\begin{itemize}
    \item ``Have you ever used a driver-assistance system? Examples of driver-assistance systems are adaptive cruise control, lane keeping assistance, and parking assistance.'' (SAE L1-2);
    \item ``Have you ever used a partially autonomous self-driving car? A partially autonomous self-driving car is one which can drive on its own, but still requires an attentive human driver. An example is the Tesla Autopilot.'' (SAE L3);
    \item ``Have you ever used a fully autonomous self-driving car? A fully autonomous self-driving car is one which does not require a human driver. An example is the Waymo One ride-hailing service.'' (SAE L4).
\end{itemize}

The core of Survey 2 was the explanation rating scales.
For each participant, we randomly and with equal chance picked two of the four scenarios (\textit{Scenario}), as well as 10 explanations to be ranked.
The explanations were picked from a set of 30 total explanations per scenario consisting of 5 CEMA-generated explanations the 25 selected explanations from Survey 1 (\textit{Type}).
For each scenario, similarly to Survey 1, we first showed each participant the video recording of the scenario, though this time we did not tell them what the goal of the ego vehicle was to avoid biasing people's ratings.
We then asked participants to rate each explanation on a 5-point Likert scale based on how well each statement explained the actions of the ego vehicle (\textit{MeanRating}).
With 50\% chance, we also highlighted to participants in bold typeface the explanations that were generated by CEMA (\textit{Highlight}), and made sure that they were informed that these explanations were generated by the ego agent itself.

To measure changes to trust levels, we created the second survey with a between-subjects design, where the ``treatment'' was showing the participants the explanations and having them engage with the explanations through their ratings.
Therefore, we assigned each participant randomly with a 50\% chance into one of two groups: either answering the trust questions before seeing the explanations or after (\textit{PreExplanation}).
To actually measure people's trust (\textit{TrustRating}) in AVs, we used the trust scales recommended by~\citet{hoffmanMetricsExplainableAI2019} adapted to the AD domain.
People were asked the following nine questions in some random order (\textit{TrustQuestion}) and asked to rate them on a 5-point Likert scale:
\begin{enumerate}
    \item ``I am confident in self-driving cars. I feel that they work well'' (confidence in AVs).
    \item ``The decisions of self-driving cars are predictable.'' (AV decisions are predictable)
    \item ``Self-driving cars are reliable. I can count on them to be correct at all times.'' (AVs are reliable)
    \item ``I feel safe relying on a self-driving car. It will make the right decisions.'' (AVs are always correct)
    \item ``Self-driving cars are efficient, in that they make decisions very quickly.'' (AVs are efficient)
    \item ``I am wary of self-driving cars.'' (Wary of AVs)
    \item ``Self-driving cars make better decisions than novice drivers.'' (AVs better than novice)
    \item ``Self-driving cars make better decisions than experienced drivers.'' (AVs better than expert)
    \item ``I would like to use self-driving cars for decision making.'' (Willing to AVs)
\end{enumerate}

\subsubsection{Post-Survey Questions}

At the end of both surveys, participants were asked to fill out a brief survey about their driving experience about whether they hold a valid driver's license (\textit{License}), how many years of driving experience they have (\textit{Experience}), how many days a week they drive (Frequency), and how many miles on average they drive in a year (\textit{Distance}).
We also asked them voluntary demographic questions about their age range (\textit{Age}), gender (\textit{Gender}), and education level (\textit{Education}).
Finally, participants had the opportunity at the end of each survey to give any manner of feedback they thought worthy of mentioning.

\subsection{Analysis}

To understand the effects of explanations and the ratings of different types of explanations, we conducted a detailed statistical analysis of the survey data.
For our analysis, we used the R programming language~\cite{rmanual}.

\subsubsection{Pre-processing}
We pre-processed the raw survey data to make it more amenable to analysis with R.
This involved calculating the mean ratings for each participant grouped by the type of explanation (CEMA or Human) and the scenario (Scenario).
It also involved collating the trust responses across participants into a single column grouped by the trust scale (TrustQuestion) and turning the data into long format. 
Finally, in addition to the raw explanation ratings, we also included fields for the mean explanation ratings of participants grouped by their Type (\textit{MeanCEMARating} and \textit{MeanHumanRating}).

\subsubsection{Modelling}
We used Gaussian-family mixed-effects linear models to understand the relationships between the various variables.

To understand how participants' ratings of explanations varied with different variables (corresponding to testing \textbf{H1}), we fitted a model to our data predicting the mean explanation ratings (MeanRating) of participants.
We included the crossing between HighlightAV and Type, and AVExperience as independent variables encoding them using dummy coding.
We set the reference levels: HighlightAV=TRUE, Type=CEMA, AVExperience=0 (no experience).
The model also included varying intercepts for each scenario.

To understand how participants' trust levels changed after seeing explanations (corresponding to testing \textbf{H2}), we predicted the mean trust ratings (TrustRating) of participants.
We included as factors, PreExplanation, HighlightAV, AVExperience, and the crossing between MeanCEMARating and MeanHumanRating.
We encoded the categorical factors with dummy coding, setting the reference levels as follows: PreExplanation=TRUE, HighlightAV=TRUE, AVExperience=0.
We also included varying intercepts grouped by TrustQuestion.

We have also tested more complex models with more variables, however, there was no significant change in results by including these, and the model would often not converge, so we opted to use the least complex but most expressive model, which we have reported here.

\subsection{Results}
The following section presents a detailed view of the quantitative results of our analysis of Survey 2.

\subsubsection{Summary Statistics}

\begin{figure*}
    \centering
    \includegraphics[height=3.9cm]{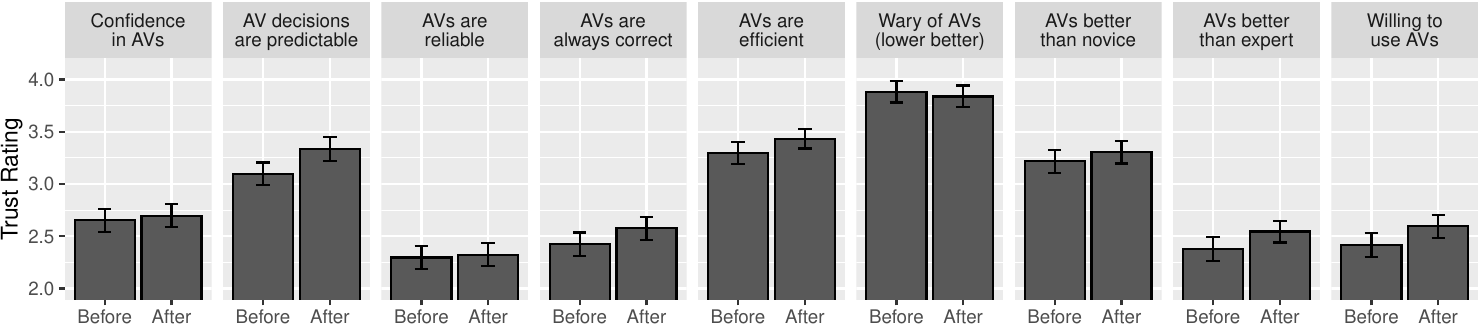}
    \caption{Mean ratings of participants on the 9 trust scales before and after having seen explanations with standard error.}
    \label{fig:trust}
\end{figure*}

\begin{figure*}
    \centering
    \includegraphics[width=\textwidth]{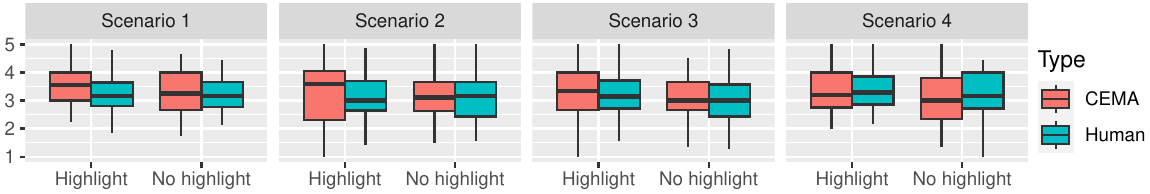}
    \caption{Ratings of CEMA's and human explanations with bootstrapped 95\% confidence interval (CI) grouped by whether CEMA's explanations were highlighted. The effect of highlighting is significant under our mixed-effects model ($p<0.05$).}
    \label{fig:expl}
\end{figure*}

We present in \cref{tab:pet-trust,tab:pot-trust} the summary statistics for the trust levels.
In \cref{tab:expl1,tab:expl2,tab:expl3,tab:expl4}, we give the summary statistics of the ratings of all 30 questions for each scenario. 
The questions in bold were generated by CEMA.
We also plot the distributions of trust before and after seeing explanations in \cref{fig:expl,fig:trust}.

\begin{table*}[]
    \centering
    \caption{Trust levels \textbf{before} seeing explanations (PreExplanation = TRUE). Trust scale corresponding to the enumeration in \cref{sssec:survey2}.}
    \label{tab:pet-trust}
    \begin{tabular}{lllllllll}
    \toprule
        \textbf{trust scale} &  \textbf{count} &        \textbf{mean} &         \textbf{std} &    \textbf{min} &     \textbf{25\%} &    \textbf{50\%} &     \textbf{75\%} &     \textbf{max} \\
    \midrule
        1           &  101.0 &    2.653465 &    1.117458 &    1.0 &    2.00 &    2.0 &    4.00 &     5.0 \\
        2           &  101.0 &    3.099010 &    1.081711 &    1.0 &    2.00 &    3.0 &    4.00 &     5.0 \\
        3           &  101.0 &    2.297030 &    1.072796 &    1.0 &    1.00 &    2.0 &    3.00 &     5.0 \\
        4           &  101.0 &    2.425743 &    1.116661 &    1.0 &    2.00 &    2.0 &    3.00 &     5.0 \\
        5           &  101.0 &    3.297030 &    1.091280 &    1.0 &    3.00 &    3.0 &    4.00 &     5.0 \\
        6           &  101.0 &    3.881188 &    1.022616 &    1.0 &    3.00 &    4.0 &    5.00 &     5.0 \\
        7           &  101.0 &    3.217822 &    1.109991 &    1.0 &    3.00 &    3.0 &    4.00 &     5.0 \\
        8           &  101.0 &    2.376238 &    1.164916 &    1.0 &    1.00 &    2.0 &    3.00 &     5.0 \\
        9           &  101.0 &    2.415842 &    1.168480 &    1.0 &    2.00 &    2.0 &    3.00 &     5.0 \\
    \bottomrule
    \end{tabular}
\end{table*}

\begin{table*}[]
    \centering
    \caption{Trust levels \textbf{after} seeing explanations (PreExplanation = FALSE). Trust scale corresponding to the enumeration in \cref{sssec:survey2}.}
    \label{tab:pot-trust}    
    \begin{tabular}{lllllllll}
    \toprule
        \textbf{trust scale} &  \textbf{count} &        \textbf{mean} &         \textbf{std} &    \textbf{min} &     \textbf{25\%} &    \textbf{50\%} &     \textbf{75\%} &     \textbf{max} \\
    \midrule
        1           &   99.0 &    2.696970 &    1.092222 &    1.0 &    2.00 &    2.0 &    3.00 &     5.0 \\
        2           &   99.0 &    3.333333 &    1.133893 &    1.0 &    3.00 &    4.0 &    4.00 &     5.0 \\
        3           &   99.0 &    2.323232 &    1.095709 &    1.0 &    1.50 &    2.0 &    3.00 &     5.0 \\
        4           &   99.0 &    2.575758 &    1.107405 &    1.0 &    2.00 &    2.0 &    4.00 &     5.0 \\
        5           &   99.0 &    3.434343 &    0.949346 &    1.0 &    3.00 &    4.0 &    4.00 &     5.0 \\
        6           &   99.0 &    3.838384 &    1.037138 &    1.0 &    3.00 &    4.0 &    5.00 &     5.0 \\
        8           &   99.0 &    3.303030 &    1.063826 &    1.0 &    3.00 &    3.0 &    4.00 &     5.0 \\
        9           &   99.0 &    2.545455 &    1.032855 &    1.0 &    2.00 &    3.0 &    3.00 &     5.0 \\
        10          &   99.0 &    2.595960 &    1.105821 &    1.0 &    2.00 &    2.0 &    3.50 &     5.0 \\
    \bottomrule
    \end{tabular}
\end{table*}

\begin{table*}
    \centering
    \caption{Base statistics of explanation ratings for the selected 30 questions in Scenario 1.}
    \label{tab:expl1}    
        \begin{tabular}{lrrrrrrrr}
        \toprule
        \textbf{question} &  \textbf{count} &      \textbf{mean} &       \textbf{std} &  \textbf{min} &  \textbf{25\%} &  \textbf{50\%} &   \textbf{75\%} &  \textbf{max} \\
        \midrule
        1        &  100.0 &  2.640000 &  1.267464 &  1.0 &  1.0 &  3.0 &  3.00 &  5.0 \\
        2        &   26.0 &  3.461538 &  1.103839 &  1.0 &  3.0 &  4.0 &  4.00 &  5.0 \\
        3        &  100.0 &  3.730000 &  1.081105 &  1.0 &  3.0 &  4.0 &  5.00 &  5.0 \\
        4        &   26.0 &  2.807692 &  1.059027 &  1.0 &  2.0 &  3.0 &  3.75 &  5.0 \\
        5        &  100.0 &  4.100000 &  1.010051 &  2.0 &  3.0 &  4.0 &  5.00 &  5.0 \\
        6        &   26.0 &  3.384615 &  0.803837 &  2.0 &  3.0 &  3.0 &  4.00 &  5.0 \\
        7        &   23.0 &  3.217391 &  1.241572 &  1.0 &  2.5 &  3.0 &  4.00 &  5.0 \\
        8        &   26.0 &  3.153846 &  1.255143 &  1.0 &  2.0 &  3.0 &  4.00 &  5.0 \\
        9        &   25.0 &  3.400000 &  1.118034 &  1.0 &  3.0 &  3.0 &  4.00 &  5.0 \\
        10       &   26.0 &  2.653846 &  1.198075 &  1.0 &  2.0 &  3.0 &  3.00 &  5.0 \\
        11       &   27.0 &  3.259259 &  1.163304 &  1.0 &  2.0 &  3.0 &  4.00 &  5.0 \\
        12       &   27.0 &  3.777778 &  1.012739 &  2.0 &  3.0 &  4.0 &  4.50 &  5.0 \\
        13       &   24.0 &  2.541667 &  1.062367 &  1.0 &  2.0 &  2.0 &  3.25 &  4.0 \\
        14       &   24.0 &  2.958333 &  1.334465 &  1.0 &  2.0 &  3.0 &  4.00 &  5.0 \\
        15       &   26.0 &  3.076923 &  1.262476 &  1.0 &  2.0 &  3.0 &  4.00 &  5.0 \\
        16       &   27.0 &  3.518519 &  1.087353 &  1.0 &  3.0 &  4.0 &  4.00 &  5.0 \\
        17       &   28.0 &  3.428571 &  0.997351 &  1.0 &  3.0 &  3.0 &  4.00 &  5.0 \\
        18       &   26.0 &  2.923077 &  1.293772 &  1.0 &  2.0 &  3.0 &  4.00 &  5.0 \\
        19       &   27.0 &  3.333333 &  1.176697 &  1.0 &  2.0 &  3.0 &  4.00 &  5.0 \\
        20       &   24.0 &  2.958333 &  1.122078 &  1.0 &  2.0 &  3.0 &  4.00 &  5.0 \\
        21       &   27.0 &  2.851852 &  0.988538 &  1.0 &  2.0 &  3.0 &  3.00 &  5.0 \\
        22       &   26.0 &  3.384615 &  1.202561 &  1.0 &  3.0 &  3.5 &  4.00 &  5.0 \\
        23       &   27.0 &  3.962963 &  0.939782 &  2.0 &  3.0 &  4.0 &  5.00 &  5.0 \\
        24       &   27.0 &  4.111111 &  1.219500 &  1.0 &  3.5 &  5.0 &  5.00 &  5.0 \\
        25       &   26.0 &  3.153846 &  1.120439 &  1.0 &  3.0 &  3.0 &  4.00 &  5.0 \\
        26       &   27.0 &  3.740741 &  0.813000 &  2.0 &  3.0 &  4.0 &  4.00 &  5.0 \\
        27       &   25.0 &  2.800000 &  1.258306 &  1.0 &  2.0 &  3.0 &  3.00 &  5.0 \\
        28       &   27.0 &  3.629630 &  1.005682 &  2.0 &  3.0 &  4.0 &  4.00 &  5.0 \\
        29       &   25.0 &  3.560000 &  0.916515 &  1.0 &  3.0 &  4.0 &  4.00 &  5.0 \\
        30       &   25.0 &  1.800000 &  0.912871 &  1.0 &  1.0 &  2.0 &  2.00 &  4.0 \\
        \bottomrule
        \end{tabular}
\end{table*}

\begin{table*}
    \centering
    \caption{Base statistics of explanation ratings for the selected 30 questions in Scenario 2.}
    \label{tab:expl2}    
        \begin{tabular}{lrrrrrrrr}
        \toprule
        \textbf{question} &  \textbf{count} &      \textbf{mean} &       \textbf{std} &  \textbf{min} &  \textbf{25\%} &  \textbf{50\%} &   \textbf{75\%} &  \textbf{max} \\
        \midrule
1        &  101.0 &  2.861386 &  1.191887 &  1.0 &  2.00 &  3.0 &  4.00 &  5.0 \\
2        &   27.0 &  3.555556 &  1.250641 &  1.0 &  3.00 &  4.0 &  4.00 &  5.0 \\
3        &  101.0 &  3.316832 &  1.264363 &  1.0 &  2.00 &  4.0 &  4.00 &  5.0 \\
4        &   27.0 &  2.740741 &  1.163304 &  1.0 &  2.00 &  3.0 &  4.00 &  5.0 \\
5        &  101.0 &  3.386139 &  1.288179 &  1.0 &  2.00 &  4.0 &  4.00 &  5.0 \\
6        &   24.0 &  1.750000 &  1.032094 &  1.0 &  1.00 &  1.0 &  2.25 &  4.0 \\
7        &   25.0 &  2.640000 &  1.319091 &  1.0 &  1.00 &  3.0 &  3.00 &  5.0 \\
8        &   25.0 &  3.440000 &  1.157584 &  1.0 &  3.00 &  4.0 &  4.00 &  5.0 \\
9        &   26.0 &  3.769231 &  1.274604 &  1.0 &  3.00 &  4.0 &  5.00 &  5.0 \\
10       &   27.0 &  3.740741 &  1.195909 &  1.0 &  3.50 &  4.0 &  4.50 &  5.0 \\
11       &   26.0 &  3.076923 &  1.293772 &  1.0 &  2.00 &  3.0 &  4.00 &  5.0 \\
12       &   25.0 &  3.160000 &  1.312758 &  1.0 &  2.00 &  3.0 &  4.00 &  5.0 \\
13       &   27.0 &  3.222222 &  1.368136 &  1.0 &  2.50 &  3.0 &  4.00 &  5.0 \\
14       &   26.0 &  2.884615 &  1.423430 &  1.0 &  2.00 &  3.0 &  4.00 &  5.0 \\
15       &   26.0 &  3.846154 &  0.967153 &  1.0 &  3.25 &  4.0 &  4.00 &  5.0 \\
16       &   27.0 &  2.851852 &  1.199478 &  1.0 &  2.00 &  3.0 &  4.00 &  5.0 \\
17       &   26.0 &  3.153846 &  1.461296 &  1.0 &  2.00 &  3.5 &  4.00 &  5.0 \\
18       &   27.0 &  3.518519 &  1.369176 &  1.0 &  2.00 &  4.0 &  4.50 &  5.0 \\
19       &   26.0 &  1.538462 &  0.859338 &  1.0 &  1.00 &  1.0 &  2.00 &  3.0 \\
20       &   25.0 &  2.960000 &  1.513275 &  1.0 &  2.00 &  3.0 &  5.00 &  5.0 \\
21       &   26.0 &  2.923077 &  1.262476 &  1.0 &  2.00 &  3.0 &  4.00 &  5.0 \\
22       &   26.0 &  3.269231 &  1.372813 &  1.0 &  2.00 &  3.5 &  4.00 &  5.0 \\
23       &   25.0 &  3.080000 &  1.222020 &  1.0 &  2.00 &  3.0 &  4.00 &  5.0 \\
24       &   27.0 &  2.555556 &  1.527525 &  1.0 &  1.00 &  2.0 &  4.00 &  5.0 \\
25       &   26.0 &  3.500000 &  0.948683 &  2.0 &  3.00 &  3.5 &  4.00 &  5.0 \\
26       &   27.0 &  2.925926 &  1.327981 &  1.0 &  2.00 &  3.0 &  4.00 &  5.0 \\
27       &   27.0 &  3.740741 &  1.403090 &  1.0 &  3.00 &  4.0 &  5.00 &  5.0 \\
28       &   27.0 &  3.222222 &  1.086042 &  1.0 &  2.00 &  4.0 &  4.00 &  5.0 \\
29       &   27.0 &  2.925926 &  1.412198 &  1.0 &  2.00 &  3.0 &  4.00 &  5.0 \\
30       &   27.0 &  3.111111 &  1.476309 &  1.0 &  2.00 &  3.0 &  4.00 &  5.0 \\
        \bottomrule
        \end{tabular}
\end{table*}

\begin{table*}
    \centering
    \caption{Base statistics of explanation ratings for the selected 30 questions in Scenario 3.}
    \label{tab:expl3}    
        \begin{tabular}{lrrrrrrrr}
        \toprule
        \textbf{question} &  \textbf{count} &      \textbf{mean} &       \textbf{std} &  \textbf{min} &  \textbf{25\%} &  \textbf{50\%} &   \textbf{75\%} &  \textbf{max} \\
        \midrule
1        &  98.0 &  2.857143 &  1.192762 &  1.0 &  2.00 &  3.0 &  4.00 &  5.0 \\
2        &  26.0 &  3.307692 &  1.123182 &  1.0 &  3.00 &  3.0 &  4.00 &  5.0 \\
3        &  98.0 &  3.693878 &  1.106757 &  1.0 &  3.00 &  4.0 &  5.00 &  5.0 \\
4        &  25.0 &  2.600000 &  1.258306 &  1.0 &  2.00 &  3.0 &  3.00 &  5.0 \\
5        &  98.0 &  3.163265 &  1.289880 &  1.0 &  2.00 &  3.0 &  4.00 &  5.0 \\
6        &  23.0 &  2.478261 &  1.591728 &  1.0 &  1.00 &  3.0 &  3.50 &  5.0 \\
7        &  26.0 &  3.346154 &  1.263085 &  1.0 &  2.00 &  4.0 &  4.00 &  5.0 \\
8        &  26.0 &  3.884615 &  1.275207 &  1.0 &  3.00 &  4.0 &  5.00 &  5.0 \\
9        &  26.0 &  3.653846 &  1.354764 &  1.0 &  3.00 &  4.0 &  5.00 &  5.0 \\
10       &  25.0 &  3.320000 &  1.314027 &  1.0 &  2.00 &  3.0 &  4.00 &  5.0 \\
11       &  25.0 &  3.360000 &  1.254326 &  1.0 &  3.00 &  3.0 &  5.00 &  5.0 \\
12       &  24.0 &  3.458333 &  1.215092 &  1.0 &  3.00 &  4.0 &  4.00 &  5.0 \\
13       &  25.0 &  3.160000 &  1.247664 &  1.0 &  2.00 &  3.0 &  4.00 &  5.0 \\
14       &  26.0 &  2.192308 &  1.414757 &  1.0 &  1.00 &  2.0 &  3.00 &  5.0 \\
15       &  26.0 &  3.692308 &  1.378963 &  1.0 &  3.00 &  4.0 &  5.00 &  5.0 \\
16       &  26.0 &  2.730769 &  1.457606 &  1.0 &  1.00 &  3.0 &  4.00 &  5.0 \\
17       &  26.0 &  2.807692 &  1.497177 &  1.0 &  1.25 &  3.0 &  4.00 &  5.0 \\
18       &  26.0 &  3.692308 &  1.225373 &  1.0 &  3.00 &  4.0 &  5.00 &  5.0 \\
19       &  24.0 &  2.333333 &  1.403928 &  1.0 &  1.00 &  2.0 &  3.25 &  5.0 \\
20       &  23.0 &  3.478261 &  1.441892 &  1.0 &  2.50 &  4.0 &  5.00 &  5.0 \\
21       &  26.0 &  2.692308 &  1.257592 &  1.0 &  2.00 &  3.0 &  3.00 &  5.0 \\
22       &  27.0 &  3.925926 &  1.106829 &  1.0 &  3.00 &  4.0 &  5.00 &  5.0 \\
23       &  25.0 &  3.760000 &  1.267544 &  1.0 &  3.00 &  4.0 &  5.00 &  5.0 \\
24       &  25.0 &  3.040000 &  1.171893 &  1.0 &  2.00 &  3.0 &  4.00 &  5.0 \\
25       &  27.0 &  3.814815 &  1.075498 &  1.0 &  3.00 &  4.0 &  5.00 &  5.0 \\
26       &  26.0 &  2.730769 &  1.218448 &  1.0 &  2.00 &  3.0 &  3.75 &  5.0 \\
27       &  26.0 &  2.538462 &  1.475961 &  1.0 &  1.00 &  2.0 &  4.00 &  5.0 \\
28       &  25.0 &  2.520000 &  1.326650 &  1.0 &  1.00 &  2.0 &  3.00 &  5.0 \\
29       &  26.0 &  2.115385 &  1.107318 &  1.0 &  1.00 &  2.0 &  3.00 &  4.0 \\
30       &  25.0 &  3.680000 &  1.069268 &  2.0 &  3.00 &  4.0 &  5.00 &  5.0 \\
        \bottomrule
        \end{tabular}
\end{table*}

\begin{table*}
    \centering
    \caption{Base statistics of explanation ratings for the selected 30 questions in Scenario 4.}
    \label{tab:expl4}    
        \begin{tabular}{lrrrrrrrr}
        \toprule
        \textbf{question} &  \textbf{count} &      \textbf{mean} &       \textbf{std} &  \textbf{min} &  \textbf{25\%} &  \textbf{50\%} &   \textbf{75\%} &  \textbf{max} \\
        \midrule
1        &  101.0 &  2.871287 &  1.270145 &  1.0 &  2.00 &  3.0 &  4.00 &  5.0 \\
2        &   27.0 &  3.629630 &  1.079464 &  1.0 &  3.00 &  4.0 &  4.00 &  5.0 \\
3        &   25.0 &  3.560000 &  1.083205 &  2.0 &  3.00 &  3.0 &  5.00 &  5.0 \\
4        &  101.0 &  2.900990 &  1.360183 &  1.0 &  2.00 &  3.0 &  4.00 &  5.0 \\
5        &  101.0 &  3.722772 &  1.078136 &  1.0 &  3.00 &  4.0 &  5.00 &  5.0 \\
6        &   27.0 &  2.777778 &  1.154701 &  1.0 &  2.00 &  3.0 &  3.00 &  5.0 \\
7        &   24.0 &  3.833333 &  0.868115 &  2.0 &  3.00 &  4.0 &  4.25 &  5.0 \\
8        &   26.0 &  3.346154 &  1.324909 &  1.0 &  2.00 &  3.0 &  4.75 &  5.0 \\
9        &   27.0 &  3.074074 &  1.639088 &  1.0 &  1.00 &  3.0 &  4.50 &  5.0 \\
10       &   26.0 &  3.076923 &  1.440085 &  1.0 &  2.00 &  3.0 &  4.00 &  5.0 \\
11       &   27.0 &  3.925926 &  1.141050 &  1.0 &  3.50 &  4.0 &  5.00 &  5.0 \\
12       &   26.0 &  3.269231 &  1.250846 &  1.0 &  3.00 &  3.0 &  4.00 &  5.0 \\
13       &   26.0 &  3.730769 &  1.041449 &  1.0 &  3.00 &  4.0 &  4.00 &  5.0 \\
14       &   26.0 &  3.346154 &  1.383974 &  1.0 &  2.00 &  3.0 &  5.00 &  5.0 \\
15       &   26.0 &  3.730769 &  1.401647 &  1.0 &  3.00 &  4.0 &  5.00 &  5.0 \\
16       &   26.0 &  3.269231 &  1.401647 &  1.0 &  2.25 &  3.0 &  4.75 &  5.0 \\
17       &   25.0 &  2.520000 &  1.084743 &  1.0 &  2.00 &  3.0 &  3.00 &  5.0 \\
18       &   27.0 &  3.148148 &  1.166972 &  1.0 &  2.00 &  4.0 &  4.00 &  5.0 \\
19       &   27.0 &  2.555556 &  1.187542 &  1.0 &  2.00 &  3.0 &  3.00 &  5.0 \\
20       &   27.0 &  3.518519 &  1.155934 &  1.0 &  3.00 &  4.0 &  4.00 &  5.0 \\
21       &   27.0 &  3.000000 &  1.441153 &  1.0 &  2.00 &  3.0 &  4.00 &  5.0 \\
22       &   25.0 &  3.840000 &  1.027943 &  2.0 &  3.00 &  4.0 &  5.00 &  5.0 \\
23       &   27.0 &  2.555556 &  1.187542 &  1.0 &  2.00 &  2.0 &  3.50 &  5.0 \\
24       &   26.0 &  3.461538 &  1.475961 &  1.0 &  2.25 &  4.0 &  4.75 &  5.0 \\
25       &   26.0 &  2.846154 &  1.461296 &  1.0 &  2.00 &  3.0 &  4.00 &  5.0 \\
26       &   26.0 &  3.269231 &  1.250846 &  1.0 &  2.00 &  3.0 &  4.00 &  5.0 \\
27       &   26.0 &  3.923077 &  0.934797 &  2.0 &  3.00 &  4.0 &  5.00 &  5.0 \\
28       &   26.0 &  3.230769 &  1.242826 &  1.0 &  2.25 &  3.0 &  4.00 &  5.0 \\
29       &   26.0 &  3.346154 &  1.294366 &  1.0 &  2.25 &  4.0 &  4.00 &  5.0 \\
30       &   27.0 &  4.074074 &  0.916764 &  2.0 &  4.00 &  4.0 &  5.00 &  5.0 \\
        \bottomrule
        \end{tabular}
\end{table*}

\begin{table}[]
    \centering
    \caption{Estimated coefficients ($\hat{\beta}$) of the fixed effects of the mixed-effects model predicting \textbf{explanation} ratings.}
    \label{tab:expl-fixed}    
    \begin{tabular}{lrrr}
    \toprule
         {}                        & \textbf{Estimate}    & \textbf{Std. Error} & \textbf{t value}  \\
         \midrule
        (Intercept)                &  3.31082    & 0.07517    & 44.044 \\
        HighlightAVFALSE           & -0.22321    & 0.08054    & -2.771 \\
        TypeHuman                  & -0.15920    & 0.08038    & -1.981 \\
        AVExperienceTRUE           &  0.09922    & 0.05926    &  1.674 \\
        HighlightAVFALSE &  0.17437    & 0.11367    &  1.534 \\
        :TypeHuman       & {} & {} & {} \\
        \bottomrule
    \end{tabular}
\end{table}

\begin{table}[]
    \centering
    \caption{Variation of intercepts by scenario of the mixed-effects model predicting explanation ratings.}
    \label{tab:expl-intercepts}    
    \begin{tabular}{lr}
    \toprule
        {}            & \textbf{(Intercept)}   \\
    \midrule
        Scenario 1    & 3.376390 \\
        Scenario 2    & 3.259917 \\
        Scenario 3    & 3.279770 \\
        Scenario 4    & 3.327218 \\
    \bottomrule
    \end{tabular}
\end{table}

\begin{table}[]
    \centering
    \caption{Estimated variation of the random effects, in this case just the Scenario, of the mixed-effects model predicting explanation ratings.}
    \label{tab:expl-random}    
    \begin{tabular}{llll}
    \toprule
         \textbf{Groups}   & \textbf{Name}        & \textbf{Variance} & \textbf{Std.Dev.} \\
    \midrule
         Scenario & (Intercept) & 0.004611 & 0.0679  \\
         Residual &             & 0.646100 & 0.8038  \\
    \bottomrule
    \end{tabular}
\end{table}

\begin{table}[]
    \centering
    \caption{Analysis of Deviance Table (Type II Wald chi-square tests) with response variable of explanation ratings. Significance codes:  0 `***' 0.001 `**' 0.01 `*' 0.05 `.' 0.1 ` ' 1}
    \label{tab:expl-significance}    
    \begin{tabular}{lrrr}
    \toprule
        {}               & \textbf{Chisq}   & \textbf{Df}   & \textbf{Pr($>$Chisq)}   \\
        \midrule
        HighlightAV      & 5.6826  & 1    & * 0.01713 \\
        Type             & 1.6055  & 1    & 0.20513   \\
        AVExperience     & 2.8032  & 1    & . 0.09408 \\
        HighlightAV:Type & 2.3529  & 1    & 0.12505  \\
        \bottomrule
    \end{tabular}
\end{table}

\begin{table}[]
    \centering
    \caption{Estimated coefficients ($\hat{\beta}$) of the fixed effects of the mixed-effects model predicting \textbf{trust} ratings.}
    \label{tab:trust-fixed}    
    \begin{tabular}{lrrr}
    \toprule
         {}                        & \textbf{Estimate}    & \textbf{Std. Error} & \textbf{t value}  \\
         \midrule
        (Intercept)                &  1.52881   & 0.50029  &3.056\\
        MeanCEMARank               &  0.35401   & 0.15279  &2.317\\
        MeanHumanRank              &  0.11929   & 0.15360  &0.777\\
        PreExplanationFALSE        &  0.11044   & 0.05158  &2.141\\
        HighlightAVFALSE           &  0.02302   & 0.05186  &0.444\\
        AVExperienceTRUE           &  0.33170   & 0.05248  &6.321\\
        MeanCEMARank & -0.04054   & 0.04477  &-0.906\\
        :MeanHumanRank & {} & {} & {} \\
        \bottomrule
    \end{tabular}
\end{table}

\begin{table}[]
    \centering
    \caption{Variation of intercepts by trust scales of the mixed-effects model predicting trust ratings.}
    \label{tab:trust-intercepts}    
    \begin{tabular}{lr}
    \toprule
        {}            & \textbf{(Intercept)}  \\
    \midrule
        {}         & (Intercept) \\
        Question 1   & 1.3028370 \\
        Question 2   & 1.8321074\\
        Question 3   & 0.9450894\\
        Question 4   & 1.1313141\\
        Question 5   & 1.9791269\\
        Question 6   & 2.4642915\\
        Question 7   & 1.8762132\\
        Question 8   & 1.0921089\\
        Question 9   & 1.1362148 \\
    \bottomrule
    \end{tabular}
\end{table}

\begin{table}[]
    \centering
    \caption{Estimated variation of the random effects, in this case just the trust scales, of the mixed-effects model predicting trust ratings.}
    \label{tab:trust-random}    
    \begin{tabular}{rrll}
    \toprule
         \textbf{Groups}   & \textbf{Name}        & \textbf{Variance} & \textbf{Std.Dev.} \\
    \midrule
         TrustQuestion & (Intercept) &0.2787  & 0.5279 \\  
         Residual      &            &1.1299 &  1.0630  \\
    \bottomrule
    \end{tabular}
\end{table}

\begin{table}[]
    \centering
    \caption{Analysis of Deviance Table (Type II Wald chi-square tests) with response variable of trust ratings. Significance codes:  0 `***' 0.001 `**' 0.01 `*' 0.05 `.' 0.1 ` ' 1}
    \label{tab:trust-significance}    
    \begin{tabular}{lrrr}
    \toprule
        {}               & \textbf{Chisq}   & \textbf{Df}   & \textbf{Pr($>$Chisq)}   \\
        \midrule
        MeanCEMARank               & 21.8790  & 1  & *** 2.904e-06\\
        MeanHumanRank              &  0.0379  & 1  &   0.84563    \\
        PreExplanation             &  4.5852  & 1  &   * 0.03225   \\
        HighlightAV                &  0.1970  & 1  &   0.65719    \\
        AVExperience               & 39.9520  & 1  & *** 2.603e-10\\
        MeanCEMARank &  0.8202  & 1  &   0.36511   \\
        :MeanHumanRank & {} &{} &{}\\
        \bottomrule
    \end{tabular}
\end{table}

\clearpage

\subsubsection{Results of Statistical Analysis}
We present in the previous page the full results of our statistical analysis using the mixed-effects model for both the explanation ratings-related and trust ratings-related experiments
In \cref{tab:expl-fixed,tab:trust-fixed}, we report the estimated means and standard errors of the fixed effects.
In \cref{tab:expl-intercepts,tab:trust-intercepts}, we report the changes to the estimated mean of the intercepts due to the grouping by the random effects variables in the model.
In \cref{tab:expl-random,tab:trust-random}, we report the variation in the random effects variables.
Finally, in \cref{tab:expl-significance,tab:trust-significance}, we report the significance of each fixed effects variable on the predicted value of the outcome variable.

\subsection{ChatGPT Prompts and Replies}
One possible additional baseline that has recently emerged is the family of large language models and their public-facing demonstrations such as ChatGPT~\cite{openaiChatGPTOptimizingLanguage2022}.
While it is strongly debated whether these systems have a grasp on causality, nonetheless, their impressive fluency and apparent reasoning abilities should not be ignored.

Therefore, we have also explored the kind of explanations that ChatGPT can give by describing each scenario and then prompting it for an explanation.
For all scenarios, tedious prompt editing and re-prompting were required to produce useful responses. 
We found that ChatGPT could identify causes correctly behind S1 and S4, and failed to explain S2 and S3.
However, it is impossible to determine how and why specific causes are selected by ChatGPT. 
It is a closed system which changes frequently and its responses vary for the same prompts making it difficult to use as a baseline.

We give the prompts and the obtained responses from our exploration of ChatGPT's ability to explain behaviour in our scenarios. 
We used GPT-3.5 for these explanations.

\subsubsection{Scenario 1}
ChatGPT identifies a safety risk in the orange vehicle (non-ego) cutting in front of the blue vehicle (ego).
After additional prompting, it gives further correct and plausible reasons to why the ego would change lanes left in this scenario.

\begin{description}
    \item\textbf{Prompt:} 
    There is a blue vehicle and an orange vehicle on a two-laned road. 
    The orange vehicle is in the left lane 20 meters ahead of the blue vehicle which is in the right lane. 
    The road is approaching a T-junction with an exit on the right. 
    The orange vehicle changes lanes to the right and begins to slow down. 
    In response, the blue vehicle changes lanes to the left. Why did the blue vehicle change lanes to the left?
    
    \item\textbf{Response:} 
    The blue vehicle most likely changed lanes to the left to avoid a potential collision with the orange vehicle, which was slowing down and changing lanes to the right.
    Changing lanes to the left would give the blue vehicle more space and distance from the orange vehicle, reducing the risk of a collision as the road approaches the T-junction with an exit on the right. 
    It's a defensive driving technique to maintain a safe distance from other vehicles and anticipate any potential hazards or obstacles on the road.
    
    \item\textbf{Prompt:}
    Besides avoiding a collision, what else could have prompted the blue vehicle to change lanes left?
    
    \item\textbf{Response:}
    There could be several other reasons why the blue vehicle changed lanes to the left besides avoiding a collision with the orange vehicle. Here are some possibilities:
    
    \begin{enumerate}
        \item Overtaking: The blue vehicle may have wanted to overtake another slower-moving vehicle in front of it, and changing lanes to the left provided a clear path to do so.
        \item Obstruction: There may have been an obstruction in the right lane that the blue vehicle wanted to avoid, such as debris, a pothole, or a parked vehicle.
        \item Turning left: The blue vehicle may have been planning to turn left at the T-junction, and changing lanes to the left put it in the correct position for the turn.
        \item Personal preference: The driver of the blue vehicle may have simply preferred driving in the left lane for personal reasons, such as feeling more comfortable or having better visibility.
    \end{enumerate}
    However, without more context about the specific situation, it's impossible to know for sure why the blue vehicle changed lanes to the left. \textit{[This last sentence is also by ChatGPT]}
\end{description}

\subsubsection{Scenario 2}
ChatGPT cannot identify the correct causes behind the blue vehicle's actions (ego) even when we give a large number of contextual clues.
It gives completely incorrect guesses about why the blue vehicle might not give way to the orange vehicle (non-ego).

\begin{description}
    \item\textbf{Prompt:}
    A blue vehicle is approaching a junction without traffic lights from the south on a secondary road. 
    There is an orange vehicle coming from the left and a green vehicle coming from the right on a main road. 
    The green vehicle is maintaining a high speed and is heading straight through the junction, and the orange vehicle is slowing down to a stop likely to turn left and waiting for the green vehicle to pass. 
    In response, the blue vehicle turns right onto the main road without giving way to the orange vehicle. 
    Why did the blue vehicle not give way?
    
    \item\textbf{Response:}
    Based on the scenario you described, the blue vehicle should have given way to the orange vehicle before turning right onto the main road.
    Failing to give way to the orange vehicle would be a violation of traffic laws in most jurisdictions, and could result in a collision or other dangerous situation.
    
    Without further information about the specific circumstances and the actions of the drivers involved, it's difficult to determine why the blue vehicle did not give way. 
    However, some possible reasons could include the driver not being aware of the orange vehicle's presence or intention to turn left, or misjudging the distance and speed of the orange vehicle. 
    It's important for drivers to always be aware of their surroundings and give way to other vehicles when required by traffic laws and common sense.
\end{description}

At this point, further prompting (e.g., telling ChatGPT explicitly that it is safe not to give way) does not change the responses of ChatGPT significantly.

\subsubsection{Scenario 3}
ChatGPT cannot identify the correct causes behind the blue vehicle's actions (ego) even when we give a large number of contextual clues.
Further prompting (e.g., telling ChatGPT explicitly that it is safe not to give way) does not change the responses of ChatGPT significantly.

\begin{description}
    \item\textbf{Prompt:} 
    There is a blue vehicle and an orange vehicle. 
    The orange vehicle is in a two-laned roundabout and it has just changed lanes from the inner lane to the outer lane. 
    The blue vehicle is on a road that enters the roundabout and it has to give way to the orange vehicle. 
    On observing the orange vehicle change lanes, the blue vehicle decides to enter the roundabout without giving way to the orange vehicle.
    Why was it safe to enter the roundabout?
    
    \item\textbf{Response:}
    As an AI language model, I do not have access to real-time information about the traffic conditions in your specific location. 
    However, based on the scenario you have described, it is not safe for the blue vehicle to enter the roundabout without giving way to the orange vehicle, even if the orange vehicle had just changed lanes.
    
    Failing to give way to a vehicle that has already entered a roundabout can cause a collision, particularly if the other vehicle is already in the driver's blind spot. 
    It is important to remember that all drivers on the road have a responsibility to drive safely and follow traffic laws to prevent accidents.
    
    Therefore, it is recommended that the blue vehicle wait until the orange vehicle has passed 
    
    \item\textbf{Prompt:}
    However, the orange vehicle is also slowing down which indicates that it will exit the roundabout without crossing the path of the blue vehicle. 
    Why was it safe to enter the roundabout for the blue vehicle now?
    
    \item\textbf{Response:}
    If the orange vehicle is slowing down and is about to exit the roundabout without crossing the path of the blue vehicle, then it may be safe for the blue vehicle to enter the roundabout.
    
    However, it is important to note that the blue vehicle should only enter the roundabout if it is safe to do so, and after ensuring that there are no other vehicles entering or exiting the roundabout that could potentially cross its path.
    
    Additionally, the blue vehicle should always follow traffic laws and signals, and should be prepared to stop or yield to other vehicles if necessary.
    
    Therefore, if the blue vehicle determines that it is safe to enter the roundabout and that it will not impede the flow of traffic or cause a potential collision, then it may proceed to enter the roundabout.
\end{description}

\subsubsection{Scenario 4}
Initially, ChatGPT assumes certain facts about the environment which are not true, but on some further clarifications, it can identify the correct causes behind the orange vehicle stopping at the junction. 

\begin{description}
    \item\textbf{Prompt:}
    There is a waiting line of cars behind a red traffic light. 
    Just behind this line of cars is a secondary road that enters the main road in a T-junction. 
    There is a blue vehicle on the secondary road waiting to turn left onto the main road. 
    An orange vehicle on the main road stops before the T-junction. 
    In reaction, the blue vehicle turns left onto the main road and joins the waiting line of cars. 
    Why did the blue vehicle turn left?
    
    \item\textbf{Response:}
    Based on the information provided, it seems that the orange vehicle on the main road has stopped before the T-junction, potentially indicating that there is an obstruction or some other reason preventing it from continuing forward. 
    This may have prompted the blue vehicle on the secondary road to take advantage of the opportunity to turn left onto the main road and join the waiting line of cars, rather than waiting for the traffic light to turn green.
    
    Alternatively, it is possible that the blue vehicle was already intending to turn left onto the main road, and the stopping of the orange vehicle simply provided a clear opening for the blue vehicle to make the turn safely. 
    Without further information, it is difficult to determine the exact reason why the blue vehicle turned left onto the main road.
    
    \item\textbf{Prompt:}
    There are no obstacles on the road so the orange vehicle could have just joined the line of waiting cars, thus blocking the way of the blue vehicle. 
    Yet it did not. 
    In light of this, why could the blue vehicle turn left?
    
    \item\textbf{Respose:}
    If there were no obstacles on the road and the orange vehicle could have joined the line of waiting cars, but did not, it is possible that the orange vehicle driver intentionally stopped before the T-junction to let the blue vehicle turn left onto the main road. 
    This could have been a courtesy gesture on the part of the orange vehicle driver, particularly if the blue vehicle had been waiting for a while or if the orange vehicle driver noticed that the blue vehicle was in a hurry.
    
    Alternatively, it is possible that the blue vehicle had the right of way to turn left onto the main road and the orange vehicle driver recognized this, and therefore stopped to allow the blue vehicle to turn safely.
    
    Without further information, it is difficult to determine the exact reason why the blue vehicle turned left onto the main road. 
    However, it is possible that the actions of the orange vehicle driver played a role in allowing the blue vehicle to make the turn.
\end{description}

\end{document}